\documentclass{article}

\PassOptionsToPackage{dvipsnames,table}{xcolor}
\usepackage{capt-of}
\usepackage{tikz,lipsum,lmodern}
\usepackage[most]{tcolorbox}
\tcbuselibrary{breakable}
\usepackage{ulem}

\usepackage[12pt]{extsizes}
\usepackage[utf8]{inputenc}
\usepackage{amsmath}
\usepackage{amssymb}
\usepackage{amsthm}
\usepackage{amsfonts}
\usepackage[numbers]{ai4science}
\makeatletter
\@ifundefined{ver@natbib.sty}{%
  \@namedef{ver@natbib.sty}{}}{}
\makeatother
\usepackage{graphicx}
\usepackage{lipsum}
\usepackage{enumitem}
\usepackage{caption}
\usepackage{mdframed}
\usepackage{tabularx}
\usepackage{pifont}

\newcommand{\authorblock}[1]{%
  \begin{center}
  {\authorfontsize #1}
  \end{center}
}
\newcommand{\authorrow}[1]{#1\\[0.3em]}

\newcommand{\aumark}[1]{\textsuperscript{#1}}
\newcommand{\emark}{\raisebox{-0.5ex}{*}}

\newcommand{\authorfontsize}{\fontsize{11}{13}\selectfont}

\usepackage{url}
\usepackage{subfigure}
\usepackage[subfigure]{tocloft}
\usepackage{bibunits}
\usepackage[dvipsnames,table]{xcolor}
\usepackage{hyperref}
\usepackage{mathtools}
\usepackage{relsize}
\usepackage{multirow}
\usepackage{booktabs}
\usepackage{tabularx}
\usepackage{bbding}
\usepackage{listings}
\usepackage{xspace}
\usepackage[ruled,vlined]{algorithm2e}
\usepackage{multirow}
\usepackage[p,osf]{cochineal}
\usepackage[scale=.95,type1]{cabin}
\usepackage[zerostyle=c,scaled=.94]{newtxtt}
\usepackage[cal=boondoxo]{mathalfa}
\usepackage{microtype}
\usepackage{multirow}
\usepackage{adjustbox}
\usepackage{etoolbox}
\usepackage{lmodern}
\usepackage{cleveref}
\usepackage{titletoc}

\usepackage{longtable}
\usepackage{rotating}
\usepackage[most]{tcolorbox}
\tcbuselibrary{breakable,skins,raster,listings}

\usepackage[font=small,labelfont=bf]{caption}

\crefname{appendix}{supplementary information}{supplementary information}
\Crefname{appendix}{Supplementary Information}{Supplementary Information}

\crefname{suppsection}{supplementary information}{supplementary information}
\Crefname{suppsection}{Supplementary Information}{Supplementary Information}

\crefname{suppfig}{supplementary figure}{supplementary figures}
\Crefname{suppfig}{Supplementary Figure}{Supplementary Figures}

\crefname{supptab}{supplementary table}{supplementary tables}
\Crefname{supptab}{Supplementary Table}{Supplementary Tables}

\usepackage{bibunits}
\defaultbibliographystyle{unsrtnat}
\defaultbibliography{ref}

\usepackage{csquotes}

\usepackage{geometry}
\definecolor{myurlcolor}{HTML}{123463}
\definecolor{dc_color}{RGB}{230, 245, 244}
\definecolor{ds_color}{RGB}{195, 230, 227}
\definecolor{ms_color}{RGB}{150, 214, 209}
\hypersetup{
    colorlinks=true,
    linkcolor=black,
    filecolor=black,      
    urlcolor=black,
    citecolor=black
}

\hypersetup{
   breaklinks=true,   
   colorlinks=true,  
   linkcolor=Bittersweet,
   citecolor=Periwinkle,
   urlcolor=gray
}
\definecolor{red}{HTML}{ca0020}
\definecolor{lightred}{HTML}{f4a582}
\definecolor{lightblue}{HTML}{92c5de}
\definecolor{green}{HTML}{008837}
\definecolor{blue}{HTML}{2c7bb6}

\apptocmd{\thebibliography}{\raggedright}{}{}

\makeatletter
\patchcmd{\@maketitle}{\LARGE \@title}{\fontsize{30}{19.2}\selectfont\@title}{}{}
\makeatother
\usepackage{amsmath,amsfonts,bm}

\def\eqref#1{equation~\ref{#1}}

\DeclareMathAlphabet{\mathsfit}{\encodingdefault}{\sfdefault}{m}{sl}
\SetMathAlphabet{\mathsfit}{bold}{\encodingdefault}{\sfdefault}{bx}{n}

\newcommand{\method}{\textsc{MolRSI}\xspace}

\definecolor{categoryheader}{RGB}{244,243,240}
\definecolor{ourshighlight}{RGB}{225,240,252}
\definecolor{codebg}{rgb}{0.97,0.97,0.95}
\definecolor{caseheader}{RGB}{159,225,203}
\definecolor{casebg}{RGB}{225,245,238}
\definecolor{casetext}{RGB}{4,52,44}
\definecolor{altheader}{RGB}{211,209,199}
\definecolor{altbg}{RGB}{241,239,232}
\definecolor{alttext}{RGB}{44,44,42}

\newtcblisting{promptbox}[1]{%
  enhanced, breakable,
  colback=codebg, colframe=gray!60,
  coltitle=white, fonttitle=\bfseries\small,
  title={#1},
  boxrule=0.5pt, arc=2pt,
  left=6pt, right=6pt, top=2pt, bottom=2pt,
  listing only,
  listing options={%
    basicstyle=\footnotesize\ttfamily,
    breaklines=true,
    columns=fullflexible,
    keepspaces=true,
    aboveskip=0pt,
    belowskip=0pt,
  },
}

\usepackage{placeins}
\newcommand{\namens}{MolRSI}
\newcommand{\name}{\namens\xspace}

\begin{document}

\title{\textbf{\method: A Recursive Self-Improving AI agent for Molecular Discovery}}
\date{}

\maketitle
\vspace{-6em}

\authorblock{
 \authorrow{
 Yikun Zhang\aumark{1,2,\emark},
 Xiwei Cheng\aumark{1,2,\emark},
 Mingze Yuan\aumark{3},
 Yu Zhang\aumark{2,4,5},
 Brian Shim\aumark{2,3},
 Aarti Krishnan\aumark{2,4,5},
 Daniel Rosen\aumark{2,14},
 }
 \authorrow{
 Kunyang Sun\aumark{6},
 Yingze Wang\aumark{6},
 Zhenfeng Deng\aumark{7},
 Botao Yu\aumark{8},
 Bingxu Liu\aumark{9},
 David Liu\aumark{2,3,10},
 Quanzheng Li\aumark{13},
 }
 \authorrow{
 Tianyu Liu\aumark{11,${\dagger}$},
 Yuanqi Du\aumark{12,${\dagger}$},
 Nir Hacohen\aumark{2,3,13,${\dagger}$},
 James J. Collins\aumark{2,4,5,${\dagger}$},
 Wengong Jin\aumark{1,2,${\dagger}$},
 }
}
\begingroup
 \renewcommand{\thefootnote}{}
 \footnotetext{\scriptsize
 $^{1}$Northeastern University, Boston, MA, USA;
 $^{2}$Broad Institute of MIT and Harvard, Cambridge, MA, USA;
 $^{3}$Harvard University, Cambridge, MA, USA;
 $^{4}$Wyss Institute for Biologically Inspired Engineering, Harvard University, Boston, MA, USA;
 $^{5}$Massachusetts Institute of Technology, Cambridge, MA, USA;
 $^{6}$University of California, Berkeley, Berkeley, CA, USA;
 $^{7}$University of Toronto, Toronto, ON, Canada;
 $^{8}$The Ohio State University, Columbus, OH, USA;
 $^{9}$University of Washington, Seattle, WA, USA;
 $^{10}$Howard Hughes Medical Institute, Cambridge, MA, USA;
 $^{11}$Yale University, New Haven, CT, USA;
 $^{12}$Microsoft Research New England, Cambridge, MA, USA;
 $^{13}$Massachusetts General Hospital, Boston, MA, USA;
 $^{14}$Brigham and Women's Hospital and Dana-Farber Cancer Institute, Boston, MA, USA.
 $^{\star}$These authors contributed equally.
 $^{\dagger}$Co-corresponding authors:
 Tianyu Liu, Yuanqi Du, Nir Hacohen, James J. Collins, and Wengong Jin.
 Contact: Wengong Jin (\texttt{wjin@broadinstitute.org}).
 }
\endgroup

\begin{bibunit}

\begin{abstract}

\noindent Molecular discovery requires navigating vast combinatorial search spaces while balancing competing biological objectives. AI agents can automate expert-defined molecular design protocols, but automation alone does not address how to optimize workflows for new tasks or improve their development through experience. Here we introduce \method, an AI agent that combines systematic optimization of workflows and model architectures with persistent cross-task memory to recursively improve its search policy for molecular discovery. Workflow performance, refinement outcomes, and execution feedback guide subsequent searches, allowing the agent to improve both its workflows and its strategy for developing them. We evaluate \method in two challenging molecular discovery settings that require balancing competing properties. In antibiotic discovery, \method developed a screening workflow balancing antibacterial activity, cytotoxicity and structural novelty, achieving an experimental hit rate of 15.9\%, compared with 8.7\% for a previous expert-led workflow using the same training data and screening library. In \textit{de novo} peptide–MHC binder design, \method developed workflows that jointly consider target binding and specificity, producing eight experimentally validated hits against MART-1 and SARS-CoV-1 pMHCs with preferential binding over tested off-targets while Claude Science completely failed. These findings demonstrate how experience-guided workflow optimization can address competing discovery objectives while improving the search strategies used in subsequent campaigns.

\end{abstract}

\section{Introduction}

Molecular discovery requires searching vast chemical and biological spaces for candidates that satisfy multiple competing objectives, including potency, binding specificity, toxicity, structural novelty, and stability \cite{wong2024discovery,liu2025design,bennett2023improving,dauparas2022robust}. These demands have been traditionally addressed by experimental workflows comprised of library preparation, screening, counter-screening, and hit validation \cite{wong2024discovery,liu2025design}. Recently, computational workflows have shown great success in molecular design \cite{watson2023novo,liu2025design,bennett2023improving}. They often involve different molecular representations, model architectures, training objectives, and optimization strategies with the available data \cite{yang2019analyzing,wu2018moleculenet}. They must also be integrated with evaluation and selection procedures that reflect the discovery objectives \cite{wong2024discovery,liu2025design}. Virtual screening, for example, requires jointly considering predictions of activity \cite{Liu2023Abaucin} and toxicity alongside structural constraints \cite{wong2024discovery,Stokes2020AntibioticDiscovery}, while protein design requires the coordination of target preparation, backbone generation \cite{watson2023novo}, sequence design \cite{dauparas2022robust}, and evaluation of binding and specificity \cite{liu2025design}. Decisions about both the models and their integration determine which candidates advance to experimental testing \cite{wong2024discovery,bennett2023improving}, making task-specific model and workflow optimization central to successful discovery.

Finding an effective workflow for a given task is a challenging search problem \cite{feurer2015autosklearn,zhang2025aflow}. The growing diversity of molecular representations, model architectures, training strategies, and optimization strategies creates a vast space of possible configurations whose performance varies across biological targets, available data, and desired properties (Figure 1a) \cite{yang2019analyzing,bennett2023improving,wu2018moleculenet}. Navigating this space requires an effective search policy for designing and training models, evaluating candidate workflows, and deciding how to customize them \cite{feurer2022autosklearn2,Toledo2025AIRA}. For human scientists, this policy is shaped by experience across projects. Previous successes and failures provide lessons in workflow and model engineering: which architectures suit the available data, which training strategies improve generalization, and which practical constraints must be addressed. These lessons help researchers identify strong starting points and prioritize beneficial changes (Figure 1b). As experience accumulates, scientists can adapt models and workflows to new problems with fewer rounds of trial and error. This expertise is particularly valuable, as the evaluation of each candidate requires expensive wet lab validation. Each project can contribute not only a workflow for a given task but also experience that improves the strategy used to develop future workflows.

Recent advances in scientific agents have begun to automate aspects of model and workflow development that traditionally depend on human expertise \cite{m2024augmenting,swanson2025virtual,miao2026paper2agent,aygun2026empirical,Lu2026AIScientist,Toledo2025AIRA,Schmidgall2025AgentLaboratory,Martinek2026Agentomics}. Biomni retrieves relevant tools, databases and software from a unified biomedical environment, then uses language model reasoning and code execution to compose them into workflows \cite{Huang2026BiomniScience}. In protein design, Claude Science has produced experimentally validated binders through campaigns guided by expert-written protocols specifying design objectives and evaluation criteria \cite{claudescience2026binder,claudescience2026scorpion}. Developing and executing these workflows produces evidence about which approaches succeed and which attempts fail \cite{aygun2026empirical}. However, the reported studies do not establish how this experience improves workflow development in subsequent problems. When campaigns remain isolated, lessons from one project cannot inform the search policy for the next. Retaining and applying these lessons across tasks would allow agents to build on earlier work \cite{fang2025memp,kim2026memory} and develop new workflows with less repeated trial and error.

\begin{figure}
    \centering
    \includegraphics[width=\linewidth]{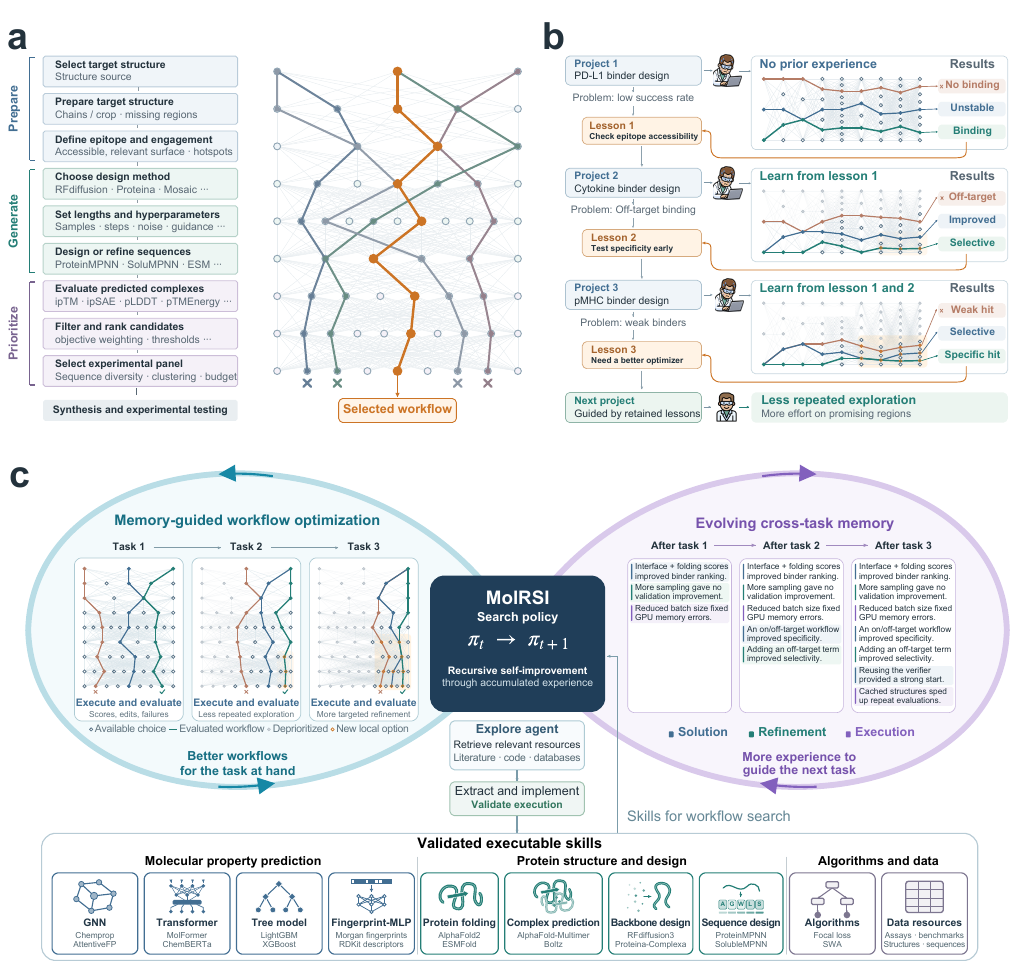}
    \caption{\textit{MolRSI motivation and architecture.}
    (a) Molecular discovery involves a combinatorial space of decisions across preparation, generation and prioritization, illustrated for \textit{de novo} protein binder design. Nodes represent options at each stage. The orange path traces one possible workflow. 
    (b) Scientists accumulate lessons from experimental outcomes across projects, learning to avoid unproductive choices and focus exploration on promising regions.
    (c) \method couples systematic workflow search with persistent cross-task memory. The Explore agent retrieves relevant resources from scientific literature and code repositories, then converts them into reusable, executable skills through implementation and validation. Workflow evaluations update solution memory (workflows and performance), refinement memory (modifications and their effects) and execution memory (execution strategies and repairs). Accumulated experience guides subsequent workflow selection, refinement and execution, recursively improving the effective search policy from $\pi_t$ to $\pi_{t+1}$. Colored paths indicate evaluated workflows, faded elements denote deprioritized choices, and orange-outlined nodes represent new local options. Projects, outcomes and memory entries are illustrative.}
    \label{fig:fig1}
\end{figure}

Here we introduce \method, a molecular discovery agent that couples model and workflow engineering with persistent cross-task memory to improve its search policy (\Cref{fig:fig1}c). The agent converts scientific methods and associated code into validated, executable skills, then uses experience-guided Monte Carlo tree search (MCTS) to assemble and refine candidate workflows, including their model architectures and training strategies. Inspired by how experienced researchers draw lessons from past projects, \method records these choices and their outcomes in persistent memory together with the context of each task. Past workflow performance and modification outcomes guide selection and refinement on subsequent tasks, while records of execution failures and verified repairs help the agent resolve similar problems. Each search adds to this experience and changes how later models and workflows are developed, forming a cycle of recursive self-improvement (RSI).

We evaluate \method in virtual antibiotic screening and \textit{de novo} peptide--MHC (pMHC) binder design, two challenging problems that require balancing competing objectives. These applications test our hypothesis that systematic optimization of models and workflows, guided by experience from previous tasks, can improve discovery under these constraints. In antibiotic discovery, \method draws on experience from 25 molecular property prediction tasks to develop a customized screening workflow balancing antibacterial activity, cytotoxicity and structural novelty. Experimental testing against \textit{Staphylococcus aureus} identifies 17 of 107 compounds with minimum inhibitory concentrations of 32$\mu$g/mL or lower. The observed hit rate of 15.9\% is substantially higher than the 8.7\% achieved by our previous expert-driven workflow using the same training data and screening library~\cite{wong2024discovery}. In pMHC binder design, \method draws on experience from 14 minibinder design tasks to develop customized workflows for pMHCs. Experimental screening identifies eight hits against MART-1 and SARS-CoV-1 pMHCs, each showing preferential enrichment for its intended target over the tested off-targets, whereas the workflow developed by Claude Science yields no hits. These results highlight the value of accumulated experience and task-specific optimization for molecular discovery under competing objectives.

\section{Results}
\label{sec:experiments}

\subsection{Overview of \method}

\method combines workflow search with persistent memory to implement recursive self-improvement (RSI) for molecular discovery (\Cref{fig:fig1}c). Each task specifies a biological problem, available data, an evaluation criterion and a computational budget. An Explore Agent organizes task-relevant scientific papers and associated code, tools, benchmarks and algorithms into validated, executable skills. These skills form the resource library from which \method constructs candidate workflows.

\method develops these workflows through experience-guided Monte Carlo tree search (MCTS), starting from configurations assembled from the resource library or retrieved from solution memory. Its search policy determines which workflows to explore, how to allocate search effort and how to refine and execute candidates.At each iteration, selection rules draw on workflows and scores in solution memory, combining performance on the current and related tasks with an exploration bonus \cite{kocsis2006bandit} for less-visited alternatives. Refinement memory supplies previous modifications and their outcomes to guide changes to the selected workflow. The agent then executes and evaluates the revised candidate, using execution memory to allocate resources and resolve errors. Search returns the workflow with the best development score within the task's computational budget.

Persistent memory carries the outcomes of these searches forward to guide later decisions. Each attempt updates the stored workflows, performance scores, modification outcomes and execution records. These updates change subsequent search decisions within the task and remain available for later tasks. Search experience can also lead to revised or additional skills. Searches on new tasks therefore use a policy informed by earlier outcomes and generate further experience that revises the policy applied to subsequent tasks, completing the RSI cycle (Methods and \Cref{app:explore,app:memory-schema,app:search-routing}).

We assess the contribution of this cross-task policy adaptation by comparing \method with and without RSI. Both settings use experiences obtained during the current task to guide search, whereas only \method with RSI also draws on experience from previous tasks. Zero-search transfer tests direct reuse of previously developed workflow configurations without new search. Model fitting on target training data and evaluation on target validation data are retained.

\subsection{MolRSI for antibiotic discovery}

\begin{figure}
    \centering
    \includegraphics[width=0.95\linewidth]{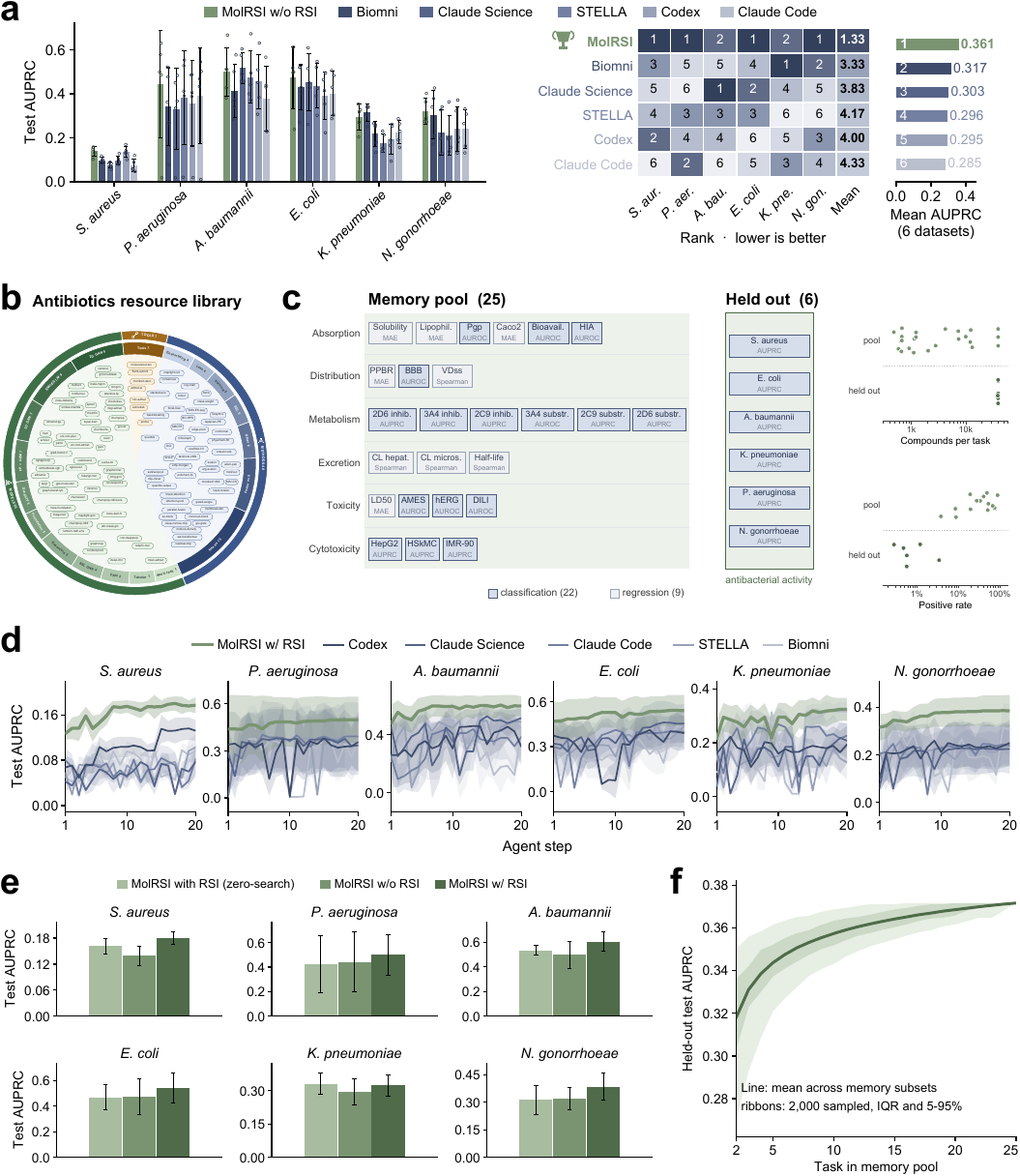}
    \caption{\textbf{Computational evaluation of \method for antibiotic discovery.}
(a) Test AUPRC and rankings of \method without RSI and baseline agents across six antibacterial tasks.
(b) Resource library.
(c) Experience from 25 source tasks, excluding the antibacterial targets.
(d) Search trajectories over 20 iterations.
(e) Search with and without RSI and zero-search transfer, which retains model fitting and validation.
(f) Zero-search transfer performance versus memory-pool size. The line shows test AUPRC averaged analytically across the six held-out antibacterial tasks and all possible subsets of each size from the 25-task memory pool. Darker and lighter shading indicate the interquartile range (IQR; 25th--75th percentiles) and the 5th--95th percentile range, respectively, across 2,000 sampled memory subsets at each pool size. In (a,d,e), means and $\pm1$ sample SD are shown across five data splits. Circles in (a) denote individual splits.}
    \label{fig:fig2}
\end{figure}

Antimicrobial resistance (AMR) is one of the most pressing global public health threats, steadily eroding our ability to treat common bacterial infections and imposing a large and growing burden of mortality worldwide\cite{gbd2024amr}.
The spread of antibiotic resistance creates an urgent need for compounds that inhibit bacterial growth while limiting toxicity to human cells. In our previous work~\cite{wong2024discovery},  we built an expert-driven virtual screening workflow that combined ensembles of Chemprop graph neural networks predicting activity against Methicillin-Resistant \textit{Staphylococcus aureus} (MRSA) with separate ensembles predicting cytotoxicity in human cell lines such as HepG2. We applied these models to more than 12 million compounds and used additional filters for undesirable substructures and similarity to known active compounds, followed by graph-based analyses that identified chemical substructures associated with predicted antibacterial activity. This workflow enabled the discovery of a new structural class of antibiotics. Developing such workflows requires navigating interdependent choices in molecular representation, model selection and training, screening criteria and candidate prioritization\cite{yang2019analyzing,wong2024discovery}. We therefore use antibiotic discovery to assess whether \method can autonomously find better virtual screening workflows and learn from its own searches to improve the search policy applied to subsequent tasks. We first benchmark its workflow search and then experimentally test compounds selected by the resulting virtual screening workflow.

\textbf{\method outperforms existing agents in antibacterial activity prediction benchmarks.}
Accurate prediction of antibacterial activity is central to prioritizing compounds for experimental testing. We therefore benchmarked \method and existing agents on six antibacterial activity prediction tasks, each corresponding to a different bacterial species (\Cref{fig:fig2}a). We compared \method with the scientific research agents Biomni, Claude Science and STELLA and general-purpose coding agents Codex and Claude Code~\citep{anthropic_claude_code}. For this comparison, we first evaluated \method without RSI, starting each task without experience from previous tasks. This establishes a baseline for assessing the contribution of RSI. \method develops workflows by selecting and refining combinations of models and methods from the full resource library of 110 reusable skills, comprising 55 models, 48 methods and seven tools or resources (\Cref{fig:fig2}b). All agents used the same five similarity-based data splits. Mean validation area under the precision--recall curve (AUPRC) guides workflow selection, and mean held-out test AUPRC measures predictive performance (Methods and \Cref{app:datasets,app:baselines}). \method achieves the highest mean test AUPRC among the evaluated agents, ranking first on four tasks and second on the remaining two.

\textbf{\method's performance and search efficiency improves as it accumulates experience from more tasks.}
We next tested whether experience from other molecular prediction tasks improves the search policy used to develop antibacterial workflows. \method first searched independently on 25 source tasks, comprising 22 Therapeutics Data Commons (TDC) \cite{huang2021therapeutics} tasks and three cytotoxicity tasks for HepG2, HSkMC and IMR-90 cells\cite{wong2024discovery} (\Cref{tab:tdc-source-tasks}). The six antibacterial benchmark tasks were excluded from this experience pool (\Cref{fig:fig2}c). \method used workflow performance, refinement outcomes and execution feedback from these searches to guide workflow selection, refinement and execution on the antibacterial tasks (Methods and \Cref{app:experience-accumulation}). Under the same search budget, \method generally achieves higher mean test AUPRC than the baseline agents across the search trajectories (\Cref{fig:fig2}d). On \textit{E.~coli} and \textit{N.~gonorrhoeae}, its first evaluated workflow already outperforms the final workflows selected by every baseline agent after 20 search iterations.

To assess the contribution of RSI, we compared searches with and without experience from previous tasks using the same initial resource library, backbone model and search budget (\Cref{fig:fig2}e). RSI improves mean test AUPRC on all six antibacterial tasks. We also included zero-search transfer to test whether previously developed workflows can be applied without further workflow optimization. This setting retains model fitting and validation on target data (\Cref{app:antibiotic-case-studies}). Zero-search transfer outperforms search without RSI on \textit{S. aureus}, \textit{A. baumannii} and \textit{K. pneumoniae}, achieving the highest mean test AUPRC among the three conditions on \textit{K. pneumoniae}. Experience therefore improves subsequent search and, on several targets, provides workflows that outperform search from scratch without further optimization.

We further examined whether the benefits of accumulated experience increase as more tasks contribute to it (\Cref{fig:fig2}f). We increased the number of source tasks contributing experience from 2 to 25 and assessed workflow transfer to the six antibacterial targets. We averaged performance across all combinations containing the same number of source tasks to ensure that the observed trend is not driven by a particular task order. The mean test AUPRC of the transferred workflows increases from approximately 0.32 with two source tasks to approximately 0.37 with all 25. With experience from all 25 tasks, this performance exceeds the mean AUPRC of 0.361 achieved by search without RSI after 20 iterations. These results show that accumulating experience across more tasks makes increasingly effective workflows available for subsequent tasks, extending the benefits of RSI from improved search to better performance (\Cref{app:experience-accumulation}).

\begin{figure}
    \centering
    \includegraphics[width=\linewidth]{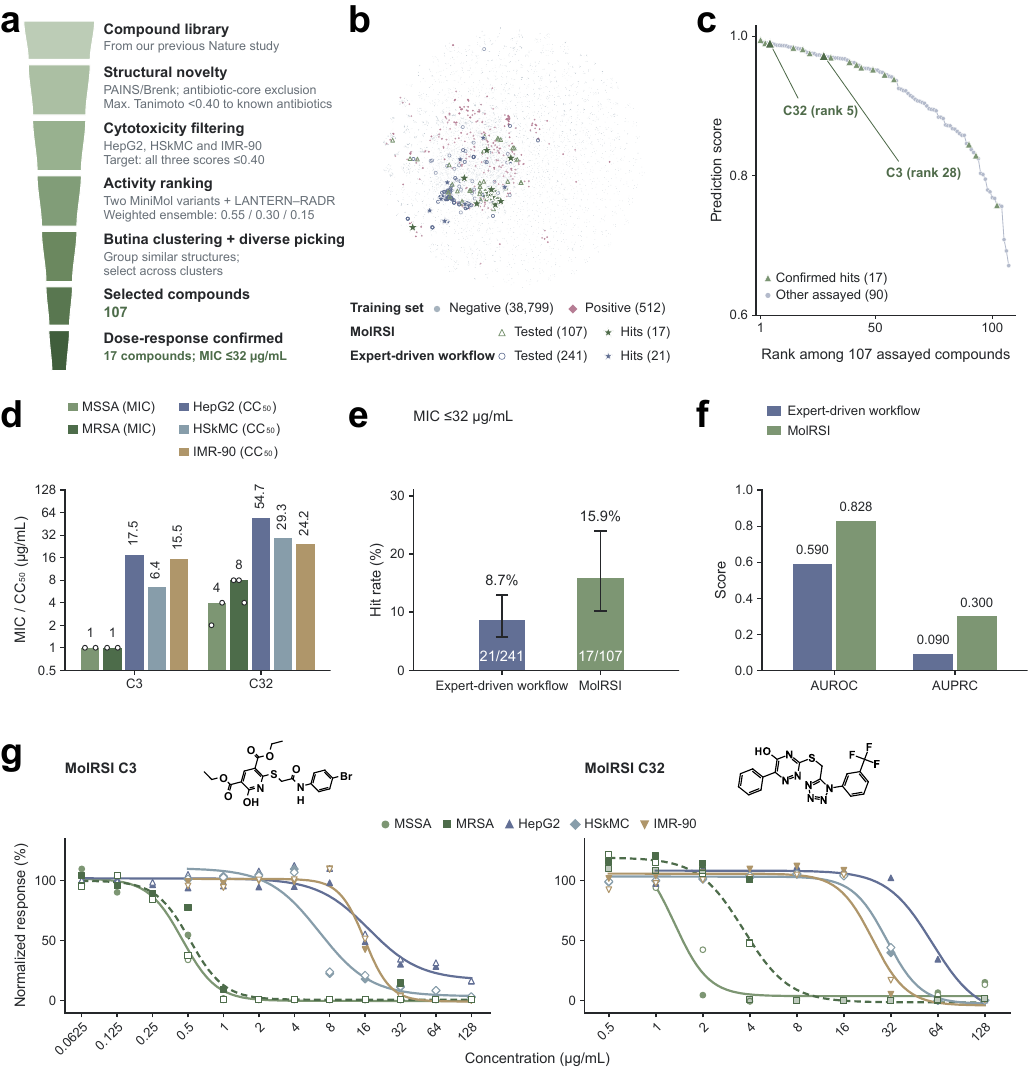}
    \caption{\textit{{Virtual screening and experimental validation with \method.}}
(a) Screening workflow yielding 107 tested compounds; cytotoxicity-filter exceptions are detailed in \Cref{app:screening-funnel}. 
(b) t-SNE projection~\citep{vandermaaten2008visualizing} of training and tested compounds; stars indicate hits. 
(c) Activity-score ranking of tested compounds. 
(d) MSSA and MRSA MICs and HepG2, HSkMC and IMR-90 $\mathrm{CC}_{50}$ values for two selected hits, C3 and C32; dots show individual MIC measurements. 
(e) Hit rates for 107 \method-selected and 241 expert-selected compounds~\cite{wong2024discovery}, with 95\% Wilson confidence intervals. 
(f) AUROC and AUPRC on the shared 271-compound held-out \textit{S. aureus} test set. 
(g) Chemical structures and dose--response curves of C3 and C32. Symbols show responses; lines indicate descriptive logistic fits. Two response series are shown per curve. Hits are defined by an MSSA MIC of 32$\mu$g/mL or lower.} 
    \label{fig:fig3}
\end{figure}

\textbf{\method identifies compounds with potent \textit{in vitro} activity against MRSA.}
We next evaluated whether the screening workflow developed by \method can identify antibacterial compounds while balancing activity, cytotoxicity and structural novelty. We used the same training datasets and screening library of approximately 12.1 million compounds as in our previous expert-driven study~\cite{wong2024discovery}. The workflow integrates optimized antibacterial activity and cytotoxicity predictors with structural filtering and diversity-based selection. Structural filters exclude specified antibiotic cores and compounds similar to known antibiotics, while cytotoxicity predictions for HepG2, HSkMC and IMR-90 cells inform candidate prioritization. Antibacterial activity ranking followed by selection across structural clusters yields 107 compounds for experimental testing (\Cref{fig:fig3}a; Methods; \Cref{app:screening-models,app:screening-funnel}).

Experimental validation identifies 17 compounds with minimum inhibitory concentrations (MICs) of 32$\mu$g/mL or lower against methicillin-susceptible \textit{S. aureus} (MSSA). The hits occupy multiple regions of the chemical space (\Cref{fig:fig3}b). Among these hits, C3 and C32 also exhibit potent activity against methicillin-resistant \textit{S. aureus} (MRSA), with MICs of 1 and 8$\mu$g/mL, respectively. C3 and C32 are ranked at 28th and 5th by predicted antibacterial activity among the tested compounds (\Cref{fig:fig3}c).
Both compounds inhibit MRSA at concentrations that do not cause cytotoxicity in each of the three types of human cells. Their half-maximal cytotoxic concentrations ($\mathrm{CC}_{50}$) range from 6.4 to 17.5$\mu$g/mL for C3 and from 24.2 to 54.7$\mu$g/mL for C32. These measurements establish a separation between antibacterial potency and mammalian cytotoxicity under the tested conditions (\Cref{fig:fig3}d,g; \Cref{app:antibiotic-assays}).

\textbf{\method improves experimental hit rate and antibacterial activity prediction over the previous expert-driven workflow.}
The 17 hits among 107 tested compounds correspond to an experimental hit rate of 15.9\%, compared with 8.7\% (21 of 241 compounds) in the previous expert-driven study at the same MSSA MIC threshold~\cite{wong2024discovery}. Thus, using the same training datasets and screening library, the workflow developed by \method achieves an approximately 82\% relative increase in observed hit rate (\Cref{fig:fig3}e; \Cref{app:antibiotic-experimental-results}).

To assess predictive performance independently of differences in compound selection, we compared the antibacterial activity predictors on a shared held-out set of 271 compounds experimentally evaluated in the previous study (\Cref{app:antibiotic-heldout}). \method achieves an area under the receiver operating characteristic curve (AUROC) of 0.828, compared with 0.590 for the previous predictor, and an area under the precision--recall curve (AUPRC) of 0.300, compared with 0.090 (\Cref{fig:fig3}f). The improved predictive discrimination on a common test set complements the higher prospective hit rate, demonstrating gains in both activity prediction and the yield of experimentally active compounds.

\subsection{MolRSI for \textit{de novo} peptide--MHC binder design}
\label{sec:minibinder}

\begin{figure}
    \centering
    \includegraphics[width=\linewidth]{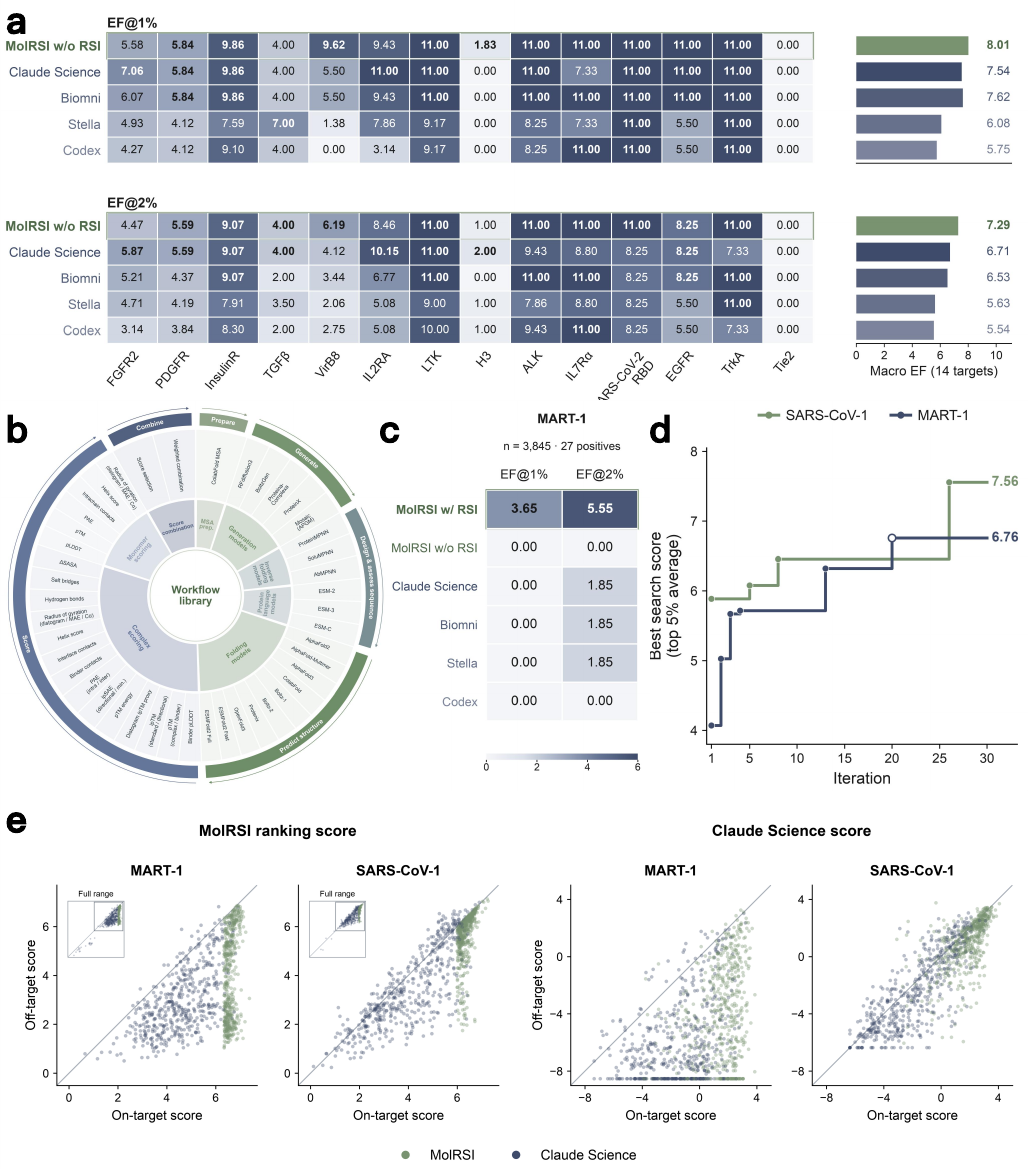}
    \caption{\textit{Benchmarking and experience-guided optimization of pMHC binder-design workflows.}
(a) Enrichment Factor (EF) at the top-ranked 1\% and 2\% cutoffs (EF@1\% and EF@2\%) for \method and four agent baselines across 14 miniprotein-binder tasks. Heatmaps show individual tasks; bars show the unweighted macro average across tasks.
(b) Protein resource library organized into input preparation, generation, sequence design and assessment, structure prediction, scoring and score combination.
(c) EF@1\% and EF@2\% on the held-out MART-1 pMHC target for \method with and without RSI and four agent baselines.
(d) Best-so-far search objective, defined as the mean ranking score among the top 5\% of candidates, during MART-1 and SARS-CoV-1 design-workflow optimization.
(e) On-target versus off-target scores for selected MART-1 and SARS-CoV-1 designs from \method and Claude Science, evaluated with the \method ranking score (left) and Claude Science score (right).
}
\label{fig:fig4}
\end{figure}

Class I major histocompatibility complex (MHC-I) molecules display intracellular peptides on cell surfaces, providing access to disease-associated antigens for engineered cell therapies and protein therapeutics\cite{liu2025design,Nathan2021Tebentafusp}. Designing binders for these pMHC complexes requires recognition of the exposed peptide while avoiding the same MHC presenting peptides from healthy cells. This task is demanding because target and off-target complexes share the large MHC scaffold and may differ only in a few exposed peptide residues. 
In our previous work, an expert-driven workflow combined RFdiffusion \cite{watson2023novo}, ProteinMPNN \cite{dauparas2022robust} and a pMHC-specialized AlphaFold2 model \cite{jumper2021highly,motmaen2023peptide} to generate small proteins that contact the presented peptide \cite{liu2025design}. The workflow produced binders that distinguished target pMHCs from closely related peptide-swapped off-targets and supported antigen-dependent T-cell activation when used in chimeric antigen receptors. Reaching this outcome required experts to choose peptide hotspots, generation strategies and criteria for both target binding and off-target discrimination. Here, we use this application to test whether \method can automate and improve this process in three stages: benchmarking binder prioritization across protein targets, applying the accumulated experience to pMHC ranking and prospective design, and experimentally screening the resulting candidates.

\textbf{\method improves binder prioritization than existing agents on a 14 target benchmark.}
Accurate candidate prioritization is central to recovering experimental hits from protein-design campaigns \cite{bennett2023improving}. Across 14 miniprotein targets, \method more effectively prioritizes experimentally confirmed binders than the comparison agents. We asked \method and four leading scientific or general-purpose agents~\citep{Huang2026BiomniScience,anthropic_claude_science,jin2025stella,openai_codex} to rank candidates in experimentally annotated tasks assembled from previous studies~\citep{cao2020novo,cao2022design,bennett2023improving}. Each task represents a different protein target, and all agents received the same inputs, objective and computational budget (\Cref{sec:protein-rsi-evaluation,app:protein-baselines}). We first evaluated \method without drawing on experience from previous tasks, while retaining access to the full resource library of 65 executable skills spanning candidate generation, structure prediction and scoring (\Cref{fig:fig4}b).
We assessed how effectively each workflow prioritizes experimentally confirmed binders using the Enrichment Factor (EF) at the top-ranked 1\% and 2\% of candidates (\Cref{sec:protein-benchmark-evaluation}). \method achieved the highest macro EF at both cutoffs: 8.01 at 1\%, compared with 7.62 for the next-best agent, and 7.29 at 2\%, compared with 6.71 for the next-best agent (\Cref{fig:fig4}a). Thus, even without prior-task memory, \method develops a stronger binder-ranking workflow than the comparison agents. Additional benchmark results and comparisons with fixed expert-defined objectives are reported in \Cref{app:protein-results,app:protein-expert-comparison}.

\textbf{Cross-task experience helps \method develop better pMHC binder design workflows.}
Drawing on experience from the miniprotein tasks, \method achieves better candidate ranking on a new pMHC task. \method developed and validated the pMHC ranking workflow using two gp100 datasets. MART-1 was reserved to test whether the resulting workflow generalizes to an unseen pMHC target (\Cref{sec:protein-benchmark-evaluation,app:protein-datasets}).
We compared the resulting workflow with those developed by the four agent baselines. \method achieves the highest MART-1 test enrichment, with EF@1\% and EF@2\% of 3.65 and 5.55, respectively (\Cref{fig:fig4}c). None of the comparison agents identifies an experimentally confirmed binder among the top-ranked 1\% of candidates, and the highest baseline EF@2\% is 1.85.
To isolate the contribution of RSI, we also compared against an otherwise identical \method workflow search without access to accumulated experience from the 14 miniprotein tasks. Without this experience, the same system achieves an EF of 0.00 at both cutoffs. 
Additional AUPRC and EF results are reported in \Cref{app:protein-results}.

We next used the ranking workflow developed with accumulated experience to guide pMHC design workflow optimization for MART-1 and SARS-CoV-1.
For each target, a peptide-swapped off-target retains the same MHC molecule, making peptide discrimination part of the design setting (\Cref{sec:method-pmhc-design,app:pmhc-generation}). \method kept the ranking procedure fixed while conducting a 30-round search over generation methods and settings, using the mean score of the top-ranked 5\% of candidates as the search objective. The best observed objective reaches 6.76 for MART-1 and 7.56 for SARS-CoV-1, improvements of 66\% and 28\% over the initial configurations, respectively (\Cref{fig:fig4}d).

We then compared the resulting pMHC design pipelines with Claude Science under the same target and off-target specifications, design constraints and computational budget.
We selected Claude Science for this comparison because it achieves the highest average macro EF among the agent baselines on the miniprotein benchmark and previously demonstrated autonomous design of experimentally validated protein binders~\citep{claudescience2026binder}. Each agent retained its own design workflow (\Cref{sec:method-pmhc-design,app:protein-workflow-comparison,app:claude-prospective}). 
To compare their outputs, we evaluated the top 500 selected candidates using the fixed \method ranking score and the sum of ipSAE$_{\min}$ z-scores \cite{dunbrack2025res} from the three prediction models used by Claude Science. Z-scores were calculated separately for each target and prediction model, pooling candidates from both design methods.
By both measures, \method designs have higher median on-target scores than Claude Science designs for MART-1 and SARS-CoV-1 (\Cref{fig:fig4}e).

\FloatBarrier 
\begin{figure}[!ht]
    \centering
    \includegraphics[width=\linewidth]{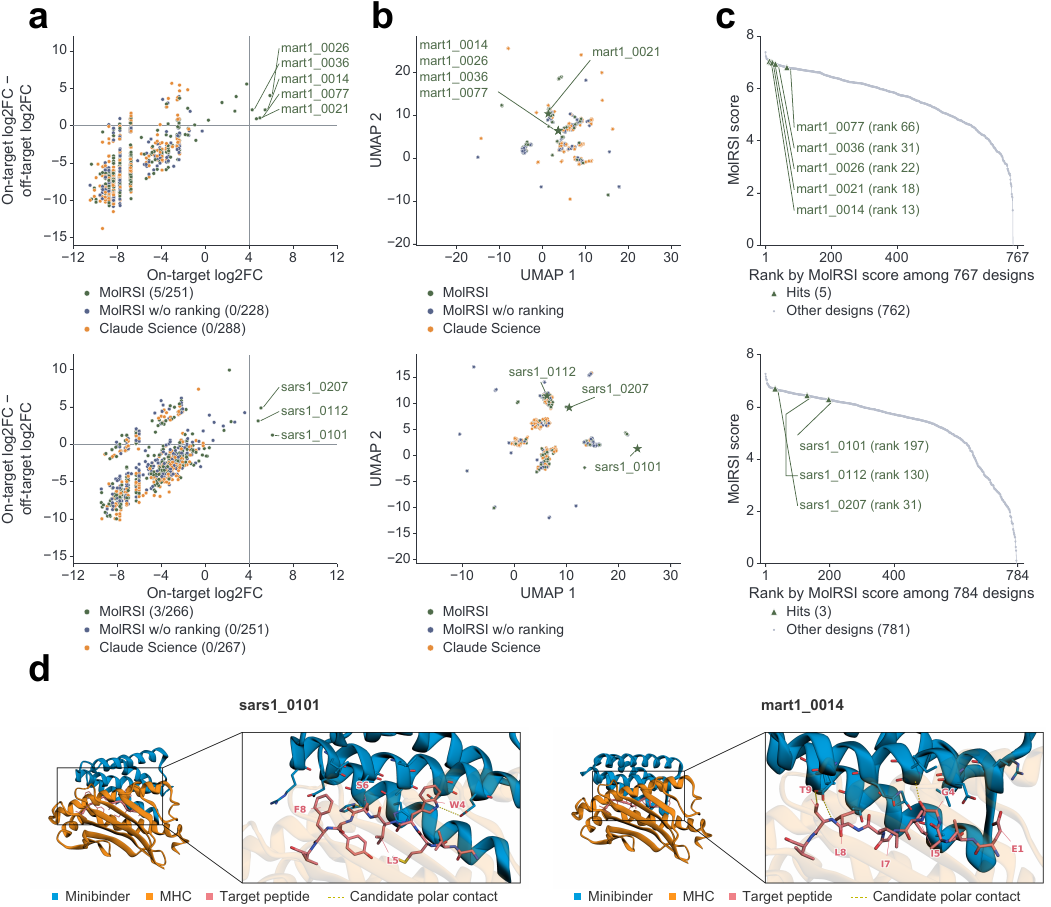}
    \caption{\textit{Prospective experimental validation of pMHC binder designs.}
Upper and lower rows in a--c show MART-1 and SARS-CoV-1, respectively.
(a) On-target log2FC and its difference from the maximum available off-target log2FC. Dashed lines mark hit thresholds ($x>4$, $y>0$); legend values indicate hits/evaluable designs. (b) UMAP~\citep{mcinnes2018umap} of predicted minibinder structures based on TM-align~\citep{zhang2005tmalign} similarity; stars mark experimental hits. (c) Designs ranked by \method score, with hits highlighted in green. (d) Predicted complexes of \texttt{sars1\_0101} and \texttt{mart1\_0014}. Minibinders, MHC and peptides are blue, orange and pink, respectively; yellow dashed lines indicate candidate polar contacts. } 
\label{fig:fig5}
\end{figure}

\textbf{Prospective experimental screening identifies specific pMHC binders designed by \method.}
We next ask whether the computational advantages of MolRSI translate into experimental binding. Candidates from \method and Claude Science are screened using an \textit{E.~coli} phage-display system through two rounds of in vitro selection against MART-1 (ELAGIGILTV) and SARS-CoV-1 (GLMWLSYFV) pMHCs. Sequencing-based enrichment is quantified for each design–pMHC pair, with hits requiring strong on-target enrichment (log2FC > 4) that exceeds the highest measurable off-target enrichment. \method yielded five MART-1 hits among 251 evaluable designs and three SARS-CoV-1 hits among 266, whereas Claude Science yielded no hits among 288 and 267 evaluable designs, respectively. Removing the learned ranking procedure likewise yielded no hits among 479 evaluable designs across the two targets, supporting the contribution of candidate prioritization to experimental success (\Cref{fig:fig5}a). This difference emerged within a broadly shared structural landscape: designs from all three methods occupied overlapping regions of the structural-similarity map, while the recovered hits spanned multiple structural neighborhoods (\Cref{fig:fig5}b). Experimental hits also tended to receive high \method scores. When evaluable designs from all three groups were ranked together, all five MART-1 hits fell within the highest-scoring 9\% of 767 designs, and the three SARS-CoV-1 hits fell within approximately the highest-scoring quarter of 784 designs, linking the computational score to prospective hit recovery (\Cref{fig:fig5}c). Predicted complexes of representative hits, \texttt{mart1\_0014} and \texttt{sars1\_0101}, positioned the minibinders over the exposed peptide and suggested contacts with multiple peptide residues, including potential polar interactions (\Cref{fig:fig5}d). These models are consistent with the peptide-engaging recognition mode sought by the workflow, while the screening results establish experimental enrichment and target preference for designs recovered by \method.

\section{Discussion}

\method improves molecular-discovery workflows through recursive self-improvement of the search policy used to develop them. Performance on related tasks changes the selection scores that determine where the agent directs further exploration. Current-task evaluations revise these priorities, while earlier modifications and execution outcomes guide refinement and repair. The policy therefore changes throughout workflow development and carries the effects of earlier searches into subsequent tasks. Those searches produce new results that revise later decisions, completing the recursive process. This links two levels of improvement. At the task level, the agent selects and refines computational methods to produce an effective workflow. Across tasks, the results of these attempts change how it approaches the next search.

The computational comparisons show that adapting this policy helps the agent make better use of a fixed search budget. Under matched conditions, incorporating prior-task results improves the workflows selected for antibacterial prediction and pMHC ranking. The antibacterial search trajectories also show that effective workflows can be found early, sometimes exceeding the final performance of baseline searches at the first evaluation. The resulting workflows can also remain useful beyond their original tasks. Direct transfer outperforms search from scratch on several antibacterial targets, providing an additional benefit of earlier workflow development. The retrospective pool analysis further identifies better available workflows as experience covers more source tasks. Together, these findings show how earlier searches can supply effective configurations and improve the policy used to develop new ones.

The experimental studies demonstrate why improved workflow development matters for molecular discovery. In antibiotic discovery, the \method workflow achieves a higher observed hit rate than the previous expert-driven workflow using the same training data and screening library~\cite{wong2024discovery}. Evaluation on a shared held-out compound set supports the predictive value of the workflow alongside its ability to prioritize active compounds for testing. The pMHC application extends the framework to design, where selecting candidates requires considering both target binding and discrimination against related off-targets. The ranking workflow developed with prior experience guides subsequent generation and selection, yielding binders with preferential binding to their intended complexes over tested off-targets. The workflow developed by Claude Science yields no hits under the same screening conditions. These experiments assess complete workflows that coordinate several computational methods. Their outcomes show the practical value of improving how such methods are selected, combined and refined before candidates advance to experimental testing.

These results also suggest how scientific agents that improve through use should be assessed. A successful final workflow establishes performance on a particular biological problem. Policy improvement concerns how the agent conducts subsequent searches, including which alternatives it explores and how effectively it uses the available budget. Comparisons between search with prior experience, search from scratch and direct transfer help distinguish the value of inherited configurations from the benefit of further workflow development. This makes the decisions involved in developing workflows open to systematic evaluation. Which methods are combined, which modifications are pursued and which attempts fail all provide evidence that can change later search priorities. Such evidence complements the scientific knowledge embodied in existing models and tools. Future benchmarks could assess how effectively an agent adapts its search policy over successive tasks.

Further progress depends on whether policy updates remain useful as biological targets and evaluation criteria change. The variation among experience pools highlights the importance of the contributing tasks, while current-task evaluation provides a way to reassess priorities inherited from earlier searches. Biological validation remains necessary to establish whether better computational scores lead to more effective molecular candidates. Future systems could incorporate experimental outcomes into policy updates, allowing laboratory successes and failures to influence which workflows receive further exploration. This would connect the improvement of search strategies more closely to the biological goals of discovery. More broadly, \method shows that scientific experience can be transferred between related tasks to improve the policy that guides workflow development. This provides a basis for discovery systems that refine their search strategies through use, with each project contributing to how future problems are approached.

\section{Methods}
\label{sec:method}

\subsection{\method framework}
\label{sec:method-architecture}

\subsubsection{Resource library construction}
\label{sec:skills}

\method uses an Explore Agent to build the resource library. The agent searches for scientific papers and their associated code repositories, tools, benchmarks and algorithms relevant to the task. It uses the task description and data characteristics to formulate queries about data representations, model architectures, training objectives and domain constraints. For each retrieved resource, the Explore Agent creates a reusable skill describing how to use it. Each skill specifies the resource's purpose, software dependencies, input and output requirements and execution steps, with references to available code, documentation and usage examples. The agent checks each skill's syntax, dependencies and interfaces during library construction. When a newly constructed skill is first used for a task, the agent runs a usage example in an isolated environment. If execution fails, it uses the error messages and repository information to revise the code, dependencies or environment settings, then reruns the example. Skills that pass execution validation are tagged as validated and retained for reuse. Skills that still fail after repair are removed, so that only validated skills are available for subsequent tasks. Full implementation details are given in \Cref{app:explore}.

\subsubsection{Persistent memory}
\label{sec:memory}

\method retains experience from workflow development and evaluation in solution, refinement and execution memory. These memories are updated during each task and retained for use in later tasks. \method uses the accumulated experience to guide later iterations of the same search and searches on new tasks. Experience from previous tasks could also be used to select an existing workflow without a new search, a mode referred to as zero-search transfer.

\textbf{Solution memory.} Solution memory stores each workflow's executable code, configuration, execution status and any available development scores. Development scores are calculated using the task-specific evaluation criterion to compare workflows during search. The configuration specifies the component methods, data processing, model architecture, parameter settings and training or generation procedures. Each score is stored with the task, dataset, evaluation metric, and data split and random seed identifiers where applicable. \method uses the stored workflows and scores to select workflows for further refinement on the current task or transfer to a new task.

\textbf{Refinement memory.} Refinement memory stores the original and revised workflows, the changes made and the reasons for proposing them. For revisions that complete evaluation, it also stores the development scores before and after the change, measured under the same task-specific evaluation protocol, and their difference. Modifications are retained whether performance improves, declines or remains unchanged. Revisions that fail during execution are stored with their failure status. \method uses these previous modifications and their outcomes when proposing further changes on the current task or related tasks.

\textbf{Execution memory.} Execution memory stores the computational requirements of workflows and the skills available in the resource library. Runtime, GPU time, processing speed and GPU memory use are stored together with the dataset, dataset size, hardware configuration and software environment. \method uses these measurements and their associated execution conditions to plan computing resources for later runs. When execution fails, execution memory stores the stage at which the failure occurs, the error messages and logs, and the cause when identified. It also stores each attempted repair, including changes to code, dependencies, environment settings or resource allocation, and whether the repair succeeds. When similar errors occur in the same or later tasks, \method retrieves the previous repairs and their outcomes to guide further debugging. Details of memory storage and updates are given in \Cref{app:memory-schema}.

\subsubsection{Recursive self-improvement of the search policy}
\label{sec:RSI for search policy}

\method uses experience from its own workflow searches to improve its search policy. Evaluation results from the current task and experience retained in persistent memory guide which workflows to explore, which modifications to pursue and how to resolve execution failures. Each attempt provides further evidence about workflow performance, the effects of modifications and execution requirements. This experience informs subsequent search decisions within and across tasks, completing the feedback loop through which recursive self-improvement occurs.

Workflow search uses experience-guided Monte Carlo tree search (MCTS)~\citep{kocsis2006bandit}. Initial workflows form the roots of separate trees, with each node representing a workflow. A workflow selected for modification is the parent, and a revised version is its child. Workflows based on a common model or computational approach are grouped into families. Each search iteration comprises the following five steps:

\textbf{Step 1: Selecting workflow families.} \method selects a workflow family using current task performance, an exploration bonus and a historical transfer score. For families already evaluated on the current task, the performance term uses the best development score obtained in earlier iterations. The exploration bonus \cite{kocsis2006bandit} is larger for less visited families. The historical transfer score summarizes the family's performance on related tasks using solution memory. Historical experience guides initial exploration, and current task scores are incorporated as evaluations become available. The family with the highest combined score is selected. The selection rule and historical transfer score calculations are given in \Cref{app:search-routing}.

\textbf{Step 2: Selecting parent workflows.} \method samples a parent workflow from those already evaluated on the current task. Sampling favours better development performance, penalizes workflows that have already produced more candidates, and incorporates historical transfer scores from solution memory. The sampling rule and eligibility criteria are given in \Cref{app:search-routing}.

\textbf{Step 3: Proposing workflow modifications.} \method retrieves previous modifications and their outcomes on related tasks from refinement memory. It uses these modifications and outcomes to propose changes to the selected parent workflow, such as replacing a method, changing how components are combined or adjusting parameter settings. Each proposal specifies the changes and their rationale. Before execution, \method checks that the proposed workflow follows the required format, does not duplicate an existing candidate and is feasible to run.

\textbf{Step 4: Executing and evaluating workflows.} Using the resource requirements stored in execution memory, \method configures the runtime environment and allocates computing resources. It then runs the revised workflow and assesses its performance using the task-specific evaluation procedure. If execution fails, \method first searches execution memory for similar errors and the repairs that have resolved them. When relevant experience is available, it uses the previous repair to guide changes to the code, dependencies or environment settings. Otherwise, it diagnoses the problem from the current error messages and makes the necessary changes. The workflow is rerun after each repair. If execution still fails, \method uses the new error messages to guide further changes, up to a predefined limit on repair attempts and within the task's computational limits. Successful evaluations provide development scores for subsequent search iterations.

\textbf{Step 5: Updating search policy to guide subsequent searches.} After each execution attempt, \method stores the workflow, its execution status and any available development score in solution memory. It adds the modification and its outcome to refinement memory and stores runtime, resource use, errors and repair outcomes in execution memory. \method then uses the updated memories to select families and parents, propose modifications, allocate resources and repair execution failures in later iterations (Steps 1--4).

We denote the effective policy governing the decisions in Steps 1--4 by \(\pi_t(a\mid s)\), the distribution over search actions \(a\) for the current task and search state \(s\). Let \(M_t\) denote persistent memory after \(t\) updates and \(E_t\) the new execution and evaluation experience. The policy before and after a memory update is expressed as
\begin{equation}
\begin{aligned}
\pi_t(a\mid s) &= \pi(a\mid s;M_t),\\[4pt]
M_{t+1} &= \operatorname{Update}(M_t,E_t),\\[4pt]
\pi_{t+1}(a\mid s) &= \pi(a\mid s;M_{t+1}).
\end{aligned}
\label{eq:memory-conditioned-policy}
\end{equation}

The updated policy guides further searches, whose results supply experience for the next memory update (\Cref{fig:fig1}). This feedback loop implements recursive self-improvement of the effective search policy. \method uses the outcomes of its own searches to focus exploration on promising workflows, reduce repeated unsuccessful modifications and reuse effective repairs. The same process operates within a task and across tasks as experience accumulates. Search stops when the task-specific limit on evaluation attempts or computation time is reached. \method returns the successfully evaluated workflow with the best development score on the current task.

\method uses experience from previous tasks to select a workflow for direct transfer or as the starting point for a new search. It first identifies the best matching previously completed task, referred to as the source task, using task type, evaluation metric, dataset size and task description similarity. It then retrieves the workflow with the best development score on that source task from solution memory. In zero-search mode, the retrieved configuration is applied to the new task without searching over alternative workflows. Model parameters are fitted using the new task's training data where required. Alternatively, the retrieved workflow serves as the starting point for further search using Steps 1--5, within the task-specific limits on evaluation attempts or computation time.

\subsection{MolRSI for antibiotic discovery}
\label{sec:method-molecule}
\subsubsection{Datasets and benchmark evaluation}
\label{sec:method-molecule-model}

\textbf{Source tasks for experience accumulation.} We used 25 source tasks to accumulate experience for antibiotic workflow development: 22 absorption, distribution, metabolism, excretion and toxicity tasks from the Therapeutics Data Commons (TDC) \citep{huang2021therapeutics}, together with three cytotoxicity tasks for HepG2, HSkMC and IMR-90 cells from our previous study\citep{wong2024discovery}. The TDC tasks used the official benchmark partitions and task-specific evaluation metrics. The cytotoxicity datasets were obtained from our previous study, with the same compound structures and experimental measurements. Each cytotoxicity dataset was divided into training, validation and test sets at proportions of 8:1:1 using molecular similarity, with a Tanimoto threshold of 0.40. Five splits were generated using different random seeds.

\textbf{Antibacterial target tasks.} We evaluated \method on six antibacterial activity prediction tasks for \textit{Staphylococcus aureus}, \textit{Pseudomonas aeruginosa}, \textit{Acinetobacter baumannii}, \textit{Escherichia coli}, \textit{Klebsiella pneumoniae} and \textit{Neisseria gonorrhoeae}. These tasks were excluded from the source experience pool. The datasets were obtained from our previous study, with the same compound structures and experimental measurements. Each dataset used five similarity-based training, validation and test splits with a Tanimoto threshold of 0.40, proportions of 8:1:1 and the same five random seeds used for the cytotoxicity tasks. All methods used identical splits. Each task involved a single workflow search, during which every candidate was fitted and evaluated across all five splits. Mean validation AUPRC guided the search, and the workflow with the highest value was selected. Test AUPRC is reported as the mean and sample standard deviation across the five splits. Test outcomes were not used to guide workflow search.

\textbf{Activity \textit{S. aureus} held-out test set.} For the predictive comparison in \Cref{fig:fig3}f, we used a held-out set of 271 compounds from our previous study~\citep{wong2024discovery}. This set comprised 241 candidate compounds and 30 low predicted activity controls; the 12 known antibiotic class positive controls were excluded. \method and the published antibacterial activity model were evaluated on the same compounds using the same experimental activity labels. AUROC and AUPRC are reported for both models.

Dataset sizes and evaluation metrics for all datasets are provided in \Cref{app:datasets}, together with the split sizes and random seeds used for workflow development.

We benchmarked \method with RSI and \method without RSI against Biomni, Claude Science, STELLA, Codex and Claude Code on the six antibacterial target tasks (\Cref{fig:fig2}). All agents received the same task prompt, datasets and development objective, with up to 20 attempted workflow evaluations per task. Each agent retained its native tools and workflow development procedures. \method and all agent baselines except Codex used Claude Sonnet~5; Codex used GPT-5.5. \method accessed its backbone model through the Anthropic API, whereas Biomni ran on its own platform. Complete task prompts, model identifiers, API settings and baseline configurations are provided in \Cref{app:antibiotic-baselines}.

\subsubsection{Evaluation of recursive self-improvement}

\method searched for workflows independently on the 25 source tasks described in \Cref{sec:method-molecule-model}. Experience from these searches was pooled in persistent memory before evaluation on the antibacterial target tasks.

We compared \method with and without RSI. \method with RSI began with the accumulated source memory, whereas \method without RSI began without experience from previous tasks. Both conditions used the same resource library and backbone model, together with the data partitions, evaluation procedure and workflow evaluation budget specified in \Cref{sec:method-molecule-model}. Memory was updated during each search to guide subsequent iterations. Experience generated on one antibacterial task was not passed to the other antibacterial tasks.

For the search trajectories in \Cref{fig:fig2}d, we identified the workflow with the highest mean validation AUPRC obtained so far at each iteration. Its test performance was calculated after the search to construct the plotted trajectory.

For zero-search transfer (\Cref{fig:fig2}e), \method first identified the best matching source task using task type, evaluation metric, dataset size and task description similarity. It then selected the workflow with the best validation performance on that source task. Both selections were made before evaluating the workflow on the antibacterial task. Model parameters were fitted using the target task's training data where required, and the transferred workflow was evaluated without further workflow search.

We compared the transfer performance available from different numbers of source tasks with the performance of \method without RSI after workflow search (\Cref{fig:fig2}f). One workflow from each of the 25 source tasks was transferred to each antibacterial task, yielding 150 source target evaluations. Experience pools contained 2--25 source tasks. For each pool, the transfer test AUPRC on each antibacterial task was identified retrospectively, and these values were averaged across the six tasks. The mean was calculated over all source task combinations of each size, with variation among combinations estimated from 2,000 random orderings of the source tasks. The transfer evaluations and calculation of means and percentile ranges are described in \Cref{app:experience-accumulation,app:antibiotic-search}.

\subsubsection{Virtual screening and compound selection}

\method uses the experience accumulated across source tasks to guide the development of a virtual screening workflow for identifying candidate antibiotics against methicillin-resistant \textit{Staphylococcus aureus} (MRSA). The workflow combines antibacterial activity prediction with cytotoxicity prediction for HepG2, HSkMC and IMR-90 cells. The activity models were trained on methicillin-susceptible \textit{S. aureus} (MSSA; strain RN4220). All activity and cytotoxicity models used the same training datasets as our previous study \citep{wong2024discovery} and both studies used the same screening library.

Screening began with structural filtering. Compounds with a maximum Tanimoto similarity of 0.40 or greater to the antibacterial reference set were excluded. Structural filtering also removed compounds matching PAINS \cite{baell2010pains} or Brenk alerts \cite{brenk2008lessons} and specified antibiotic cores. Cytotoxicity was then predicted separately for the three cell types. The HepG2 and IMR-90 predictors combined MiniMol embeddings~\citep{klaser2024minimol} with Morgan fingerprints~\citep{rogers2010extended} and RDKit descriptors~\citep{rdkit}, whereas the HSkMC predictor used MiniMol embeddings. Cytotoxicity filtering used a score threshold of 0.40 for each cell type. Model architectures, reference compounds and filter settings are described in \Cref{app:screening-models,app:screening-funnel}.

The remaining compounds were ranked by a weighted combination of predictions from two MiniMol-based variants and a LANTERN--RADR ensemble, with weights of 0.55, 0.30 and 0.15, respectively. Each component prediction was averaged across five fitted models. Butina clustering~\citep{butina1999unsupervised} grouped structurally similar compounds, and candidates were selected across clusters to retain chemical diversity.

We selected and experimentally tested 107 compounds. Activity against MRSA was assessed during experimental validation. Compound identities, prediction scores and sourcing details, including supplier substitutions and exceptions to the cytotoxicity criterion, are provided in \Cref{app:screening-funnel}.

\subsubsection{Experimental validation}

\textbf{Antibacterial activity assays.} Antibacterial activity was measured against methicillin-susceptible \textit{Staphylococcus aureus} RN4220 (MSSA; DSM 26309) and methicillin-resistant \textit{S. aureus} USA300 FPR3757 (MRSA; ATCC BAA-1556). Compounds were first screened at a single dose of $128\,\mu\mathrm{g\,mL^{-1}}$, and those that inhibited growth were advanced to two-fold dose--response testing from $128\,\mu\mathrm{g\,mL^{-1}}$ (8--9 concentrations, to 1 or $0.5\,\mu\mathrm{g\,mL^{-1}}$). Overnight LB cultures grown from single colonies were diluted to approximately $10^{5}\,\mathrm{CFU\,mL^{-1}}$ in fresh LB and added to plates containing compound in DMSO (final DMSO $\leq1\%$). Plates were covered with a lid and incubated for 18--24~h at $37\,^{\circ}\mathrm{C}$ with 5\% $\mathrm{CO}_{2}$. Growth was measured as $\mathrm{OD}_{600}$. The minimum inhibitory concentration (MIC) was defined as the lowest concentration giving $\geq90\%$ growth inhibition relative to untreated growth controls, and compounds with an MSSA MIC $\leq32\,\mu\mathrm{g\,mL^{-1}}$ were defined as hits. DMSO vehicle, untreated growth and Triton X-100 kill controls were included. MICs were determined in technical replicate ($n=2$-$3$) in a single experiment, and the higher replicate value was reported.

\textbf{Cytotoxicity assays.} Cytotoxicity was measured in human hepatocellular carcinoma (HepG2; ATCC HB-8065), primary human skeletal muscle (HSkMC; ATCC PCS-950-010) and human fetal lung fibroblast (IMR-90; ATCC CCL-186) cells. Cells were seeded at $5$--$10\times10^{3}$ cells per well in 96-well clear flat-bottom black tissue-culture-treated plates ($100\,\mu\mathrm{L}$ per well) and incubated for 24~h at $37\,^{\circ}\mathrm{C}$ and 5\% $\mathrm{CO}_{2}$ before dosing. Compounds were added as a two-fold dilution series from 128 to $0.5\,\mu\mathrm{g\,mL^{-1}}$, with 1\% DMSO at the highest concentration diluted in step with compound, and cells were incubated for 48--72~h. Viability was quantified with PrestoBlue HS ($10\,\mu\mathrm{L}$ per well, 4~h at $37\,^{\circ}\mathrm{C}$) by fluorescence (excitation/emission, 550/590~nm). Untreated cells defined 100\% viability, with a matched DMSO series and Triton X-100 (0.1\%) lysis wells as vehicle and floor controls and media-only wells as background. The half-maximal cytotoxic concentration ($\mathrm{CC}_{50}$) was calculated by log-linear interpolation between the two concentrations bracketing 50\% viability. Each compound and concentration was tested in two technical wells in a single experiment.

\subsection{MolRSI for peptide–MHC binder design}
\label{sec:method-protein}
\subsubsection{Datasets and benchmark evaluation}
\label{sec:protein-benchmark-evaluation}

The protein study instantiated the cross-task update framework in three stages. First, ranking workflow development on 14 miniprotein-binder tasks supplied protein-domain source experience. Second, \method used the resulting experience memory to develop workflows for one pMHC binder design task. Third, the selected pMHC workflow guided prospective binder design workflow optimization for MART-1 and SARS-CoV-1 pMHC complexes.

\textbf{Miniprotein-binder tasks.} We assembled experimentally annotated de novo miniprotein-binder datasets for 14 protein targets from three published studies~\citep{cao2020novo,cao2022design,bennett2023improving}. Each target defined a task using a leave-one-target-out partition: workflow search used the other 13 targets, and the selected workflow was then evaluated on the held-out target. We retained all 1,565 positive designs and sampled up to ten negatives per positive within each target, yielding 17,215 designs. Target identities, sample counts and label definitions are reported in \Cref{app:protein-datasets}.

\textbf{pMHC task.} The pMHC benchmark comprised three target datasets: two for gp100 and one for MART-1. Both gp100 datasets were used directly for workflow search, while MART-1 was reserved for testing.
Each candidate was represented as a four-chain complex comprising the HLA heavy chain, $\beta_2$-microglobulin, the presented peptide and the designed binder. The datasets were generated using the experimental screening procedure described by Liu et al.~\citep{liu2025design}. Peptide identities, dataset sizes and label definitions are reported in \Cref{app:protein-datasets}.

\textbf{Ranking metrics.} We evaluated candidate rankings using the area under the precision--recall curve (AUPRC), the area under the receiver operating characteristic curve (AUROC) and the Enrichment Factor (EF). EF measures the positive fraction among top-ranked candidates relative to the positive fraction in the full target dataset. For a target with \(N\) candidates and \(N_+\) experimentally positive binders, let \(n_f\) be the number retained at the top-ranked fraction \(f\) and \(n_{+,f}\) the number of positives among them. We calculated \begin{equation} \mathrm{EF}_{f}=\frac{n_{+,f}/n_f}{N_+/N},\qquad f\in\{0.01,0.02,0.05,0.1\}. \label{eq:protein-ef} \end{equation} We denote these metrics as EF@1\%, EF@2\%, EF@5\% and EF@10\%, respectively. Random ranking has an expected EF of 1. At each cutoff, macro EF is the unweighted average across the 14 miniprotein tasks (\Cref{fig:fig4}a). For the pMHC benchmark, EF is reported separately for the held-out MART-1 target (\Cref{fig:fig4}c). Full EF results at all four cutoffs and AUPRC results are reported in \Cref{app:protein-results}. Comparisons with fixed expert-defined scoring criteria are provided in \Cref{app:protein-expert-comparison}.

\textbf{Design targets.} We designed binders for HLA-A*02:01 complexes presenting MART-1 (\texttt{ELAGIGILTV}) and SARS-CoV1 (\texttt{GLMWLSYFV}). Each intended pMHC was paired with an off-target complex presenting a different peptide while retaining the same HLA heavy chain and $\beta_2$-microglobulin. Target sequences and exposed peptide hotspots were derived from source structures. The source coordinates were not supplied as design inputs. Off-target peptide sequences, source structures and hotspot definitions are given in \Cref{tab:pmhc-design-inputs}.

\subsubsection{Evaluation of recursive self-improvement}
\label{sec:protein-rsi-evaluation}

\textbf{Resource library and workflow search.} \method selected and configured components from the protein resource library, which contained 65 executable skills comprising 28 models and 37 registered scoring terms (\Cref{fig:fig4}b). For ranking workflow development, these components covered input and feature preparation, complex-structure prediction, sequence and interface scoring, and score aggregation. Each candidate workflow mapped a target--binder pair to a scalar ranking score. AUPRC on development data from the 13 non-held-out targets guided workflow search and selection within each miniprotein-binder task, whereas AUROC guided search and selection for the single pMHC task.

\textbf{Benchmark baselines and budget.} We compared \method with Biomni~\citep{Huang2026BiomniScience}, Claude Science~\citep{anthropic_claude_science}, STELLA~\citep{jin2025stella} and Codex~\citep{openai_codex}. For each benchmark task, all agents received the same task inputs, data partitions, and evaluation objective, with a budget of 300 A100 GPU-hours and 20 attempted workflow evaluations. Each agent retained its native tools and workflow-development procedures. Folded structures and the scores produced by each model were cached and reused across workflow evaluations to avoid repeated computation. Task instructions and baseline configurations are provided in \Cref{app:baselines,app:protein-baselines}.

\textbf{Memory ablation.} We compared pMHC workflow search initialized with pooled experience from the 14 miniprotein-binder tasks against search without this prior experience. The initial memory contained no pMHC or prospective design outcomes. Both conditions used identical resource libraries, candidate data, search controllers, evaluation objectives and procedures, and the per-task budget specified above. Memory was updated during search in both conditions, so the comparison isolated the contribution of prior cross-task experience.

\subsubsection{Binder generation and candidate selection}
\label{sec:method-pmhc-design}

\textbf{Generation-workflow optimization.} The prospective pipeline combined candidate generation, hotspot-contact filtering, inverse folding and ranking. \method optimized generation methods and configurations separately for each target, using RFDiffusion3~\citep{butcher2025novo}, Proteina-Complexa~\citep{didi2026scaling}, PXDesign~\citep{team2025pxdesign}, BoltzGen~\citep{stark2025boltzgen} and the MOSAIC~\citep{boyd2025mosaic} simplex accelerated proximal gradient method (APGM). Each target-specific search received up to 30 attempted rounds and 1,000 A100 GPU-hours. The searches ran in parallel without sharing evaluations or search records.

For each evaluated configuration, we generated 20 backbones with binder lengths of 80--120 residues. SolubleMPNN~\citep{goverde2024computational} generated eight sequences per retained backbone, with cysteine excluded from the sequence alphabet. The ranking procedure selected in the memory-guided pMHC search remained fixed throughout prospective search and candidate selection. The mean score among the top-ranked 5\% of generated candidates served as the configuration-selection objective. 

\textbf{Final candidate generation and selection.} After each target-specific search, the three highest-scoring workflows each generated 8,000 additional designs. Candidates from the search and expansion stages were pooled and deduplicated. Final \method candidates were required to satisfy $\texttt{on\_target\_ipsae\_min}\geq0.50$, $\texttt{monomer\_plddt}\geq0.80$, at least 15 HLA contacts ($\texttt{hla\_res\_contact}\geq15$), and $s_{\mathrm{on}}-s_{\mathrm{off}}>0$. Passing designs were ranked by the on-target frozen-verifier score $s_{\mathrm{on}}$. Experimental outcomes are reported separately over evaluable designs, as defined below. Detailed generation and selection settings are provided in \Cref{app:pmhc-generation}. The Claude Science protocol is described in \Cref{app:protein-baselines,app:protein-workflow-comparison}.

\subsubsection{Experimental validation}
\label{sec:method-pmhc-validation}
Biotinylated pMHC monomers were purchased from KACTUS Bio and comprised HLA-A*02:01 (Gly25--Thr305), $\beta_2$-microglobulin (Ile21--Met119) and the MART-1 (ELAGIGILTV) and SARS-CoV-1 (GLMWLSYFV) peptide.
Binder designs were synthesized by Twist Biosciences and cloned into engineered M13 bacteriophage lacking gIII as previously described~\cite{esvelt2011system}. Phage libraries were then selected for specific binding activity using host E. coli cells encoding positive selection circuits for on-target pMHC binding activity and negative selection circuits for off-target pMHC binding using an adaptation of previously described methods~\cite{badran2016continuous,styles2025pancs}. Next generation sequencing of phage-encoded designs was used to assess enrichment of functional designs throughout selection.

\section*{Data Availability}
The molecular property prediction, antibacterial activity, human-cell cytotoxicity and miniprotein-binder datasets used for computational workflow development and benchmarking are publicly available. Molecular-property prediction datasets were obtained from the Therapeutics Data Commons~\citep{huang2021therapeutics}. Antibacterial activity and human-cell cytotoxicity datasets, including the 271-compound held-out \textit{S. aureus} test set, were obtained from our previous study~\citep{wong2024discovery}. Miniprotein-binder datasets were obtained from the studies by Cao et al.~\citep{cao2020novo,cao2022design} and Bennett et al.~\citep{bennett2023improving}. The pMHC binder datasets from Liu et al.~\citep{liu2025design} are available from the corresponding authors upon reasonable request. Dataset sources, composition and partitions are described in \Cref{app:datasets,app:protein-datasets}.

\section*{Acknowledgments}
Wengong Jin acknowledges support from the Google Research Scholar Award, the NVIDIA Academic Grant Program, the Google TPU Research Cloud Award, and NSF ACCESS.
Nir Hacohen acknowledges support from the Torrey Coast Foundation and NIH/NCI IMAT R61 CA281807.

\section*{Author Contributions}
Coordination and planning: Wengong Jin;
Agent framework implementation: Yikun Zhang, Xiwei Cheng;
Computational design of antibiotics: Xiwei Cheng, Yikun Zhang;
Computational design of proteins: Xiwei Cheng, Yikun Zhang, Mingze Yuan;
Antibiotic benchmark data provision: Yu Zhang, Aarti Krishnan, James J Collins;
pMHC binder data provision: Bingxu Liu;
Experimental validation of antibiotics: Yu Zhang, Aarti Krishnan;
Experimental validation of pMHC binders: Brian Shim, Daniel Rosen;
Supervision on experimental validation: David Liu, Nir Hacohen, James J Collins.
Supervision of methodology: Tianyu Liu, Yuanqi Du, Wengong Jin.
All authors reviewed and edited the manuscript.

\section*{Competing Interests}
The authors declare that they have no conflict of interests at this time.

\clearpage
\putbib
\end{bibunit}

\clearpage

\makeatletter
\def\NAT@hyper@#1{%
\hyper@natlinkstart{supp.\@citeb\@extra@b@citeb}#1\hyper@natlinkend}
\def\NAT@anchor#1#2{%
\hyper@natanchorstart{supp.#1\@extra@b@citeb}%
  \def\@tempa{#2}\@ifx{\@tempa\@empty}{}{\@biblabel{#2}}%
\hyper@natanchorend}
\makeatother

\begin{bibunit}
\appendix
\emergencystretch=2em

\crefalias{section}{suppsection}
\crefalias{figure}{suppfig}
\crefalias{table}{supptab}

\begin{center}
\vspace*{1em}
{\fontsize{30}{19.2}\selectfont\textit{Supplementary Information} for \name\par}
\vspace{1.5em}
\end{center}

\setcounter{section}{0}
\renewcommand{\thesection}{S\arabic{section}}
\renewcommand{\thefigure}{S\arabic{figure}}
\renewcommand{\thetable}{S\arabic{table}}
\setcounter{figure}{0}
\setcounter{table}{0}

\startcontents[supplementary]
\section*{Contents}
\printcontents[supplementary]{}{1}[3]{}
\clearpage

\clearpage
\section*{Supplementary Figures}
\addcontentsline{toc}{section}{Supplementary Figures}

\noindent\begin{minipage}{\linewidth} \centering \includegraphics[width=\linewidth,height=.60\textheight,keepaspectratio]{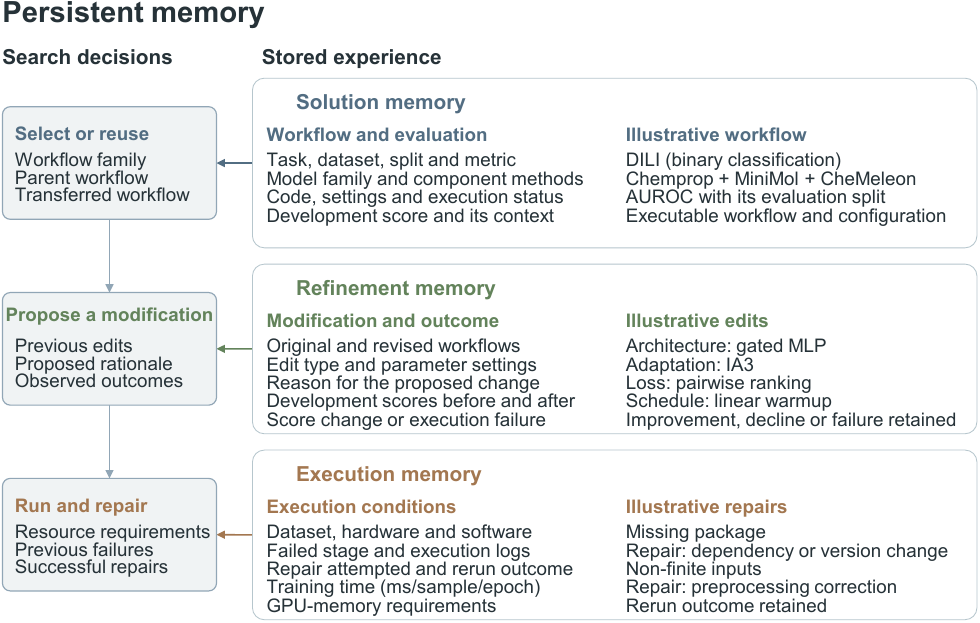} \captionof{figure}{\textbf{Contents and uses of persistent memory.} Solution memory links a workflow's code, configuration and development performance to its task and evaluation conditions. This information supports workflow-family selection, parent selection and workflow transfer. Refinement memory links original and revised workflows to the proposed changes, their rationale and observed outcomes. Previous modifications, including unsuccessful ones, inform subsequent proposals. Execution memory links runtime measurements and failures to the conditions under which they occurred, the repairs attempted and the results of rerunning the workflow. It supports resource allocation and reuse of successful repairs. Examples illustrate the types of information stored, rather than new performance measurements. The same memories inform later iterations within a task and searches on subsequent tasks, subject to the evaluation protocol. DILI, drug-induced liver injury; AUROC, area under the receiver operating characteristic curve; MLP, multilayer perceptron. IA3 denotes a parameter-efficient fine-tuning method~\citep{liu2022fewshot}.} \label{fig:supp-memory-contents} \end{minipage}

\clearpage

\noindent\begin{minipage}{\linewidth}
\centering
\includegraphics[width=.72\linewidth,height=.66\textheight,keepaspectratio]{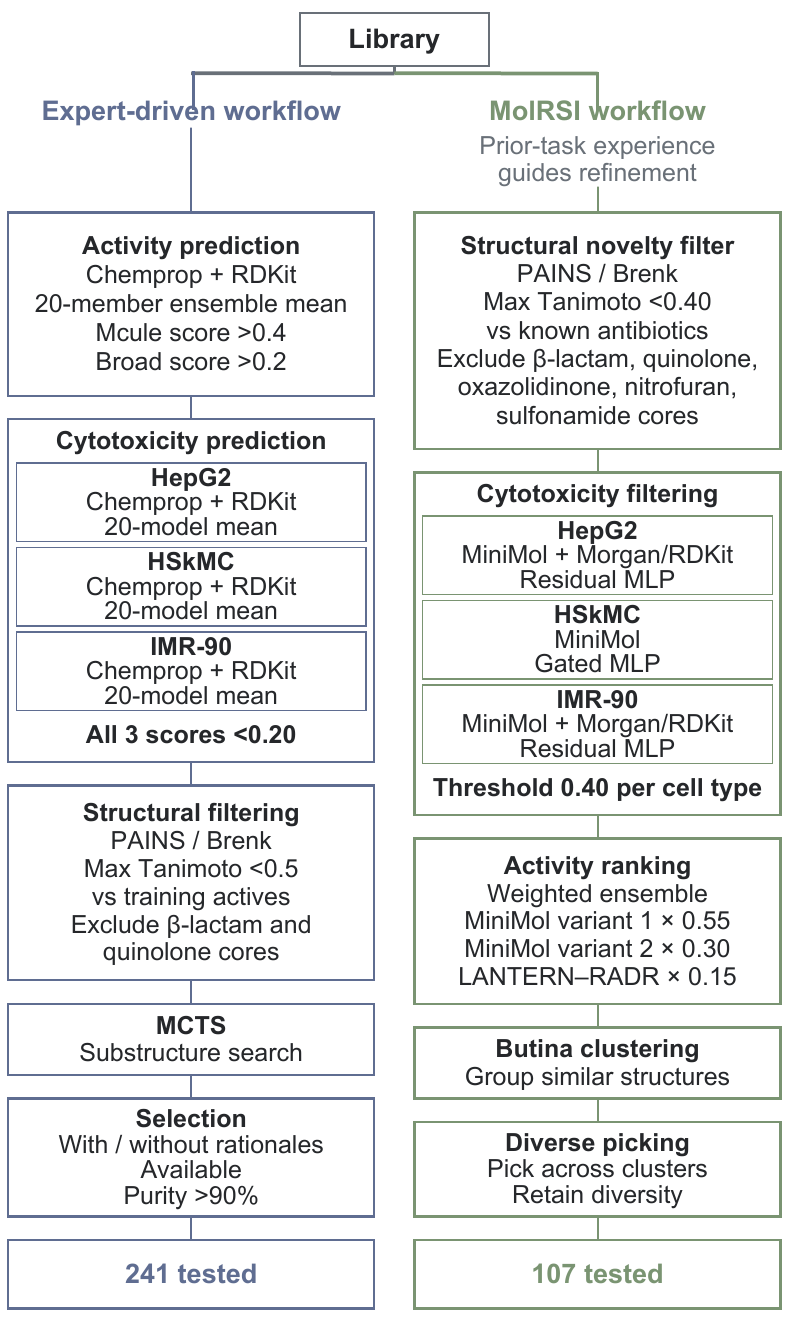}
\captionof{figure}{\textbf{Comparison of the previous expert-driven and \method antibiotic-screening workflows.} Both workflows used the same prospective training datasets and screening library. The expert-driven workflow applied activity prediction before cytotoxicity and structural filtering, and used substructure rationales to inform compound selection. The \method workflow applied structural filtering before cell-type-specific cytotoxicity prediction, antibacterial-activity ranking and selection across structural clusters. The displayed models, thresholds and selection steps are described in \Cref{app:screening-models,app:screening-funnel,app:antibiotic-workflow-comparison}. The campaigns tested 241 expert-selected and 107 \method-selected compounds, respectively. The displayed cytotoxicity thresholds denote nominal criteria; final compound selection included documented exceptions.}
\label{fig:supp-antibiotic-workflow}
\end{minipage}

\clearpage

\noindent\begin{minipage}{\linewidth}
\centering
\includegraphics[width=\linewidth]{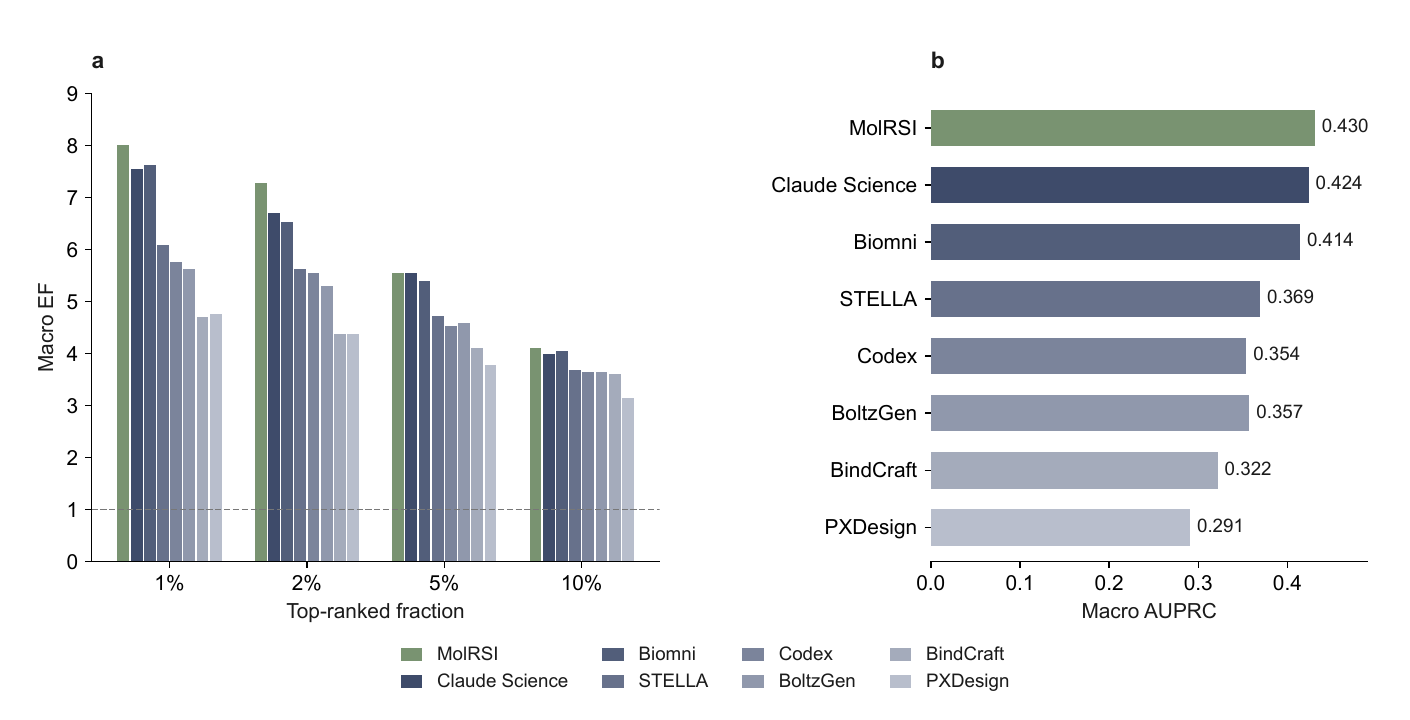}
\captionof{figure}{\textbf{Miniprotein-binder ranking performance across agent workflows and fixed expert-defined objectives.} (a) Macro Enrichment Factor (EF) at four top-ranked fractions; each cutoff is averaged separately across the 14 held-out targets. The dashed line marks EF~=~1. (b) Macro area under the precision--recall curve (AUPRC). Each task used 13 targets for agent workflow search and the remaining target for testing, without cross-task memory. BoltzGen, BindCraft and PXDesign denote the fixed internal ranking criteria described in \Cref{app:protein-expert-comparison}. Bars show point estimates. \Cref{tab:protein-ef-minibinder} provide the target-level numerical results.}
\label{fig:supp-protein-minibinder}
\end{minipage}

\clearpage

\noindent\begin{minipage}{\linewidth}
\centering
\includegraphics[width=\linewidth]{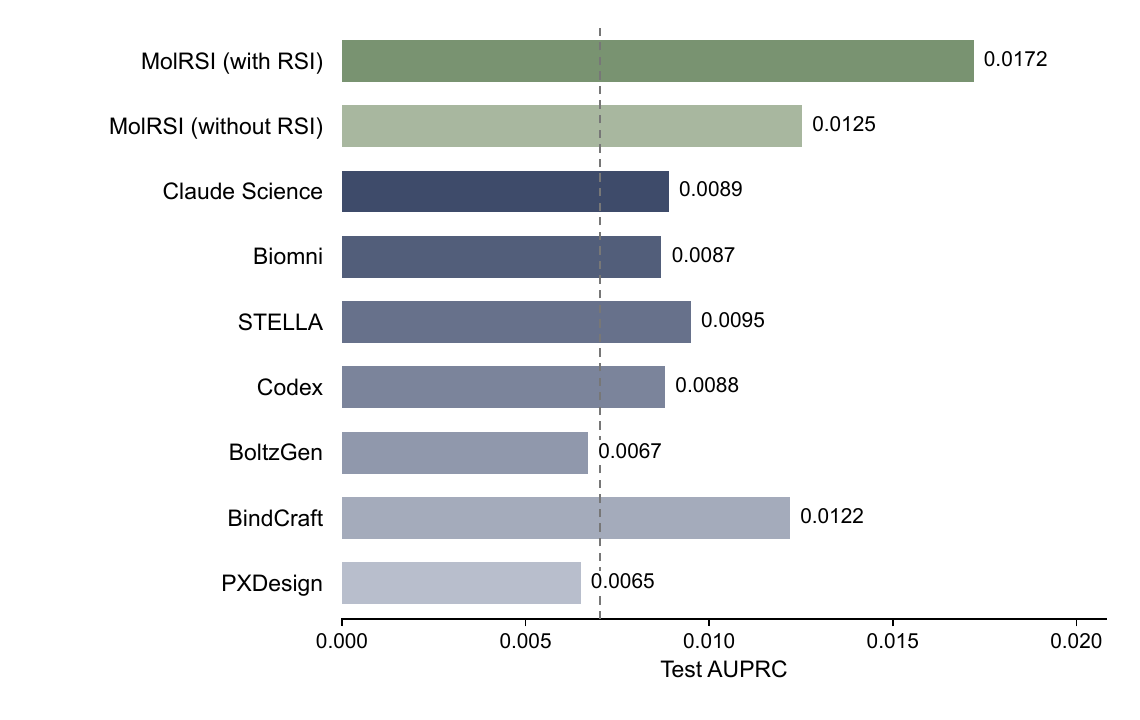}
\captionof{figure}{
\textbf{MART-1 test precision--recall performance and the effect of source-task memory.} AUPRC for \method with and without RSI, the four agent baselines and the three fixed expert-defined ranking objectives. Agent workflows were selected using the two gp100 collections and frozen before evaluation on MART-1. The dashed line indicates the MART-1 positive fraction, 27/3,845. \Cref{tab:protein-ef-pmhc} and \Cref{tab:protein-memory-results} report EF at four cutoffs and AUPRC.
}
\label{fig:supp-protein-pmhc}
\end{minipage}

\clearpage

\section{\method framework}
\label{app:implementation}

\subsection{Resource library construction}
\label{app:explore}

\textbf{Resource retrieval and synthesis.} The Explore Agent generates task-specific queries about scientific methods, software and benchmark evaluation using the task description and available dataset information. It searches literature and web sources, removes duplicate results and ranks the remaining resources by relevance to the task. The retrieved material is summarized to identify models, methods, tools, benchmarks and associated implementations for skill construction.

\textbf{Resource analysis and skill construction.} The Explore Agent examines the retrieved resources and their available code, documentation and usage examples. It prioritizes resources for skill construction according to task relevance and the requested skill categories. When repository code is available, static analysis of Python abstract syntax trees recovers public interfaces and their arguments. These interfaces, together with model definitions and training or inference examples, inform the skill instructions. Additional searches supply supporting repositories or documentation when the initial material is incomplete.

\textbf{Library organization and skill contents.} A shared index lists each skill's name, category and purpose. Each skill describes how to use the resource and links these instructions to available implementations, dependencies, usage examples and source references. Where relevant, it also specifies feature representations and whether computed features could be reused. The contents are summarized in \Cref{tab:resource-library-files}.

\textbf{Checks during construction.} The Explore Agent checks package identity and examines Python usage examples for syntax and import-name inconsistencies. When an execution environment is available, it also checks dependency availability and selected interfaces. Detected problems prompt revisions to installation instructions, dependencies or usage examples. These checks precede execution validation when a skill is first used.

\textbf{Execution validation and reuse.} When a newly constructed skill is first used, \method runs a usage example in an isolated environment. Execution errors guide repairs to the code, dependencies or environment settings. The example is rerun until it completes successfully or reaches the repair-attempt limit. Skills that pass are tagged as validated and retained in the resource library for reuse. Skills that remain unsuccessful are removed.

\textbf{Incremental expansion.} The Explore Agent could expand an existing resource library by retrieving additional resources and identifying skills not already present. Existing skills are retained. New skills undergo the same construction checks and first-use execution validation, and the shared index is updated to include them.

\begin{table}[t]
\centering
\caption{Contents of the resource library. The shared index identifies available skills. Each skill combines instructions with the implementation, source information and validation results available for that resource. The supporting material depends on the resource.}
\label{tab:resource-library-files}
\small
\renewcommand{\arraystretch}{1.18}
\begin{tabularx}{\linewidth}{@{}>{\raggedright\arraybackslash}p{0.40\linewidth}>{\raggedright\arraybackslash}X@{}}
\toprule
\textbf{Component} & \textbf{Information provided} \\
\midrule
Shared index & Skill names, categories, purposes and links to the corresponding instructions. \\
\addlinespace
Skill instructions & Resource description, intended use, installation, supported operations and usage examples. \\
Implementation & Available source code, model definitions, dependencies and supporting scripts. \\
Source information & Scientific references, repository location, revision and package information. \\
Interfaces and examples & Available operations and examples of feature extraction, model fitting or inference, as appropriate to the resource. \\
Validation information & Validation status, checks performed and their outcomes. \\
\bottomrule
\end{tabularx}
\end{table}

\subsection{Persistent memory}
\label{app:memory-schema}

Workflow development generates experience about candidate performance, proposed modifications and execution. \method retains this experience in solution, refinement and execution memory, together with the tasks and conditions under which it was obtained. These memories guide later search decisions within and across tasks (\Cref{fig:supp-memory-contents}).

\textbf{Solution memory.} For each evaluated workflow, solution memory stores the task description, task type, dataset and dataset size, together with the workflow's executable code, configuration and execution status. The configuration specifies the workflow family, component models and methods, architecture, hyperparameters, data processing, training objective and training or generation procedure. Development scores are retained with their evaluation metric and, where applicable, data split and random seed. A workflow that failed before evaluation retains its failure status without receiving a development score. Scores from completed workflows provide evidence about which families performed well on related tasks and which individual workflows are suitable for further refinement or transfer. The historical score normalization, family selection and parent selection procedures are specified in \Cref{app:search-routing}.

\textbf{Refinement memory.} Refinement memory stores the original workflow, the revised workflow, the task and the proposed changes. Changes are described by the affected component and its settings, such as an architecture change, the use of parameter-efficient fine-tuning, a different loss function, altered data processing or a change in learning rate. The agent's reason for proposing a change is retained alongside the result of testing it. When the original and revised workflows are evaluated under the same protocol, both development scores and their difference are stored. Improvements, declines, unchanged scores and execution failures are all retained. If a proposal changes several components, its outcome is associated with the combined modification. For a subsequent proposal, the agent retrieves relevant earlier changes and their outcomes, including previous attempts derived from the selected parent workflow.

\textbf{Execution memory.} Execution memory stores the execution history of workflows and the usable skills in the resource library. For each attempt, it retains the dataset and its size, hardware and software environment, execution status and available runtime measurements. Resource measurements include elapsed time, GPU time, processing speed and GPU memory use. Where measured, training time per sample per epoch relates runtime to dataset size and training duration. These measurements inform later resource allocation and timeout settings. For each failure, the memory retains the failed stage, error messages, execution logs and the cause when identified. It also stores each attempted repair, including changes to code, dependencies or environment settings, and whether re-execution succeeded. This experience guides repairs when similar errors recur.

\textbf{Updates within and across tasks.} Memory is updated after each execution attempt and used to guide subsequent iterations. Experience retained from completed tasks could also guide later searches or direct workflow transfer. When an evaluation used fixed source-task experience, each target began with the same experience, and updates from one target were not shared with other targets. Application-specific experience pools are described in \Cref{app:antibiotic-search,app:protein-datasets}. Held-out test scores are excluded from memory updates and search decisions. \Cref{app:search-routing} describes how the updated memory guides the search policy.

\subsection{Recursive self-improvement of the search policy}
\label{app:search-routing}

\textbf{Search representation.}\label{app:task-interface} \method uses experience-guided Monte Carlo tree search over a forest of executable workflows. Initial workflows form the roots of separate trees, and each node represents a concrete workflow with its specification, execution status, development scores, resource use, logs and generated artifacts. A workflow selected for modification is the parent, and the revised workflow is its child. An edge represents the proposed changes, such as a different representation, architecture, objective, data-processing procedure or composition of methods. Workflows based on a common model or computational approach are grouped into families. The five steps below specify how current-task observations and historical experience guide search decisions.

\textbf{Step 1: Selecting workflow families.}\label{app:transfer-normalization} Family selection combines the best available current-task development performance, an exploration bonus and a historical transfer score. Historical workflow scores are first standardized within each source task and then aggregated using task-similarity weights. The family-level score enters selection as an additive term. A corresponding workflow-level historical score enters parent sampling in Step 2 as a multiplicative factor. The normalization and aggregation used by both steps are defined together below.

\textbf{Historical score normalization.}\label{app:standardized-score} Let \(x_\tau(v)\) denote the development score of workflow \(v\) on source task \(\tau\), and let \(\mathcal{V}_\tau\) contain completed workflows with finite development scores on that task. Scores are first oriented so that larger values indicate better performance:
\begin{equation}
s_\tau(v) =
\begin{cases}
x_\tau(v), & \text{for a higher-is-better metric},\\[3pt]
-x_\tau(v), & \text{for a lower-is-better metric}.
\end{cases}
\label{eq:direction-align}
\end{equation}

Each score is then normalized relative to the other completed workflows on the same source task:
\begin{equation}
\widetilde{s}_\tau(v) =
\frac{s_\tau(v)-\operatorname{median}_{u\in\mathcal{V}_\tau}s_\tau(u)}
{\max\!\left\{\operatorname{MAD}_{u\in\mathcal{V}_\tau}(s_\tau(u)),\epsilon\right\}}.
\label{eq:robust-norm}
\end{equation}

Here, \(\operatorname{MAD}(s)=\operatorname{median}|s-\operatorname{median}(s)|\) is the median absolute deviation from the median, and \(\epsilon>0\) prevents division by zero. The normalized score is mapped to a bounded utility:
\begin{equation}
\bar{s}_\tau(v) = 2\sigma\!\left(\widetilde{s}_\tau(v)\right)-1,
\qquad
\sigma(x)=\frac{1}{1+e^{-x}}.
\label{eq:squash}
\end{equation}

The resulting utility lies in \([-1,1]\), allowing historical outcomes measured on different scales to be aggregated without an unbounded contribution from a single task. All scores used in this calculation were available during development of the source workflow; held-out test scores are excluded.

\textbf{Task-similarity weights.}\label{app:task-weights} For a target task \(\tau_0\), the contribution of source task \(\tau\) is weighted according to evaluation-metric compatibility, task type, dataset size and task-description similarity:
\begin{equation}
w(\tau,\tau_0) =
w_{\mathrm{metric}}(\tau,\tau_0)\,
w_{\mathrm{type}}(\tau,\tau_0)\,
w_{\mathrm{size}}(\tau,\tau_0)\,
w_{\mathrm{emb}}(\tau,\tau_0).
\label{eq:task-weight}
\end{equation}

The non-negative factors \(w_{\mathrm{metric}}\) and \(w_{\mathrm{type}}\) encode compatibility of evaluation metrics and task types. Dataset-size similarity is calculated as
\begin{equation}
w_{\mathrm{size}}(\tau,\tau_0) =
\exp\!\left[
-\gamma\left|
\log|\mathcal{D}_\tau|-\log|\mathcal{D}_{\tau_0}|
\right|
\right],
\label{eq:task-size-weight}
\end{equation}
where \(|\mathcal{D}_\tau|\) denotes the dataset size associated with task \(\tau\), and \(\gamma>0\) controls the penalty for size differences. Description similarity is calculated from task-description embeddings \(\mathbf{e}_\tau\) and \(\mathbf{e}_{\tau_0}\):
\begin{equation}
w_{\mathrm{emb}}(\tau,\tau_0) =
\max\!\left\{
\cos(\mathbf{e}_\tau,\mathbf{e}_{\tau_0}),0
\right\}.
\label{eq:task-description-weight}
\end{equation}

\textbf{Family and workflow transfer scores.}\label{app:transfer-scores} Let \(\mathcal{T}_f\) denote the eligible source tasks containing an evaluated workflow from family \(f\), and let \(\mathcal{V}_{\tau,f}\) denote the evaluated workflows from that family on source task \(\tau\). The family-level transfer score is
\begin{equation}
\operatorname{transfer}(f) =
\frac{
\displaystyle\sum_{\tau\in\mathcal{T}_f}
w(\tau,\tau_0)\,
\operatorname{mean}_{v\in\mathcal{V}_{\tau,f}}\bar{s}_\tau(v)
}{
\displaystyle\sum_{\tau\in\mathcal{T}_f}w(\tau,\tau_0)
}.
\label{eq:transfer-family}
\end{equation}

This calculation first averages performance within each source task and then combines those task-level averages. For a target-task workflow \(v_i\), let \(\mathcal{T}_i\) denote source tasks containing an eligible historical match. The matching rule uses workflow family and modification type, and \(\operatorname{matched}(v_i,\tau)\) denotes the historical workflow identified on source task \(\tau\). The workflow-level transfer score is
\begin{equation}
\operatorname{transfer}(v_i) =
\frac{
\displaystyle\sum_{\tau\in\mathcal{T}_i}
w(\tau,\tau_0)\,
\bar{s}_\tau\!\left(\operatorname{matched}(v_i,\tau)\right)
}{
\displaystyle\sum_{\tau\in\mathcal{T}_i}w(\tau,\tau_0)
}.
\label{eq:transfer-node}
\end{equation}

Tasks without an eligible record are excluded from the corresponding numerator and denominator. These weighted averages are defined when the total task weight is positive. With non-negative task weights and utilities in \([-1,1]\), each defined transfer score also lies in \([-1,1]\).

\textbf{Family-selection rule.} Let \(b\) index family-selection decisions, \(\mathcal{F}_{\mathrm{eligible}}\) denote the eligible workflow families and \(n_b(f)\geq 1\) denote the visit count for family \(f\). The next family is selected according to
\begin{equation}
f_b =
\operatorname*{arg\,max}_{f\in\mathcal{F}_{\mathrm{eligible}}}
\Biggl[
\underbrace{\operatorname{exploit}_b(f)}_{\substack{\text{current-task}\\\text{performance}}}
+\underbrace{\alpha\sqrt{\frac{\ln(b+1)}{n_b(f)}}}_{\substack{\text{exploration}\\\text{bonus}}}
+\underbrace{\operatorname{transfer}(f)}_{\substack{\text{historical}\\\text{transfer}}}
\Biggr].
\label{eq:family-select}
\end{equation}

The exploitation term summarizes the best current-task development performance within family \(f\), including stability and overfitting adjustments, on a scale from zero to one. The coefficient \(\alpha\) controls exploration, with larger bonuses for less-visited families~\citep{auer2002finite}. Historical evidence enters through the family-level transfer score defined above. Historical experience guides initial exploration, and current-task scores are incorporated as evaluations become available.

\textbf{Step 2: Selecting parent workflows.} Let \(\mathcal{V}\) contain completed, non-ensemble workflows with finite development scores on the current task. For workflow \(v_i\in\mathcal{V}\), let \(q_i\) denote the current-task score supplied to the sampler and \(c_i\) the number of existing children. Sampling weights and probabilities are
\begin{equation}
\begin{aligned}
w_i &=
\sigma\!\left(
\beta\frac{q_i-\operatorname{median}_{v_j\in\mathcal{V}}q_j}
{\max\!\left\{\operatorname{MAD}_{v_j\in\mathcal{V}}(q_j),\epsilon\right\}}
\right)
\frac{1}{1+c_i}
\left[1+\lambda\,\operatorname{transfer}(v_i)\right],\\[6pt]
p(v_i) &= \frac{w_i}{\displaystyle\sum_{v_j\in\mathcal{V}}w_j}.
\end{aligned}
\label{eq:node-select}
\end{equation}

The coefficient \(\beta\) controls sensitivity to standardized current-task scores. The factor \(1/(1+c_i)\) reduces repeated expansion of the same parent, and \(\lambda\) controls the contribution of historical transfer. The normalized sampler requires non-negative weights and a positive total weight.

\textbf{Step 3: Proposing workflow modifications.} The agent retrieves previous modifications, their rationales and their outcomes from refinement memory to guide changes to the selected parent workflow. Each proposal specifies the changes and their rationale. The proposed workflow is checked for the required format, duplication of an existing candidate and feasibility of execution. Malformed, duplicate or infeasible proposals do not become completed workflow nodes. The proposed modification and its observed outcome are added to refinement memory after execution, including changes that reduced performance or failed to execute.

\textbf{Step 4: Executing and evaluating workflows.} The proposed child workflow is implemented and run in an isolated environment using the task-specific evaluation procedure. Execution memory informs environment settings and resource allocation. If setup or execution fails, the agent first searches execution memory for similar errors and successful repairs. Relevant experience guides changes to the code, dependencies or environment settings. If no suitable repair is available or the problem persists, the agent diagnoses the current failure and proposes further changes. The workflow is rerun after each repair, within the repair-attempt limit and computational budget. Successful evaluations supply development scores; all attempts retain their execution status, logs, resource measurements and repair outcomes. Failed evaluations count toward the workflow-evaluation allowance, but repairs of the same candidate do not consume additional evaluations.

\textbf{Step 5: Updating search policy to guide subsequent searches.} After each attempt, \method adds the workflow and its development score or failure status to solution memory. It stores the proposed modification and its outcome in refinement memory, and runtime, resource use, errors and repair outcomes in execution memory. The corresponding node and edge in the search forest are also updated. The updated memories guide subsequent family selection, parent sampling, modification proposals and execution. These updates change the effective search policy \(\pi_t\), whose subsequent evaluations generate experience for the next memory update. This feedback implements the RSI loop expressed by \(M_t\), \(E_t\) and \(\pi_t\) in \Cref{eq:memory-conditioned-policy}.

\textbf{Historical and current task updates.}\label{app:proof} Current task executions update development scores, visit counts and the experience used for subsequent refinement and execution. When the source memory was fixed for an evaluation, its historical transfer scores remained unchanged during that target-task search. Under continued use across tasks, experience from newly completed tasks could contribute to transfer estimates for later searches. The bounded transfer scores serve as empirical selection priors; their boundedness alone does not establish a standard target-task or node-level regret guarantee.

\textbf{Search termination and returned workflow.} Search stops when the task-specific limit on evaluation attempts or computation time is reached. \method returns the successfully evaluated workflow with the best current-task development score. This is the best evaluated workflow under the stated criterion and budget, rather than a global optimum. The evaluation unit and accounting for failed attempts and repairs are specified in \Cref{app:baselines}.

\textbf{Workflow transfer with and without further search.} \method could use experience from previous tasks to retrieve a workflow for direct transfer or initialize a new search. It first identifies the best matching previously completed task using task type, evaluation metric, dataset size and task description similarity. It then retrieves that source task's workflow with the best development score from solution memory. This source task routing is distinct from the family- and workflow-level transfer scores used in Steps 1 and 2. In zero-search mode, the retrieved configuration is applied to the new task without proposing or comparing alternative workflows. Model parameters are fitted using the new task's training data where required, and fitting and execution still incur computational cost. Alternatively, the retrieved workflow becomes a starting point for Steps 1--5 under the new task's limits on evaluation attempts and computation time.

\section{Comparison with baseline agents}
\label{app:baselines}

\subsection{Shared evaluation protocol}
\label{app:baseline-protocol}

\textbf{Agents and comparison settings.} We compared \method with Biomni~\citep{Huang2026BiomniScience}, Claude Science~\citep{anthropic_claude_science}, STELLA~\citep{jin2025stella} and Codex~\citep{openai_codex} on antibacterial prediction and protein binder ranking workflow development. The antibacterial benchmark additionally included Claude Code~\citep{anthropic_claude_code}. Within each benchmark, agents received the same task inputs, permitted development data, evaluation objective and workflow-evaluation allowance. Each agent used its native tools and workflow development procedures without a common predefined workflow search space. The task-specific instructions and permitted tools are described below.

\textbf{Task instructions and permitted tools.} For baseline searches in antibacterial prediction and protein ranking, prompts specified the task inputs, fixed data partitions, development objective and metric, evaluation allowance, required outputs and permitted tools. No starting prediction model or ranking workflow was prescribed. The instructions prohibited changes to the fixed data, partitions, and data-loading or evaluation code, hidden candidate evaluations, and the use of held-out labels or performance for workflow development or selection. Independent baseline runs also prohibited reuse of results, code or model choices from another baseline run.

\textbf{Search permissions and execution records.} These baseline agents could search using their native capabilities, and local agents could install software packages. Search, installation, environment inspection and deterministic feature caching independent of the labels did not consume candidate evaluations. Candidates were evaluated sequentially, with subsequent choices based on the preceding development results. Each candidate's code, outputs, execution status, logs, runtime, development scores and held-out test predictions were retained.

\textbf{Evaluation allowance and repair.} A candidate counted as one workflow evaluation once its evaluation began, including evaluations that failed or timed out. Repair attempts for the same candidate did not count as additional workflow evaluations. The authors supplied the task inputs and evaluation interfaces without selecting workflow components, repairing candidates or granting additional retries during the runs.

\textbf{LLM backbones.} \method used Claude Sonnet~5 throughout the antibacterial and protein studies. In the antibacterial prediction and protein ranking benchmarks, the baseline agents also used Claude Sonnet~5, except Codex, which used GPT-5.5. Claude Science used Claude Opus~5 for prospective binder design.

\textbf{API access and agent interfaces.} \method accessed its model through the Anthropic API. Biomni ran on its own platform and used its platform-provided environment for workflow planning and execution. Its outputs were not independently reproduced outside that platform. Claude Science used its scientific workflow-development interface. STELLA ran in self-evolve mode using the implementation available at \url{https://github.com/asfarasimconcerned/STELLA}. Codex and Claude Code~\citep{anthropic_claude_code} used their native coding-agent interfaces, with writable workflow directories and fixed data and evaluation helpers.

\textbf{Computational resources.} Across antibacterial prediction, protein ranking and prospective binder design, the authors provided GPUs to agent baselines, and each agent controlled job scheduling and GPU use within its task-specific budget. 
Their prompts imposed no hard wall-clock limit. These hardware assignments describe the four local baselines and do not specify the resources provided by the Biomni platform.

\subsection{Antibacterial task instructions and permitted tools}
\label{app:antibiotic-baselines}

\textbf{Task inputs and evaluation.} We evaluated \method with and without RSI alongside the baseline agents on the six antibacterial target tasks. Systems received the same molecular inputs, fixed partitions, development objective and allowance of 20 attempted workflow evaluations per task. A completed candidate evaluation covered all five fixed splits in \Cref{app:partition}, rather than counting each split as a separate attempt. Mean validation AUPRC guided search and final workflow selection; test predictions were scored after search.

\textbf{Search permissions and execution records.} Antibacterial baselines followed the shared search permissions and execution-record requirements in \Cref{app:baseline-protocol}, using validation scores as their development feedback.

\textbf{Computational resources.} Claude Code, Claude Science, STELLA and Codex were each assigned one NVIDIA L40S GPU and 20 CPU threads. Scheduling followed the shared protocol in \Cref{app:baseline-protocol}.

\textbf{Complete task prompt and substitutions.}\label{app:antibiotic-task-prompt} The Claude Code prompt for the \textit{E. coli} task is reproduced below, with machine-specific paths anonymized and device indices omitted. Path prefixes are represented by \texttt{<BASELINE\_ROOT>}; the remaining instructions are unchanged. Biomni received the same task instructions through its native interface. Dataset names, identifiers, split sizes and loading checks were adjusted for each task (\Cref{tab:antibiotic-prompt-datasets}).

\noindent\begin{minipage}{\linewidth}
\centering
\small
\captionof{table}{Dataset-specific fields in the antibacterial baseline prompts. The dataset identifier was formed by prefixing the task name with \texttt{antibiotic\_\_}. Names retain the spelling used in the input files. Train/validation pool sizes combine the training and validation partitions; ranges reflect the five fixed splits.}
\label{tab:antibiotic-prompt-datasets}
\begin{tabularx}{\textwidth}{@{}lXrr@{}}
\toprule
\textbf{Dataset} & \textbf{Task name} & \textbf{Train/validation pool} & \textbf{Test set} \\
\midrule
\textit{S. aureus} & \texttt{activity\_S\_aureus} & 35{,}451 & 3{,}939 \\
\textit{P. aeruginosa} & \texttt{activity\_P\_aeurignosa} & 35{,}340--35{,}341 & 3{,}926--3{,}927 \\
\textit{A. baumannii} & \texttt{activity\_A\_baumanii} & 35{,}341 & 3{,}926 \\
\textit{E. coli} & \texttt{activity\_E\_coli} & 35{,}340--35{,}341 & 3{,}926--3{,}927 \\
\textit{K. pneumoniae} & \texttt{activity\_K\_pneumoniae} & 35{,}340--35{,}341 & 3{,}926--3{,}927 \\
\textit{N. gonorrhoeae} & \texttt{activity\_N\_gonorrhea} & 34{,}785 & 3{,}865 \\
\bottomrule
\end{tabularx}
\end{minipage}
\par\vspace{1em}

\begingroup
\lstset{literate={—}{{\textemdash}}1,showstringspaces=false}
\begin{promptbox}{Complete antibacterial task prompt: Claude Code, E. coli}
You are participating in the antibiotic activity benchmark. Your job is to find the best
model you can for one task, under a fixed protocol.

BENCHMARK INSTRUCTIONS
------
- The task-specific description is provided below. Follow it exactly.
- Work in `<BASELINE_ROOT>/claude_code/activity_E_coli`. The dataset is at `<BASELINE_ROOT>/claude_code/activity_E_coli/dataset_activity_E_coli/`.
- The fixed helper `prepare.py` is at `<BASELINE_ROOT>/claude_code/activity_E_coli/prepare.py`. Import and use it. Do not modify it:

      import sys; sys.path.insert(0, "<BASELINE_ROOT>/claude_code/activity_E_coli")
      from prepare import (load_seed_split, load_seed_test, evaluate,
                           save_predictions, record_step, SEEDS, METRIC,
                           budget_remaining)

      load_seed_split(seed)  -> (train_df, val_df)   both carry Y
      load_seed_test(seed)   -> test_df              carries Y, for scoring only
      evaluate(y_true, y_pred) -> float              uses this task's metric
      save_predictions(step, seed, test_df, y_pred)  -> submission/step_NN/
      record_step(step, val_scores, description)     closes one candidate
      budget_remaining() -> int

- You must train and evaluate on all five fixed seed splits `1, 2, 3, 4, 5`.
- For every candidate, number it (1, 2, 3, ...), call `save_predictions(step, seed, ...)`
  for each of the five seeds, then call `record_step(step, val_scores, description)` once.
  Nothing is overwritten between steps, so every candidate stays scoreable afterwards.
- For each seed print exactly:  [seed N] val_auprc=VALUE
- Your final message must contain the result line in the exact format:
  `[result] AUPRC = MEAN +/- STD`
  reporting the VALIDATION mean and standard deviation of your selected candidate,
  and must state which step number you selected.
- The mean validation metric across the 5 seed splits is the sole optimization and
  model-selection target.
- `test.csv` carries `Y` so that predictions can be scored offline after this run.
  Never read it, train on it, or compute a test score during the search — model
  selection is validation-only.
- Do not modify dataset files.

ADDITIONAL NOTES
------
- Compute: NVIDIA L40S GPU and 20 CPU threads. `prepare.py` pins both on import;
  do not override them.
- Host folder access: if you need it, call `request_host_access` for exactly
  `<BASELINE_ROOT>/claude_code`
  with mode `rw`. That one grant covers this task's dataset, prepare.py and submission
  directory, so do not request any narrower or additional path.
- Total runtime: there is no hard wall-clock limit. Your budget is the experiment count
  below, so spend it on better ideas rather than rushing.
- Total steps: at most 20 candidate experiments. This is enforced, not advisory:
  `record_step` raises once the 20th is recorded. Call `budget_remaining()` to check,
  and plan so your best configuration is measured before the budget runs out.
- No starting method is prescribed. Deciding what to try is part of the task.
- Live web search is available in this Claude Code session. You may search online for the
  models and methods that actually work for this kind of endpoint — recent molecular
  representation models, pretrained embeddings, domain-specific architectures — before
  committing budget to a candidate. You may install and use any package or published
  method you judge promising. Searching and installing do not consume experiment budget;
  only a recorded candidate does. This is optional, not required.
- Use the budget to iterate and improve, but keep each candidate reasonably efficient.
- Work one candidate at a time. Do not plan or fix the whole sequence in advance: after each
  `record_step`, read that candidate's five validation scores and let what you just learned
  decide the next candidate. Each candidate should be a focused change motivated by the
  results so far, not the next item on a pre-made list.

TASK DESCRIPTION
------
# Molecular Property Prediction Benchmark — activity_E_coli

## Task

Predict the molecular property endpoint **activity_E_coli** from molecular SMILES strings.

- Dataset id: `antibiotic__activity_E_coli`
- Dataset family: antibiotic
- Category: activity
- Task type: classification
- Primary metric: **AUPRC** (higher is better)
- Train/validation pool: 35340-35341 molecules depending on seed
- Held-out test set: 3926-3927 molecules depending on seed; you must produce per-seed test predictions for every candidate
- Split: {"fingerprint": {"nbits": 2048, "radius": 2, "type": "Morgan"}, "method": "greedy_leader_tanimoto", "similarity_threshold": 0.4, "stratified_by": "Y", "test_fraction": 0.1, "val_fraction": 0.1}

## Data

The uploaded `dataset_activity_E_coli/` directory belongs only to **activity_E_coli**. It contains `meta.json` and five fixed split directories, `seed_1/` through `seed_5/`. Before running any experiment, verify that `dataset_activity_E_coli/meta.json` has `name` equal to `activity_E_coli`, `dataset_id` equal to `antibiotic__activity_E_coli`, and primary metric equal to `AUPRC`; stop and report a dataset mismatch if any value differs. Each seed directory contains `train.csv`, `val.csv`, and `test.csv`. The labeled training and validation files contain `Drug_ID`, `Drug` (SMILES), `Y` (label/target), and may contain precomputed molecule-derived descriptor columns. The test files contain the same columns, including `Y`; `Y` is present only so that predictions can be aligned and scored afterwards, and must never be read during training, tuning, or model selection.

Use the fixed split files directly:

```python
import pandas as pd

SEEDS = [1, 2, 3, 4, 5]

for seed in SEEDS:
    train_df = pd.read_csv(f"dataset_activity_E_coli/seed_{seed}/train.csv")
    val_df = pd.read_csv(f"dataset_activity_E_coli/seed_{seed}/val.csv")
```

Model selection is validation-only: never let test labels or test performance influence training, tuning, search, or candidate choice. For every candidate you evaluate, do generate and save per-seed test predictions via `prepare.save_predictions` — they are scored afterwards, outside this run. Do not create alternative train/validation splits and do not merge train and validation during search.

## Evaluation

The primary validation metric is **AUPRC** (higher is better). Evaluate each candidate independently on all five fixed seed splits. Compute the metric separately for each seed, then report the arithmetic mean and sample standard deviation (`ddof=1`) across the five values. Use `roc_auc_score` for AUROC, `average_precision_score` for AUPRC, `mean_absolute_error` for MAE, and `spearmanr(...).statistic` for Spearman. Classification metrics must be computed from positive-class probabilities, not hard class labels.

For every candidate, print exactly one validation line for each seed followed by the aggregate result:

```text
[seed 1] val_auprc=VALUE
[seed 2] val_auprc=VALUE
[seed 3] val_auprc=VALUE
[seed 4] val_auprc=VALUE
[seed 5] val_auprc=VALUE
[result] AUPRC = MEAN +/- STD
```

All five seed lines and the final result line are mandatory. The five-seed validation mean is the sole optimization and model-selection target.

## Experiment Budget

Run exactly 20 complete candidate experiments, named `step_01` through `step_20`. One budget unit is one candidate method trained and evaluated on all five fixed seed splits. Evaluate one candidate at a time. A candidate whose evaluation starts but crashes or times out still consumes its numbered step. Environment inspection and deterministic label-independent feature caching do not consume a step, but unlogged pilot experiments, hidden parameter sweeps, and extra candidate evaluations are not allowed.

After each step, record the candidate description, complete source code, command, stdout/stderr, runtime, status, five validation values, validation mean, and sample standard deviation. Preserve every step's artifacts without overwriting earlier steps. After `step_20`, select the completed candidate with the best five-seed validation mean according to the stated metric direction and preserve its source code as the final solution.

## Constraints

- Do not modify the dataset files, labels, fixed splits, primary metric, or evaluation definition.
- Do not use test labels, test predictions, or test performance for training, tuning, search, or model selection.
- Keep the five-seed protocol intact; every candidate must run on seeds `[1, 2, 3, 4, 5]`.
- Do not add extra search steps.


## Final Response

At the end, report the selected step, its five per-seed validation values, validation mean +/- sample standard deviation, the number of completed steps out of 20, failed steps, and the paths to the selected source code and complete step logs.
\end{promptbox}
\endgroup

\subsection{Protein task instructions and permitted tools}
\label{app:protein-baselines}

\textbf{Task inputs and evaluation.}

\textit{Ranking workflow search.} Claude Science, Biomni, STELLA and Codex received target structures, candidate collections, permitted development labels and a task-specific ranking objective. Each baseline was asked to return a workflow assigning a scalar score to each target--binder complex. AUPRC on the 13 search targets guided workflow selection for each miniprotein-binder task. AUROC on the two gp100 collections guided selection for the pMHC task. Selected workflows were fixed before held-out evaluation using the partitions in \Cref{app:protein-datasets}.

\textit{Prospective design workflow.} The recorded Claude Science protocol used an independent session per target. The recorded instructions supplied HLA heavy-chain and \(\beta_2\)-microglobulin sequences and MSAs, on- and off-target peptide sequences, an 80--120-residue binder range, epitope hints and novelty references. Target coordinates were not supplied. The requested output was 500 ranked designs, using the prescribed on- and off-target scoring criterion and diversity constraints described in \Cref{app:claude-prospective}.

\textbf{Search permissions and execution records.}

\textit{Ranking workflow search.} All four protein ranking baselines followed the same search permissions and execution-record requirements as the antibacterial baselines (\Cref{app:baseline-protocol}), using scores on the permitted search targets as their development feedback.

\textit{Prospective design workflow.} Claude Science accessed and installed tools under its own platform rules. It chose generator settings, additional sampling and refinement steps within the task's sampling, scoring and diversity requirements. Following Anthropic's published prompt (source below), the agent was asked to save design sequences and structures, records of how each design was generated and refined, scores, run configurations and logs. Failed checks and runs were also to be recorded (\Cref{app:claude-prospective}).

\textbf{Computational resources.} Both ranking and prospective design followed the shared scheduling and wall-clock policy in \Cref{app:baseline-protocol}.

\textit{Ranking workflow search.} Claude Science, STELLA and Codex each received 300 A100 GPU-hours and 20 attempted workflow evaluations per benchmark task. Folded structures and model scores were cached and reused across evaluations. 

\textit{Prospective design workflow.} Claude Science received a budget of 1,000 A100 GPU-hours per target, matching \method. The agent determined how to schedule jobs within this budget.

\textbf{Task prompts and substitutions.}

\textit{Ranking workflow search.}\label{app:protein-task-prompt} The Codex prompt for the pMHC task is reproduced below, with machine-specific paths anonymized. Path prefixes are represented by \texttt{<BASELINE\_ROOT>}; the remaining instructions are unchanged. Claude Science, Biomni and STELLA received the same task instructions through their native interfaces. The 14 miniprotein-binder ranking tasks used the same prompt with the candidate collections, the leave-one-target-out partition of \Cref{tab:minibinder-target-data} and AUPRC as the selection metric substituted for those tasks.

\begingroup
\lstset{literate={—}{{\textemdash}}1,showstringspaces=false}
\begin{promptbox}{Complete pMHC ranking task prompt: Codex, pmhc}
You are participating in the protein binder ranking benchmark. Your job is to find the best
ranking workflow you can for one task, under a fixed protocol.

BENCHMARK INSTRUCTIONS
------
- The task-specific description is provided below. Follow it exactly.
- Work in `<BASELINE_ROOT>/codex/pmhc_binder_ranking`. The dataset is at `<BASELINE_ROOT>/codex/pmhc_binder_ranking/dataset_pmhc_binder_ranking/`.
- The fixed helper `prepare.py` is at `<BASELINE_ROOT>/codex/pmhc_binder_ranking/prepare.py`. Import and use it. Do not modify it:

      import sys; sys.path.insert(0, "<BASELINE_ROOT>/codex/pmhc_binder_ranking")
      from prepare import (load_search, load_test, evaluate,
                           save_predictions, record_step, SEARCH, METRIC,
                           budget_remaining)

      load_search(collection) -> df                  carries Y
      load_test()             -> df                  carries Y, for scoring only
      evaluate(y_true, y_score) -> float             uses this task's metric
      save_predictions(step, collection, df, y_score)  -> submission/step_NN/<collection>.csv
                                                     with design_id and score
      record_step(step, search_scores, description)    closes one candidate
      budget_remaining() -> int

- You must score every candidate on both search collections in `SEARCH`.
- For every candidate, number it (1, 2, 3, ...), call `save_predictions(step, collection, ...)`
  for both search collections and for the held-out test collection, then call
  `record_step(step, search_scores, description)` once. Nothing is overwritten between
  steps, so every candidate stays scoreable afterwards.
- For each search collection print exactly:  [COLLECTION] val_auroc=VALUE
- Your final message must contain the result line in the exact format:
  `[result] AUROC = MEAN +/- STD`
  reporting the SEARCH mean and standard deviation of your selected candidate,
  and must state which step number you selected.
- The mean search AUROC over the two gp100 collections is the sole optimization and
  model-selection target.
- The test collection carries `Y` so that predictions can be scored offline after this run.
  Never read it, train on it, or compute a test score during the search — workflow
  selection uses the search collections only.
- Do not modify dataset files.

ADDITIONAL NOTES
------
- Compute: 300 A100 GPU-hours. Jobs are single-GPU and independent; pin
  `CUDA_VISIBLE_DEVICES` on every one of them, because several folding stacks open every
  visible GPU on import and one unpinned job destroys throughput on its whole host.
- Folded structures and model scores may be cached and reused across candidates. Label-free
  caching of this kind does not consume an experiment, so fold once and score many times
  rather than refolding per candidate.
- Workspace: `<BASELINE_ROOT>/codex/pmhc_binder_ranking` is writable and network access is
  enabled, so you can install packages and fetch model weights. Write nothing outside that
  directory, and treat the dataset directory inside it as read-only.
- Total runtime: there is no hard wall-clock limit. Your budget is the experiment count
  below and the GPU-hours above, so spend them on better ideas rather than rushing.
- Total steps: at most 20 candidate experiments. This is enforced, not advisory:
  `record_step` raises once the 20th is recorded. Call `budget_remaining()` to check,
  and plan so your best workflow is measured before the budget runs out.
- No starting workflow is prescribed. Deciding what to try is part of the task.
- Web search is available in this session. You may search online for the
  models and methods that actually work for this kind of endpoint — structure predictors,
  interface-confidence metrics, published ranking criteria, pMHC-specific scoring — before
  committing budget to a candidate. You may install and use any package or published method
  you judge promising. Searching and installing do not consume experiment budget; only a
  recorded candidate does. This is optional, not required.
- Use the budget to iterate and improve, but keep each candidate reasonably efficient.
- Work one candidate at a time. Do not plan or fix the whole sequence in advance: after each
  `record_step`, read that candidate's search scores and let what you just learned decide the
  next candidate. Each candidate should be a focused change motivated by the results so far,
  not the next item on a pre-made list.

TASK DESCRIPTION
------
# Protein Binder Ranking Benchmark — pmhc_binder_ranking

## Task

Rank designed miniprotein binders against peptide-MHC class I complexes: assign a scalar score to each target--binder complex, higher for candidates that were selected in the screen.

- Dataset id: `protein__pmhc_binder_ranking`
- Dataset family: protein binder ranking
- Category: pMHC
- Task type: classification (binary selection outcome)
- Primary metric: **AUROC** (higher is better)
- Search collections: two gp100 HLA-A2 complexes, 2,611 and 3,271 candidates
- Held-out test collection: MART-1 (A2L; ELAGIGILTV), 3,845 candidates; you must produce test predictions for every candidate
- Partition: fixed and given -- both gp100 collections are the search set, MART-1 is the test set

## Data

The uploaded `dataset_pmhc_binder_ranking/` directory belongs only to **pmhc_binder_ranking**. It contains `meta.json` and three collection directories: `gp100_YLEPGPVTA/` and `gp100_IMDQVPFSV/`, which are the search collections, and `mart1_ELAGIGILTV/`, which is the test collection. Before running any experiment, verify that `dataset_pmhc_binder_ranking/meta.json` has `name` equal to `pmhc_binder_ranking`, `dataset_id` equal to `protein__pmhc_binder_ranking` and primary metric equal to `AUROC`, and that the two search collections hold 2,611 candidates with 14 positives and 3,271 candidates with 4 positives respectively; stop and report a dataset mismatch if any value differs.

Each collection directory contains `candidates.csv` and the target specification it was screened against: `target.yaml`, carrying the `hla_allele`, `hla_sequence`, `b2m_sequence` and `peptide` fields; `msa/hla.a3m` and `msa/b2m.a3m`, the frozen target-chain alignments; and `target.cif`, the reference complex structure for that pMHC. `candidates.csv` carries `design_id`, `binder_seq` (102 aa, 20 canonical residues) and `Y`, the binary outcome of the sorting experiment, where `1` means the candidate was selected. The test collection has the same columns, including `Y`; `Y` is present only so that predictions can be aligned and scored afterwards, and must never be read during workflow development, tuning or selection.

Use the fixed collections directly:

```python
import pandas as pd

SEARCH = ["gp100_YLEPGPVTA", "gp100_IMDQVPFSV"]

for collection in SEARCH:
    df = pd.read_csv(f"dataset_pmhc_binder_ranking/{collection}/candidates.csv")
```

Workflow selection is search-only: never let test labels or test performance influence scoring, tuning, search or candidate choice. For every candidate you evaluate, do generate and save test predictions via `prepare.save_predictions` -- they are scored afterwards, outside this run. Do not create alternative partitions, and do not merge the search and test collections during search.

Three properties of this data decide what a result here means. Positives are scarce: 18 across the two search collections, so a handful of candidates moves the metric and a difference of a few points of AUROC is not a reliable ordering of two workflows. The two search collections present different gp100 peptides on the same HLA-A2 scaffold and differ only in the peptide chain, while the test collection presents an unrelated peptide, so a workflow fitted to gp100-specific features will not transfer. The negatives are hard by construction: every candidate in every collection was a design that a generative pipeline already ranked highly enough to be worth ordering, so separation here will not reproduce on a designed-versus-random task.

## Evaluation

The primary metric is **AUROC** (higher is better). Evaluate each candidate independently on both search collections. Compute the metric separately for each collection, then report the arithmetic mean and sample standard deviation (`ddof=1`) across the two values. Use `roc_auc_score` for AUROC and `average_precision_score` for AUPRC, and compute both from continuous scores, not hard class labels. AUPRC and enrichment factor may be recorded as diagnostics, but the search mean AUROC is the selection target.

For every candidate, print exactly one validation line for each search collection followed by the aggregate result:

```text
[gp100_YLEPGPVTA] val_auroc=VALUE
[gp100_IMDQVPFSV] val_auroc=VALUE
[result] AUROC = MEAN +/- STD
```

Both search lines and the final result line are mandatory. The search mean AUROC is the sole optimization and model-selection target.

## Experiment Budget

Run at most 20 complete candidate experiments, named `step_01` through `step_20`. One budget unit is one candidate ranking workflow evaluated on both search collections. Evaluate one candidate at a time. A candidate whose evaluation starts but crashes or times out still consumes its numbered step; repair attempts on the same candidate do not consume an additional step. Environment inspection and deterministic label-independent caching of folded structures and model scores do not consume a step, but unlogged pilot experiments, hidden parameter sweeps and extra candidate evaluations are not allowed.

After each step, record the candidate description, complete source code, command, stdout/stderr, runtime, GPU-hours used, status, both search values, the search mean and the sample standard deviation. Preserve every step's artifacts without overwriting earlier steps. After the final step, select the completed candidate with the best search mean AUROC and preserve its source code as the final solution.

## Constraints

- Do not modify the dataset files, labels, fixed partition, primary metric or evaluation definition.
- Do not use test labels, test predictions or test performance for scoring, tuning, search or workflow selection.
- Keep the partition intact; every candidate must be scored on both search collections and must produce test predictions.
- Do not exceed 300 A100 GPU-hours, and pin `CUDA_VISIBLE_DEVICES` on every job.
- Do not add extra search steps.

## Final Response

At the end, report the selected step, its two per-collection search AUROC values, the search mean +/- sample standard deviation, the number of completed steps out of 20, failed steps, the GPU-hours used out of 300, and the paths to the selected source code and complete step logs.
\end{promptbox}
\endgroup

\textit{Prospective design workflow.} The Claude Science instructions were adapted from Anthropic's published multi-target binder-design prompt~\citep{claudescience2026binder}.\footnote{\url{https://huggingface.co/datasets/Anthropic/claude-protein-binder-design/blob/main/prompts/prompts/multi_target_binder_design_prompt.md}} pMHC-specific modifications and run settings are described in \Cref{app:claude-prospective}.

\section{\method for antibiotic discovery}
\label{app:antibacterial-prediction}

\subsection{Datasets and benchmark evaluation}
\label{app:datasets}

\textbf{Dataset roles.} We used 25 source tasks to accumulate experience for antibiotic workflow development and six antibacterial target tasks to evaluate its transfer. The source tasks comprised 22 absorption, distribution, metabolism, excretion and toxicity (ADMET) tasks from the Therapeutics Data Commons (TDC)~\citep{huang2021therapeutics} and three cytotoxicity tasks for HepG2, HSkMC and IMR-90 cells. The antibacterial targets were activity prediction for \textit{Staphylococcus aureus}, \textit{Pseudomonas aeruginosa}, \textit{Acinetobacter baumannii}, \textit{Escherichia coli}, \textit{Klebsiella pneumoniae} and \textit{Neisseria gonorrhoeae}. The cytotoxicity and antibacterial data were obtained from our previous study~\citep{wong2024discovery}. Experience accumulated on the source tasks was used to initialize searches on the antibacterial targets; outcomes from the six target tasks were excluded from this initial memory. \Cref{tab:tdc-source-tasks} lists the source tasks and their evaluation metrics.

\setlength{\LTcapwidth}{\linewidth}
\small
\begin{longtable}{
>{\raggedright\arraybackslash}p{2.0cm}
>{\raggedright\arraybackslash}p{3.2cm}
>{\raggedright\arraybackslash}p{\dimexpr\linewidth-2.0cm-3.2cm-1.2cm-1.7cm-10\tabcolsep\relax}
>{\raggedleft\arraybackslash}p{1.2cm}
p{1.7cm}
}
\caption{The 25 source tasks used for memory accumulation. The first 22 tasks are from the TDC ADMET benchmark, and the remaining three are mammalian cytotoxicity tasks. TDC sizes are the reported benchmark sizes; these tasks used the official benchmark partitions.}
\label{tab:tdc-source-tasks}\\
\toprule
\textbf{Category} & \textbf{Dataset} & \textbf{Endpoint} & \textbf{Size} & \textbf{Metric}\\
\midrule
\endfirsthead
\toprule
\textbf{Category} & \textbf{Dataset} & \textbf{Endpoint} & \textbf{Size} & \textbf{Metric}\\
\midrule
\endhead
\midrule
\multicolumn{5}{r}{\small\textit{Continued on next page}}\\
\endfoot
\bottomrule
\endlastfoot
Absorption & Caco-2 & Intestinal permeability in a Caco-2 assay & 906 & MAE\\
Absorption & HIA & Human intestinal absorption & 578 & AUROC\\
Absorption & Pgp & P-glycoprotein inhibition & 1{,}212 & AUROC\\
Absorption & Bioavailability & Oral bioavailability & 640 & AUROC\\
Absorption & Lipophilicity & Octanol--water distribution coefficient & 4{,}200 & MAE\\
Absorption & Solubility & Aqueous solubility & 9{,}982 & MAE\\
\addlinespace
Distribution & BBB (Martins) & Blood--brain barrier penetration & 1{,}975 & AUROC\\
Distribution & PPBR & Human plasma-protein binding rate & 1{,}797 & MAE\\
Distribution & VDss & Volume of distribution at steady state & 1{,}130 & Spearman\\
\addlinespace
Metabolism & CYP2C9 substrate & CYP2C9 substrate classification & 666 & AUPRC\\
Metabolism & CYP2D6 substrate & CYP2D6 substrate classification & 664 & AUPRC\\
Metabolism & CYP3A4 substrate & CYP3A4 substrate classification & 667 & AUROC\\
Metabolism & CYP2C9 inhibition & CYP2C9 inhibitor classification & 12{,}092 & AUPRC\\
Metabolism & CYP2D6 inhibition & CYP2D6 inhibitor classification & 13{,}130 & AUPRC\\
Metabolism & CYP3A4 inhibition & CYP3A4 inhibitor classification & 12{,}328 & AUPRC\\
\addlinespace
Excretion & Half life & Drug half-life & 667 & Spearman\\
Excretion & Clearance (hepatocyte) & Hepatocyte intrinsic clearance & 1{,}020 & Spearman\\
Excretion & Clearance (microsome) & Microsomal intrinsic clearance & 1{,}102 & Spearman\\
\addlinespace
Toxicity & hERG & hERG-channel blockade & 648 & AUROC\\
Toxicity & DILI & Drug-induced liver injury & 475 & AUROC\\
Toxicity & Ames & Mutagenicity in the Ames assay & 7{,}255 & AUROC\\
Toxicity & LD50 & Acute oral toxicity & 7{,}385 & MAE\\
\addlinespace
Cytotoxicity & HepG2 & Human hepatocellular cytotoxicity & 39{,}312 & AUPRC\\
Cytotoxicity & HSkMC & Human skeletal-muscle cytotoxicity & 39{,}312 & AUPRC\\
Cytotoxicity & IMR-90 & Human lung-fibroblast cytotoxicity & 39{,}312 & AUPRC\\
\end{longtable}
\normalsize

\textbf{Dataset sizes and class distributions.} The six antibacterial datasets contained between 38,650 and 39,390 labelled entries each (\Cref{tab:antibacterial-target-tasks}). Positive labels were rare, ranging from 109 for \textit{P. aeruginosa} to 1,321 for \textit{N. gonorrhoeae}. Their prevalence varied more than tenfold across species, from 0.28\% to 3.42\%; negative labels accounted for the remaining 96.58--99.72\%. AUPRC was used to evaluate the retrieval of these rare positives. Dataset sizes count labelled entries rather than unique molecular structures. Repeated SMILES entries were retained in the benchmark files, and their numbers are reported alongside the numbers of unique SMILES strings.

\noindent\begin{minipage}{\linewidth}
\centering
\small
\setlength{\tabcolsep}{4pt}
\captionof{table}{Dataset sizes and class distributions for the six antibacterial target tasks. Size denotes the number of labelled entries, and positive labels indicate antibacterial activity. Unique SMILES were counted by exact string matching; repeated entries equal the total size minus the number of unique SMILES strings. All six tasks used AUPRC as their primary evaluation metric.}
\label{tab:antibacterial-target-tasks}
\begin{tabularx}{\linewidth}{Xrrrrrr}
\toprule
\textbf{Dataset} & \textbf{Size} & \textbf{Positive} & \textbf{Negative} & \textbf{Positive (\%)} & \textbf{\shortstack{Unique\\SMILES}} & \textbf{\shortstack{Repeated\\entries}}\\
\midrule
\textit{S. aureus} & 39{,}390 & 470 & 38{,}920 & 1.19 & 39{,}266 & 124\\
\textit{P. aeruginosa} & 39{,}267 & 109 & 39{,}158 & 0.28 & 39{,}266 & 1\\
\textit{A. baumannii} & 39{,}267 & 221 & 39{,}046 & 0.56 & 39{,}266 & 1\\
\textit{E. coli} & 39{,}267 & 222 & 39{,}045 & 0.57 & 39{,}266 & 1\\
\textit{K. pneumoniae} & 39{,}267 & 185 & 39{,}082 & 0.47 & 39{,}266 & 1\\
\textit{N. gonorrhoeae} & 38{,}650 & 1{,}321 & 37{,}329 & 3.42 & 38{,}622 & 28\\
\bottomrule
\end{tabularx}
\end{minipage}
\par\medskip

\textbf{Similarity-based data partitions.}
\label{app:partition}
The 22 TDC tasks used the official benchmark partitions and task-specific evaluation metrics. For each of the six antibacterial and three cytotoxicity tasks, we generated five fixed partitions using random seeds 1, 2, 3, 4 and 5. Molecules were represented by 2,048-bit Morgan fingerprints with radius 2 and grouped by greedy leader clustering at a Tanimoto threshold of 0.40 (\texttt{greedy\_\allowbreak leader\_\allowbreak tanimoto}). Clusters were assigned to training, validation and test sets with stratification by the binary label and target proportions of 8:1:1. All methods used the same partitions for a given task. \Cref{tab:antibiotic-splits} reports the actual sizes and positive counts for all 45 dataset--seed combinations; negative counts are obtained by subtracting positives from the corresponding split size. Within each seed, the training, validation and test sets had no overlapping dataset identifiers or exact SMILES strings.

The five seeds repartitioned the same dataset, with partially overlapping training and test sets across seeds. They therefore provided repeated evaluations on different partitions of the same data. Positive counts could vary between partitions despite label stratification. For example, \textit{S. aureus} test sets contained 72--82 positives, and \textit{E. coli} validation sets contained 22--50 positives. These counts are reported separately for each seed so that the class distributions underlying the validation and test scores are explicit.

\begin{table}[p]
\centering
\normalsize
\setlength{\tabcolsep}{4pt}
\renewcommand{\arraystretch}{1.10}
\caption{Training, validation and test sizes for the nine datasets. Total gives the number of samples and Positive gives the number with a positive label. Negative counts equal Total minus Positive. Seeds are grouped only when all partition sizes and positive counts are identical; grouped seeds retain different sample assignments.}
\label{tab:antibiotic-splits}
\begin{tabular*}{\linewidth}{@{\extracolsep{\fill}}llrrrrrr@{}}
\toprule
 & & \multicolumn{2}{c}{\textbf{Training}} & \multicolumn{2}{c}{\textbf{Validation}} & \multicolumn{2}{c}{\textbf{Test}}\\
\cmidrule(lr){3-4}\cmidrule(lr){5-6}\cmidrule(l){7-8}
\textbf{Dataset} & \textbf{Seed} & \textbf{Total} & \textbf{Positive} & \textbf{Total} & \textbf{Positive} & \textbf{Total} & \textbf{Positive}\\
\midrule
\textit{S. aureus} & 1 & 31{,}512 & 340 & 3{,}939 & 48 & 3{,}939 & 82\\
 & 2 & 31{,}512 & 347 & 3{,}939 & 48 & 3{,}939 & 75\\
 & 3 & 31{,}512 & 343 & 3{,}939 & 55 & 3{,}939 & 72\\
 & 4 & 31{,}512 & 335 & 3{,}939 & 59 & 3{,}939 & 76\\
 & 5 & 31{,}512 & 337 & 3{,}939 & 54 & 3{,}939 & 79\\
\addlinespace[3pt]
\textit{P. aeruginosa} & 1, 5 & 31{,}414 & 80 & 3{,}927 & 12 & 3{,}926 & 17\\
 & 2 & 31{,}414 & 80 & 3{,}927 & 13 & 3{,}926 & 16\\
 & 3 & 31{,}414 & 85 & 3{,}926 & 11 & 3{,}927 & 13\\
 & 4 & 31{,}414 & 81 & 3{,}926 & 11 & 3{,}927 & 17\\
\addlinespace[3pt]
\textit{A. baumannii} & 1--5 & 31{,}414 & 177 & 3{,}927 & 22 & 3{,}926 & 22\\
\addlinespace[3pt]
\textit{E. coli} & 1 & 31{,}413 & 177 & 3{,}927 & 22 & 3{,}927 & 23\\
 & 2 & 31{,}414 & 154 & 3{,}927 & 41 & 3{,}926 & 27\\
 & 3 & 31{,}414 & 178 & 3{,}927 & 22 & 3{,}926 & 22\\
 & 4 & 31{,}414 & 144 & 3{,}927 & 50 & 3{,}926 & 28\\
 & 5 & 31{,}414 & 160 & 3{,}927 & 35 & 3{,}926 & 27\\
\addlinespace[3pt]
\textit{K. pneumoniae} & 1 & 31{,}414 & 135 & 3{,}927 & 20 & 3{,}926 & 30\\
 & 2 & 31{,}414 & 126 & 3{,}927 & 20 & 3{,}926 & 39\\
 & 3 & 31{,}414 & 136 & 3{,}926 & 18 & 3{,}927 & 31\\
 & 4 & 31{,}414 & 138 & 3{,}926 & 18 & 3{,}927 & 29\\
 & 5 & 31{,}414 & 137 & 3{,}927 & 19 & 3{,}926 & 29\\
\addlinespace[3pt]
\textit{N. gonorrhoeae} & 1--5 & 30{,}920 & 1{,}057 & 3{,}865 & 132 & 3{,}865 & 132\\
\addlinespace[3pt]
HepG2 & 1--5 & 31{,}450 & 2{,}673 & 3{,}931 & 334 & 3{,}931 & 334\\
\addlinespace[3pt]
HSkMC & 1, 3--5 & 31{,}450 & 1{,}192 & 3{,}931 & 149 & 3{,}931 & 149\\
 & 2 & 31{,}449 & 1{,}192 & 3{,}931 & 149 & 3{,}932 & 149\\
\addlinespace[3pt]
IMR-90 & 1 & 31{,}450 & 2{,}692 & 3{,}931 & 345 & 3{,}931 & 410\\
 & 2 & 31{,}450 & 2{,}697 & 3{,}931 & 345 & 3{,}931 & 405\\
 & 3 & 31{,}450 & 2{,}699 & 3{,}931 & 345 & 3{,}931 & 403\\
 & 4 & 31{,}450 & 2{,}757 & 3{,}931 & 345 & 3{,}931 & 345\\
 & 5 & 31{,}450 & 2{,}715 & 3{,}931 & 345 & 3{,}931 & 387\\
\bottomrule
\end{tabular*}
\end{table}

\textbf{Compound overlap across tasks.} Compound identity across datasets was assessed by exact matching of the supplied SMILES strings. The nine antibacterial and cytotoxicity tasks contained 39,274 distinct SMILES strings in total, of which 38,583 (98.2\%) were present in all nine tasks. The three cytotoxicity datasets contained the same 39,224 unique SMILES strings; 39,219 were also present in each of the \textit{P. aeruginosa}, \textit{A. baumannii}, \textit{E. coli} and \textit{K. pneumoniae} datasets. Source and target tasks thus differed in biological endpoints while drawing largely on the same compound library. These counts refer to exact SMILES strings, without additional structure standardization for the overlap analysis.

\textbf{Precomputed molecular features.} Each split included 2,048-bit Morgan fingerprints~\citep{rogers2010extended} and 210 RDKit descriptors~\citep{rdkit}. Missing descriptor values were replaced by the median for that descriptor in the training set. Descriptors were standardized using means and standard deviations estimated from the training set, and the same transformation was applied to the validation and test sets. Raw and standardized descriptors were provided in the same row order as the corresponding input data.

\textbf{Workflow selection and statistical reporting.} Each antibacterial target involved a single workflow search. Every candidate was fitted and evaluated on all five fixed splits, and mean validation AUPRC guided search and selected the final workflow. Held-out test outcomes were not supplied as search feedback or used for workflow selection. Test AUPRC was summarized as the mean and sample standard deviation across the five splits. The standard deviation describes variation across the five data partitions. Systems were ranked separately within each task, with ties assigned their average rank. Mean ranks and mean AUPRC values gave all six tasks equal weight. AUROC denotes the area under the receiver-operating-characteristic curve, and AUPRC denotes the area under the precision--recall curve.

\textbf{Held-out \textit{S. aureus} activity test set.} \label{app:antibiotic-heldout} The predictive comparison in \Cref{fig:fig3}f used 271 compounds from the \texttt{All tested compounds} worksheet of Supplementary Data 2 in our previous study~\citep{wong2024discovery}. The worksheet contained 241 screening candidates, 30 low-predicted-activity controls and 12 known-antibiotic-class positive controls. We retained the candidates and low-score controls and excluded the positive controls identified by the \texttt{BLACTAM\_OR\_QUINOLONE} categories. The resulting set contained 21 active and 250 inactive compounds. Both predictors were evaluated against the same binary labels (\texttt{ACTIVITY}). The published scores (\texttt{ANTIBIOTIC\_PS}) gave an AUROC of 0.590 and an average precision of 0.090; average precision was used to calculate the comparator AUPRC. This predictive comparison was distinct from the campaign-level hit-rate comparison, which used 107 \method-selected and 241 expert-selected compounds (\Cref{app:antibiotic-experimental-results}).

\subsection{Evaluation of recursive self-improvement}
\label{app:experience-accumulation}
\label{app:antibiotic-search}

\textbf{Source experience.} \method developed workflows independently on each of the 25 source tasks and pooled the resulting experience in solution, refinement and execution memory. Each antibacterial target began with the same pooled source experience. Experience generated during a target search guided later iterations of that search and was not shared with the other targets.

\textbf{Resource library and RSI conditions.} The small-molecule resource library contained 110 reusable skills: 55 models, 48 methods and seven tools or resources. These covered molecular standardization, feature construction, predictive models, class-imbalance handling, calibration, stacking and ensembling. \method with RSI and \method without RSI used the same resource library, backbone model, data partitions, evaluation procedure and 20-attempt search allowance. \method with RSI began with the pooled source experience; \method without RSI began without experience from previous tasks. In both conditions, memory was updated during the current task to guide later iterations.

\textbf{Search trajectories.} At each iteration, we identified the completed workflow with the highest mean validation AUPRC obtained so far. Its test performance was calculated after workflow search to construct the trajectory in \Cref{fig:fig2}d. Test scores were not returned to the agent as feedback and were excluded from memory updates.

\textbf{Zero-search transfer.} For \Cref{fig:fig2}e, \method identified the best-matching source task using task type, evaluation metric, dataset size and task-description similarity, and then selected the workflow with the validation performance on that source task. Both selections preceded evaluation on the antibacterial target. Model parameters were fitted on the target training data where required, but no alternative target-workflow configurations were searched.

\textbf{Resource accounting.} One workflow-search iteration comprised construction, execution and validation evaluation of a candidate under the accounting rules in \Cref{app:antibiotic-baselines}. Zero-search transfer omitted target-workflow search but retained any required model fitting and inference. Skill construction, experience accumulation on source tasks, target search, final fitting and screening-library inference were separate computational stages.

\textbf{Experience-pool-size analysis.} We transferred one workflow from each source task to each of the six antibacterial targets, obtaining a matrix of 150 source--target evaluations. For each subset of 2--25 source tasks, we retrospectively selected the transfer test AUPRC available within that subset separately for each target, then averaged these six task performances. The mean curve in \Cref{fig:fig2}f is the analytical expectation of this quantity over all equally sized source-task subsets. 

To estimate the percentile bands, we generated 2,000 random permutations of the 25 source tasks using \texttt{numpy.random.default\_rng} with seed 20260901. For pool size $Z$, the first $Z$ tasks in each permutation defined a sampled pool, producing nested pools across sizes. We calculated the interquartile and 5th--95th percentile ranges of the six-target mean across these pools. The seed affected these sampled bands, not the analytical mean or the underlying source--target scores. The bands describe variation in pool composition, not independent search replicates or confidence intervals for the routing method. At size 25, only the complete pool was available.

\subsection{Case studies of experience-guided search}
\label{app:antibiotic-case-studies}

\textbf{Experience-guided workflow reuse for \textit{A. baumannii}.} We examined how a workflow developed on AMES supported workflow selection for \textit{A. baumannii}. Solution memory retained an equal-weight ensemble of LANTERN--RADR, MiniMol~\citep{klaser2024minimol}, ADMETrix and MapLight GNN, together with their component configurations. MiniMol used IA3-based adaptation~\citep{liu2022fewshot} of the prediction head, a gated multilayer perceptron and class-balanced sampling. MapLight GNN used a Set2Set readout~\citep{vinyals2016order} with three processing steps. The four components achieved AMES AUROCs of 0.8480--0.8633, and their ensemble achieved 0.8731 (\Cref{tab:abaumannii-reuse}). These configurations provided previously evaluated alternatives for the antibacterial task.

We transferred the components and their ensemble without changing their architectures or training settings. Model parameters were fitted on the \textit{A. baumannii} training data, and target validation performance selected among the five inherited configurations before test evaluation. This selected LANTERN--RADR, which achieved a mean test AUPRC of 0.5356. The unchanged four-component ensemble achieved 0.4742, and the other individual components ranged from 0.3235 to 0.3863. Thus, the ensemble that performed best on AMES did not provide the best transfer performance. Retaining its component configurations allowed target validation to identify a more effective predictor without generating or refining a new workflow.

Search without experience from previous tasks evaluated 20 workflows on \textit{A. baumannii}, comprising five initial model evaluations, six refinements and nine ensemble evaluations (\Cref{tab:abaumannii-search20}). The refinements changed molecular features, prediction heads, class-imbalance handling and training procedures. For example, ADMETTBoost--XGBoost~\citep{chen2016xgboost} incorporated Morgan-fingerprint features, focal loss~\citep{lin2017focal} and bagging. Euclia-ML variants used gradient-boosting heads, class-balanced focal loss and feature selection. Ensemble evaluations combined configurations evaluated earlier in the search.

The highest observed test AUPRC increased from 0.3716 at step 1 to 0.4134 after the ADMETTBoost--XGBoost refinement at step 3. It reached 0.4335 after the Euclia-ML refinement at step 5. Ensembles then achieved 0.4572 at step 6 and 0.4988 at step 7. The remaining 13 evaluations explored additional predictors, training modifications and ensembles but did not exceed 0.4988. These evaluations included gated fusion of molecular descriptors, cosine warm restarts~\citep{loshchilov2017sgdr}, stochastic weight averaging~\citep{izmailov2018averaging} and label smoothing. Additional ensemble configurations achieved test AUPRCs of 0.4005--0.4684, showing that further combinations of the evaluated models did not improve the best observed result in this run.

The validation-selected LANTERN--RADR configuration exceeded the best observed test AUPRC in the 20-step search by 0.0368, corresponding to a relative improvement of 7.4\%. This advantage was obtained without further workflow refinement, while retaining target-task model fitting and validation-based selection. Within the RSI framework, solution memory contributed to the effective search policy by supplying previously evaluated configurations for the new task. In this case, selecting and reusing one of these configurations produced a better predictor than the 20-step search. The unchanged source ensemble did not exceed that search, emphasizing the contribution of selecting a suitable configuration from the accumulated experience.

\noindent\begin{minipage}{\linewidth}
\centering
\small
\captionof{table}{\textbf{Transfer of AMES workflow configurations to \textit{A. baumannii}.} Target test AUPRC values are means across five seeds. LANTERN--RADR was selected using target validation performance before test evaluation. Source-task performance is reported as AUROC.}
\label{tab:abaumannii-reuse}
\begin{tabularx}{\linewidth}{@{}Xrr@{}}
\toprule
\textbf{Workflow configuration} & \textbf{AMES AUROC} & \textbf{\shortstack[r]{\textit{A. baumannii}\\test AUPRC}}\\
\midrule
LANTERN--RADR & 0.8633 & 0.5356\\
MiniMol & 0.8599 & 0.3235\\
ADMETrix & 0.8576 & 0.3863\\
MapLight GNN & 0.8480 & 0.3756\\
Equal-weight ensemble of all four & 0.8731 & 0.4742\\
\bottomrule
\end{tabularx}
\end{minipage}

\noindent\begin{minipage}{\linewidth}
\centering
\footnotesize
\captionof{table}{\textbf{Twenty step workflow search without experience from previous tasks on \textit{A. baumannii}.} Rows follow the execution order. Test AUPRC values are means across five seeds; bold type identifies the highest observed value. Each ensemble combined configurations evaluated earlier in the run. Workflow selection used validation performance, as described in \Cref{app:antibiotic-search}; test scores are shown to compare the resulting predictors.}
\label{tab:abaumannii-search20}
\begin{tabularx}{\linewidth}{@{}rXr@{}}
\toprule
\textbf{Step} & \textbf{Configuration and modification} & \textbf{Test AUPRC}\\
\midrule
1 & Chemprop--RDKit ensemble, initial configuration & 0.3716\\
2 & Euclia-ML, initial configuration & 0.3561\\
3 & ADMETTBoost--XGBoost: Morgan-fingerprint features, focal loss and five-seed bagging & 0.4134\\
4 & Euclia-ML: gradient-boosting head, class-balanced focal loss, feature selection and five-seed bagging & 0.3869\\
5 & Euclia-ML: gradient-boosting head, class-balanced focal loss and feature selection & 0.4335\\
6 & Ensemble of previously evaluated configurations & 0.4572\\
7 & Ensemble of previously evaluated configurations & \textbf{0.4988}\\
8 & CheMeleon, initial configuration & 0.3483\\
9 & Ensemble of previously evaluated configurations & 0.4561\\
10 & Graph-transformer-QM-pretrain: class-balanced focal loss, gated RDKit-feature fusion and cosine warm restarts & 0.3707\\
11 & Ensemble of previously evaluated configurations & 0.4451\\
12 & Ensemble of previously evaluated configurations & 0.4632\\
13 & ADMETTBoost--XGBoost, initial configuration & 0.3690\\
14 & Ensemble of previously evaluated configurations & 0.4630\\
15 & Frad: gated descriptor fusion, stochastic weight averaging, class-balanced binary cross-entropy and label smoothing & 0.3669\\
16 & Ensemble of previously evaluated configurations & 0.4465\\
17 & MTAN--ADMET, initial configuration & 0.3578\\
18 & Ensemble of previously evaluated configurations & 0.4684\\
19 & Euclia-ML: gradient-boosting head, class-balanced focal loss and feature selection & 0.3791\\
20 & Ensemble of previously evaluated configurations & 0.4005\\
\bottomrule
\end{tabularx}
\end{minipage}

\subsection{Virtual screening and compound selection}
\label{app:antibiotic-screening}

\textbf{Training data and screening library.} The prospective activity and cytotoxicity models used the same training datasets as our previous study~\citep{wong2024discovery}. The activity models were trained on methicillin-susceptible \textit{S. aureus} (MSSA; strain RN4220), and cytotoxicity was predicted for HepG2, HSkMC and IMR-90 cells. Both studies screened the same library of 12,076,365 compounds, comprising 11,277,225 entries from the Mcule purchasable database and 799,140 entries from the Broad Institute collection.

\textbf{Selected screening models.}
\label{app:screening-models}
The \method workflow used separately optimized predictors for activity and cytotoxicity (\Cref{fig:supp-antibiotic-workflow}). HepG2 and IMR-90 cytotoxicity predictors combined MiniMol embeddings with Morgan fingerprints and RDKit descriptors and used residual multilayer-perceptron heads. The HSkMC predictor used MiniMol embeddings with gated multilayer-perceptron heads. Activity ranking combined two MiniMol-based variants and a LANTERN--RADR ensemble with weights of 0.55, 0.30 and 0.15, respectively. These weights combined the outputs of the three component-specific ensembles.

\textbf{Filtering, ranking and diversity selection.} \label{app:screening-funnel} Screening proceeded through structural novelty filtering, cytotoxicity filtering, activity ranking, Butina clustering and diverse picking (\Cref{fig:supp-antibiotic-workflow}). Structural filtering removed compounds matching PAINS or Brenk alerts~\citep{baell2010pains,brenk2008lessons} or established antibiotic cores. It also required a maximum Tanimoto similarity below 0.40 to known antibiotics. The excluded core families were $\beta$-lactams, quinolones, oxazolidinones, nitrofurans and sulfonamides. Cytotoxicity filtering required predicted scores of $\leq0.40$ for each of HepG2, HSkMC and IMR-90. The retained compounds were ranked by the weighted antibacterial-activity ensemble described above. Butina clustering~\citep{butina1999unsupervised} grouped structurally similar candidates, and diverse picking selected compounds across clusters to retain chemical diversity. We selected and experimentally tested 107 compounds; the final selection included documented exceptions to the nominal cytotoxicity criterion.

\subsection{Comparison with the previous expert-driven workflow}
\label{app:antibiotic-workflow-comparison}

\textbf{Workflow differences.} The workflows differed in their predictive models, filter order and thresholds, and compound-selection procedures (\Cref{fig:supp-antibiotic-workflow}). The expert-driven workflow applied activity prediction before cytotoxicity and structural filtering. \method applied structural filtering before cytotoxicity prediction and activity ranking. Training data and library composition are specified in \Cref{app:antibiotic-screening}.

\textbf{Expert-developed predictors and filtering.} The previous study used separate 20-member Chemprop ensembles~\citep{yang2019analyzing} with RDKit descriptors for \textit{S. aureus} activity and for each of the three cytotoxicity tasks~\citep{wong2024discovery}. Activity-score thresholds of $>0.4$ for Mcule and $>0.2$ for the Broad Institute collection retained 10,310 compounds. Requiring all three cytotoxicity scores to be $<0.2$ reduced this set to 3,646. PAINS/Brenk filtering retained 2,209 compounds, and similarity and antibiotic-core filtering retained 1,261. The similarity filter used a maximum Tanimoto similarity below 0.5 to training actives, and the excluded core families were $\beta$-lactams and quinolones.

\textbf{Candidate selection.} In the expert-driven workflow, Monte Carlo tree search identified substructures supporting antibacterial predictions and informed selection of compounds with or without such rationales. Compounds were sourced according to availability and purity greater than 90\%, and 241 filtered candidates were tested. The \method workflow instead ranked candidates with its weighted activity ensemble and selected structurally diverse compounds across Butina clusters, leading to 107 selected and tested compounds. The substructure search in the previous study served molecular interpretation and candidate selection; it was a different application of tree search from the workflow-development search performed by \method.

\textbf{Comparison endpoints.} Predictive discrimination was evaluated against shared labels, whereas experimental hit proportions were calculated within each campaign's selected candidates. These analyses assess prediction and complete-workflow yield, respectively (\Cref{app:antibiotic-heldout,app:antibiotic-experimental-results}).

\subsection{Experimental validation}
\label{app:antibiotic-assays}

\subsubsection{Antibacterial activity assays}

\textbf{Bacterial strains and culture.} \textit{S. aureus} RN4220 (MSSA) was obtained from DSMZ (DSM 26309) and \textit{S. aureus} FPR3757/USA300 (MRSA) from ATCC (BAA-1556). Strains were authenticated by the suppliers by next-generation or Sanger sequencing. Single colonies were picked from LB agar (1.5\% Difco agar; BD 244520) and grown aerobically in liquid LB (BD 244620; 2--3~mL working volume in 14~mL Falcon tubes) at $37\,^{\circ}\mathrm{C}$ with shaking at 300~rpm, without antibiotic selection.

\textbf{Compounds.} Compounds were purchased from ChemBridge (C3; ID 7359936) and Vitas-M Laboratories (C32; STL235206) and dissolved in DMSO (MilliporeSigma D5879) at $12.8\,\mathrm{mg\,mL^{-1}}$.

\textbf{Inoculum.} A suspension of approximately $10^{5}\,\mathrm{CFU\,mL^{-1}}$ was prepared either by a 1:10,000 dilution of an overnight culture picked from a single colony or by a 1:500 dilution of an $\mathrm{OD}_{600}$ 0.08 suspension in fresh LB.

\textbf{MSSA screen and dose--response plates.} Compounds were screened against MSSA at a single dose of $128\,\mu\mathrm{g\,mL^{-1}}$ with a PrestoBlue readout. Compounds that inhibited growth were advanced to an eight-point two-fold dose--response series (128 to $1\,\mu\mathrm{g\,mL^{-1}}$) in duplicate 384-well plates. Compound was dispensed from DMSO stock by acoustic dispensing (Echo; maximum DMSO 1.0\%) before addition of inoculated LB to a final volume of $64\,\mu\mathrm{L}$. Each plate carried sterility (medium only), untreated growth, Triton X-100 kill and DMSO vehicle (1.000\% to 0.008\%) controls.

\textbf{MRSA dose--response plate.} The MRSA MIC for C32 was determined in a 96-well plate with a nine-point two-fold series (128 to $0.5\,\mu\mathrm{g\,mL^{-1}}$) in three replicate rows, a DMSO vehicle series, a Triton X-100 dilution series, untreated growth wells and medium-only outer wells.

\textbf{Incubation and readout.} Plates were covered with a lid (not sealed) and incubated for 18--24~h at $37\,^{\circ}\mathrm{C}$ with 5\% $\mathrm{CO}_{2}$. $\mathrm{OD}_{600}$ was read on a SpectraMax M3 multi-mode microplate reader (Molecular Devices). MIC was the lowest concentration giving $\geq90\%$ growth inhibition relative to the untreated growth control.

\subsubsection{Cytotoxicity assays}

\textbf{Mammalian cell culture.} HepG2, HSkMC and IMR-90 cells were obtained from ATCC and maintained at $37\,^{\circ}\mathrm{C}$ and 5\% $\mathrm{CO}_{2}$ in humidified incubators. Cells were passaged fewer than 5--10 times before use. All lines were authenticated by ATCC using short tandem repeat profiling. Culture media are listed in \Cref{tab:antibiotic-cell-lines}.

\noindent\begin{minipage}{\linewidth}
\centering
\captionof{table}{Human cell lines and culture media.}
\label{tab:antibiotic-cell-lines}
\small
\renewcommand{\arraystretch}{1.15}
\begin{tabularx}{\linewidth}{@{}>{\raggedright\arraybackslash}p{0.13\linewidth}>{\raggedright\arraybackslash}p{0.17\linewidth}>{\raggedright\arraybackslash}X@{}}
\toprule
\textbf{Cell line} & \textbf{ATCC no.} & \textbf{Culture medium} \\
\midrule
HepG2 & HB-8065 & DMEM, high glucose (Corning 10-013-CV) + 10\% FBS (Thermo Fisher 16140071) + 1\% penicillin--streptomycin (Thermo Fisher 15070063) \\
\addlinespace
HSkMC & PCS-950-010 & Mesenchymal Stem Cell Basal Medium for Adipose, Umbilical and Bone Marrow-derived MSCs (ATCC PCS-500-030) + Primary Skeletal Muscle Growth Kit (ATCC PCS-950-040) + 1\% penicillin--streptomycin \\
\addlinespace
IMR-90 & CCL-186 & Eagle's Minimum Essential Medium (EMEM; ATCC 30-2003) + 10\% FBS + 1\% penicillin--streptomycin \\
\bottomrule
\end{tabularx}
\end{minipage}

\textbf{Dosing and controls.} Cells were plated at $5$--$10\times10^{3}$ cells per well in 96-well clear flat-bottom black tissue-culture-treated plates (Corning 3603), $100\,\mu\mathrm{L}$ per well, and dosed 24~h later. For dosing, column 2 was brought to $198\,\mu\mathrm{L}$ with medium and $2\,\mu\mathrm{L}$ of $12.8\,\mathrm{mg\,mL^{-1}}$ DMSO stock was added ($128\,\mu\mathrm{g\,mL^{-1}}$, 1\% DMSO). One hundred microlitres were then carried across nine columns by two-fold serial transfer and the final $100\,\mu\mathrm{L}$ was discarded, so that DMSO fell from 1\% to 0.004\% in step with compound. Untreated wells ($n=4$) defined 100\% viability. A matched DMSO series was run in separate rows. Triton X-100 (0.1\% final) lysis wells defined the assay floor and were added 30--60~min before reading. Media-only wells with dye served as background.

\textbf{Viability readout and analysis.} PrestoBlue HS Cell Viability Reagent (Invitrogen P50200) was added at 10\% of the well volume ($10\,\mu\mathrm{L}$ per $100\,\mu\mathrm{L}$) and plates were incubated for 4~h at $37\,^{\circ}\mathrm{C}$. Fluorescence (excitation/emission, 550/590~nm) was read on a SpectraMax M3 multi-mode microplate reader (Molecular Devices) from the top and bottom. Viability was calculated as $(\text{well}-\text{background})/(\text{untreated}-\text{background})\times100$. $\mathrm{CC}_{50}$ was calculated by $\log_{2}$ interpolation across the 50\% crossing, and the two bracketing concentrations are reported alongside it. HepG2 values were obtained in separate earlier plates with the same PrestoBlue readout.

\subsubsection{Predictive performance and experimental hit rates} \label{app:antibiotic-experimental-results} Predictive performance was evaluated as described in \Cref{app:antibiotic-heldout}. Experimental hit rates used the 107 \method-selected compounds and the 241 expert-selected candidates as their denominators. Based on the recorded endpoints, the campaigns yielded 17/107 (15.9\%) and 21/241 (8.7\%) hits, respectively. A hit was defined by an MSSA MIC of $\leq32\,\mu\mathrm{g\,mL^{-1}}$, distinct from the binary labels used for predictive evaluation. The comparator denominator excluded low-predicted-activity and known-antibiotic-class controls. Hit proportions were reported with 95\% Wilson binomial confidence intervals~\citep{wilson1927probable}; no between-group hypothesis tests were reported. The numerical MIC threshold was aligned across studies, without assuming identical experimental protocols.

\section{\method for peptide--MHC binder design}
\label{app:protein-verifiers}

\subsection{Datasets and benchmark evaluation}

\label{app:protein-datasets}

The protein study comprised 14 independent miniprotein-binder design tasks, each with one held-out target, followed by one pMHC binder-design task using both gp100 datasets for workflow search and MART-1 for testing. The resulting pipeline was also used for prospective binder generation (\Cref{sec:method-protein}). Protein experiments used no small-molecule memory records.

\textbf{Miniprotein-binder experience tasks.} The fibroblast growth factor receptor 2 (FGFR2), insulin receptor (InsulinR), platelet-derived growth factor receptor (PDGFR), transforming growth factor-\(\beta\) (TGF\(\beta\)), interleukin-7 receptor-\(\alpha\) (IL-7R\(\alpha\)), VirB8-like protein from \textit{Rickettsia typhi} (VirB8), epidermal growth factor receptor (EGFR), tropomyosin receptor kinase A (TrkA), influenza A H3 haemagglutinin (H3), and angiopoietin-1 receptor (TIE2) datasets were obtained from Cao et al.~\citep{cao2022design}. The SARS-CoV-2 spike receptor-binding domain (SARS-CoV-2 RBD) dataset was obtained from Cao et al.~\citep{cao2020novo}. The anaplastic lymphoma kinase (ALK), leukocyte tyrosine kinase (LTK), and interleukin-2 receptor-\(\alpha\) (IL-2R\(\alpha\)) datasets were obtained from Bennett et al.~\citep{bennett2023improving}. The source datasets provide binary binding labels at 4,000 and 400~nM. We used the source-provided 4,000-nM label as the benchmark outcome. Across the 14 retained targets, the source collections contained 901,670 designs. To construct a computationally tractable benchmark while preserving all positive examples, we retained all 1,565 positive designs and sampled up to ten negative designs per positive within each target. The resulting benchmark contained 17,215 designs: 1,565 positives and 15,650 sampled negatives (\Cref{tab:minibinder-target-data}).

Each of the 14 targets defined a distinct design task under a leave-one-target-out partition. Workflow search used the other 13 targets, with AUPRC guiding selection. The selected workflow was frozen and then evaluated on the held-out target. Held-out-target labels and performance were excluded from search and workflow selection. The tasks ran independently without cross-task memory. Records of the evaluated workflows, modifications and execution outcomes were subsequently pooled into the protein-domain memory \(\mathcal{Z}_{\mathrm{mini}}\) used for the pMHC task. This source memory contained no pMHC or prospective-design outcomes.

\begingroup
\small
\setlength{\tabcolsep}{3pt}
\begin{longtable}{@{}p{2.2cm}p{5.0cm}rrr@{}}
\caption{Composition of the 14-task miniprotein-binder benchmark. Each row identifies the held-out test target for one task; workflow search and selection used the other 13 targets. Labels use the source-provided binary binding endpoint at 4{,}000~nM. All positives were retained, and up to ten negatives per positive were sampled within each target.}
\label{tab:minibinder-target-data}\\
\toprule
\textbf{Target} & \textbf{Data source} & \textbf{Source designs} & \textbf{Retained positives} & \textbf{Sampled negatives} \\
\midrule
\endfirsthead
\multicolumn{5}{l}{\tablename~\thetable\ (continued)}\\
\toprule
\textbf{Target} & \textbf{Data source} & \textbf{Source designs} & \textbf{Retained positives} & \textbf{Sampled negatives} \\
\midrule
\endhead
\midrule
\multicolumn{5}{r}{Continued on next page}\\
\endfoot
\bottomrule
\endlastfoot
EGFR & Cao et al.~\citep{cao2022design} & 99{,}803 & 15 & 150 \\
FGFR2 & Cao et al.~\citep{cao2022design} & 59{,}999 & 601 & 6{,}010 \\
H3 & Cao et al.~\citep{cao2022design} & 59{,}941 & 49 & 490 \\
IL-7R\(\alpha\) & Cao et al.~\citep{cao2022design} & 14{,}999 & 22 & 220 \\
InsulinR & Cao et al.~\citep{cao2022design} & 59{,}973 & 256 & 2{,}560 \\
PDGFR & Cao et al.~\citep{cao2022design} & 99{,}992 & 284 & 2{,}840 \\
SARS-CoV-2 RBD & Cao et al.~\citep{cao2020novo} & 99{,}998 & 18 & 180 \\
TGF\(\beta\) & Cao et al.~\citep{cao2022design} & 15{,}000 & 100 & 1{,}000 \\
TIE2 & Cao et al.~\citep{cao2022design} & 100{,}000 & 5 & 50 \\
TrkA & Cao et al.~\citep{cao2022design} & 15{,}000 & 10 & 100 \\
VirB8 & Cao et al.~\citep{cao2022design} & 15{,}011 & 72 & 720 \\
\addlinespace
ALK & Bennett et al.~\citep{bennett2023improving} & 99{,}980 & 28 & 280 \\
IL-2R\(\alpha\) & Bennett et al.~\citep{bennett2023improving} & 61{,}994 & 55 & 550 \\
LTK & Bennett et al.~\citep{bennett2023improving} & 99{,}980 & 50 & 500 \\
\midrule
\textbf{Total} & 14 retained targets & \textbf{901{,}670} & \textbf{1{,}565} & \textbf{15{,}650} \\
\end{longtable}
\endgroup

\textbf{pMHC benchmark data.} The single pMHC task used datasets generated by the experimental screening procedure described by Liu et al.~\citep{liu2025design}. For all three candidate collections, the binary outcome indicated whether a candidate was selected in the fluorescence-activated cell sorting (FACS) experiment. Selected candidates were labelled positive and unselected candidates negative.

The task included HLA-A2 complexes presenting the gp100-derived peptides YLEPGPVTA and IMDQVPFSV, and HLA-A*02:01 complexes presenting MART-1 (A2L; ELAGIGILTV). 
The source structures for the two gp100 complexes were PDB 5EU3 and 6VM9, respectively. All three candidate collections belonged to one pMHC benchmark task. Both gp100 collections were used directly for workflow search, whereas MART-1 comprised the held-out test set. AUROC on the gp100 collections guided workflow selection. The selected workflow was frozen before evaluation on MART-1, whose labels and performance were excluded from workflow search, selection and memory updates. \Cref{tab:pmhc-datasets} gives the dataset roles, sizes and class counts. \Cref{fig:fig4}c reports test enrichment for MART-1.

\begin{table}[h]
\centering
\small
\caption{Dataset roles within the single pMHC binder-design task. Both complete gp100 collections were used for workflow search; MART-1 was reserved for test evaluation.}
\label{tab:pmhc-datasets}
\begin{tabular}{lllrr}
\toprule
\textbf{Antigen} & \textbf{Peptide} & \textbf{Role} & \textbf{Candidates} & \textbf{Positives} \\
\midrule
gp100 & YLEPGPVTA & Search & 2{,}611 & 14 \\
gp100 & IMDQVPFSV & Search & 3{,}271 & 4 \\
\addlinespace
MART-1 & ELAGIGILTV & Test & 3{,}845 & 27 \\
\bottomrule
\end{tabular}
\end{table}

\textbf{Metric calculation and aggregation.} EF was calculated as defined in \Cref{eq:protein-ef}. 
Macro EF and AUPRC gave equal weight to each held-out target for the miniprotein-binder benchmark. 

\subsection{Fixed expert-defined ranking objectives}
\label{app:protein-expert-comparison}

\textbf{Fixed objectives from established expert-defined design pipelines.} BoltzGen~\citep{stark2025boltzgen}, BindCraft~\citep{pacesa2025one}, and PXDesign~\citep{team2025pxdesign} are complete binder design pipelines. For the retrospective ranking benchmarks, we isolated a fixed loss or ranking criterion used internally by each pipeline and evaluated it retrospectively on the same target--binder sequence pairs. Thus, these references compare the criteria used to prioritize designs.

BoltzGen used its worst-rank quality criterion on Boltz-2~\citep{passaro2025boltz} predictions. If \(r_{ij}\) is the within-target percentile rank of candidate \(i\) under metric \(j\), with smaller ranks better, the reported higher-is-better score was $q_i=-\max_j r_{ij}$. The primary three-metric reference used equal rank weights for \texttt{design\_iiptm} (which restricts the design side of the pTM calculation to residues within \(8\,\text{\AA}\) of the target), \texttt{design\_ptm}, and the negative minimum design-to-target PAE. This is the confidence-only form of the BoltzGen criterion. The complete pipeline additionally ranks PLIP~\citep{salentin2015plip} hydrogen bonds, PLIP salt bridges, and \(\Delta\)SASA. We evaluated both the three-confidence-score and six-score forms and report the former as the stronger, and therefore more conservative, fixed comparator.

For BindCraft, we reproduced the default eight-term \texttt{default\_4stage\_multimer} design loss, \[ \begin{split} L_{\mathrm{BindCraft}}={}& 0.1\,\ell_{\mathrm{pLDDT}} +0.4\,\ell_{\mathrm{PAE}} +0.1\,\ell_{\mathrm{iPAE}} +1.0\,\ell_{\mathrm{contact}} +1.0\,\ell_{\mathrm{interface\ contact}}\\ &+0.3\,\ell_{\mathrm{Rg}} +0.05\,\ell_{\mathrm{ipTM}} -0.3\,\ell_{\mathrm{helix}}, \end{split} \] with the termini-distance loss disabled as in the default configuration. Terms were evaluated from AlphaFold2-Multimer-v3~\citep{evans2021protein} predictions with target-chain MSAs. The benchmark score was \(-L_{\mathrm{BindCraft}}\), so that larger values were better.

For PXDesign, we reproduced the seven-term PXDesign-h hallucination loss on Protenix-Mini~\citep{gong2025protenixmini} predictions, \[ \begin{split} L_{\mathrm{PXDesign}}={}& 0.15\,\ell_{\mathrm{pLDDT}} +0.4\,\ell_{\mathrm{PAE}} +0.1\,\ell_{\mathrm{iPAE}} +1.0\,\ell_{\mathrm{contact}} +1.0\,\ell_{\mathrm{interface\ contact}}\\ &-0.3\,\ell_{\mathrm{helix}} +0.3\,\ell_{\mathrm{Rg}}, \end{split} \] and again used its negative as the higher-is-better benchmark score.

Because the retrospective candidate collection did not contain pipeline-native starting backbones, this comparison did not reproduce generator-specific initial-guess or design-to-refold RMSD filters, iterative sequence optimization, or final diversity selection. It should therefore be interpreted as a comparison of fixed internal objectives from the three design pipelines rather than an end-to-end comparison of the pipelines themselves.

\subsection{Ranking benchmark results}
\label{app:protein-results}

\textbf{Miniprotein-binder benchmark.} \Cref{fig:supp-protein-minibinder} compares the five agent workflows and the three fixed expert-defined objectives using macro EF at the 1\%, 2\%, 5\% and 10\% cutoffs and macro AUPRC. Each macro metric gives the 14 targets equal weight. \method achieved macro EF values of 8.01 at 1\% and 7.29 at 2\%, with Biomni next at 1\% (7.62) and Claude Science next at 2\% (6.71). At 5\%, Claude Science had a macro EF of 5.56, compared with 5.55 for \method. At 10\%, \method had the highest macro EF (4.11). \method also achieved the highest macro AUPRC (0.4305), followed by Claude Science (0.4237). Among the eight methods, \method ranked first by AUPRC on six targets, with mean target-level rank 2.43, compared with 2.79 for Claude Science. \Cref{tab:protein-ef-minibinder} summarizes all target-level EF and AUPRC performance. 

\begingroup
\small
\setlength{\tabcolsep}{5pt}
\begin{longtable}{@{}lrrrrr@{}}
\caption{Target-level results for the miniprotein-binder benchmark. Each target was held out during the corresponding agent workflow search; no cross-task memory was used. Fixed objectives were evaluated on the same candidates. The target-specific rows report EF at four cutoffs and AUPRC.}
\label{tab:protein-ef-minibinder}\\
\toprule
\textbf{Method} & \textbf{EF@1}\% & \textbf{EF@2}\% & \textbf{EF@5}\% & \textbf{EF@10}\% & \textbf{AUPRC} \\
\midrule
\endfirsthead
\multicolumn{6}{l}{\tablename~\thetable\ (continued)}\\
\toprule
\textbf{Method} & \textbf{EF@1}\% & \textbf{EF@2}\% & \textbf{EF@5}\% & \textbf{EF@10}\% & \textbf{AUPRC} \\
\midrule
\endhead
\midrule
\multicolumn{6}{r}{Continued on next page}\\
\endfoot
\bottomrule
\endlastfoot
\multicolumn{6}{@{}l}{\textbf{FGFR2}}\\*
\method & 5.58 & 4.47 & 3.16 & 2.61 & 0.2041 \\*
Claude Science & 7.06 & 5.87 & 4.22 & 3.22 & 0.2871 \\*
Biomni & 6.07 & 5.21 & 3.79 & 2.69 & 0.2212 \\*
STELLA & 4.93 & 4.71 & 3.82 & 2.82 & 0.2193 \\*
Codex & 4.27 & 3.14 & 2.69 & 2.09 & 0.1554 \\*
BoltzGen & 3.28 & 3.06 & 2.33 & 2.24 & 0.1636 \\*
BindCraft & 5.25 & 4.05 & 3.46 & 2.63 & 0.2091 \\*
PXDesign & 2.46 & 2.65 & 2.53 & 1.98 & 0.1523 \\[4pt]
\multicolumn{6}{@{}l}{\textbf{PDGFR}}\\*
\method & 5.84 & 5.59 & 3.99 & 3.16 & 0.2529 \\*
Claude Science & 5.84 & 5.59 & 3.99 & 2.88 & 0.2405 \\*
Biomni & 5.84 & 4.37 & 3.71 & 2.88 & 0.2291 \\*
STELLA & 4.12 & 4.19 & 3.08 & 2.50 & 0.1998 \\*
Codex & 4.12 & 3.84 & 3.08 & 2.42 & 0.1815 \\*
BoltzGen & 4.12 & 4.02 & 3.01 & 2.46 & 0.1818 \\*
BindCraft & 4.47 & 4.71 & 3.15 & 3.06 & 0.2277 \\*
PXDesign & 4.12 & 3.84 & 2.94 & 2.85 & 0.2118 \\[4pt]
\multicolumn{6}{@{}l}{\textbf{InsulinR}}\\*
\method & 9.86 & 9.07 & 7.57 & 5.93 & 0.5854 \\*
Claude Science & 9.86 & 9.07 & 7.80 & 6.44 & 0.6364 \\*
Biomni & 9.86 & 9.07 & 7.72 & 6.16 & 0.6191 \\*
STELLA & 7.59 & 7.91 & 7.57 & 5.97 & 0.5686 \\*
Codex & 9.10 & 8.30 & 6.79 & 5.58 & 0.5326 \\*
BoltzGen & 8.72 & 7.33 & 6.32 & 5.07 & 0.4618 \\*
BindCraft & 4.93 & 5.40 & 5.85 & 5.11 & 0.4176 \\*
PXDesign & 7.97 & 6.95 & 4.99 & 4.21 & 0.3831 \\[4pt]
\multicolumn{6}{@{}l}{\textbf{TGF\(\beta\)}}\\*
\method & 4.00 & 4.00 & 2.80 & 2.00 & 0.1994 \\*
Claude Science & 4.00 & 4.00 & 2.20 & 1.80 & 0.1766 \\*
Biomni & 4.00 & 2.00 & 1.40 & 1.20 & 0.1299 \\*
STELLA & 7.00 & 3.50 & 2.60 & 2.10 & 0.1962 \\*
Codex & 4.00 & 2.00 & 1.20 & 1.40 & 0.1328 \\*
BoltzGen & 2.00 & 2.50 & 1.60 & 1.60 & 0.1260 \\*
BindCraft & 1.00 & 0.50 & 1.00 & 1.00 & 0.1054 \\*
PXDesign & 0.00 & 1.00 & 1.40 & 1.40 & 0.1027 \\[4pt]
\multicolumn{6}{@{}l}{\textbf{VirB8}}\\*
\method & 9.62 & 6.19 & 3.85 & 2.48 & 0.2394 \\*
Claude Science & 5.50 & 4.12 & 3.30 & 2.06 & 0.2154 \\*
Biomni & 5.50 & 3.44 & 2.20 & 1.93 & 0.1511 \\*
STELLA & 1.38 & 2.06 & 1.93 & 1.10 & 0.1043 \\*
Codex & 0.00 & 2.75 & 1.38 & 0.82 & 0.0957 \\*
BoltzGen & 0.00 & 0.00 & 1.65 & 1.10 & 0.0939 \\*
BindCraft & 2.75 & 2.75 & 2.48 & 1.65 & 0.1590 \\*
PXDesign & 1.38 & 2.06 & 1.65 & 1.51 & 0.1242 \\[4pt]
\multicolumn{6}{@{}l}{\textbf{IL-2R\(\alpha\)}}\\*
\method & 9.43 & 8.46 & 7.10 & 6.49 & 0.5719 \\*
Claude Science & 11.00 & 10.15 & 8.52 & 5.95 & 0.6872 \\*
Biomni & 9.43 & 6.77 & 7.45 & 6.13 & 0.5583 \\*
STELLA & 7.86 & 5.08 & 4.61 & 4.33 & 0.4159 \\*
Codex & 3.14 & 5.08 & 5.32 & 4.69 & 0.3865 \\*
BoltzGen & 4.71 & 4.23 & 5.32 & 5.95 & 0.4475 \\*
BindCraft & 1.57 & 3.38 & 4.97 & 5.05 & 0.4037 \\*
PXDesign & 3.14 & 3.38 & 3.90 & 4.15 & 0.3511 \\[4pt]
\multicolumn{6}{@{}l}{\textbf{LTK}}\\*
\method & 11.00 & 11.00 & 8.64 & 6.20 & 0.6376 \\*
Claude Science & 11.00 & 11.00 & 9.04 & 6.60 & 0.7122 \\*
Biomni & 11.00 & 11.00 & 9.43 & 6.80 & 0.7102 \\*
STELLA & 9.17 & 9.00 & 7.86 & 6.20 & 0.6222 \\*
Codex & 9.17 & 10.00 & 9.43 & 6.80 & 0.6761 \\*
BoltzGen & 9.17 & 9.00 & 8.25 & 7.40 & 0.6607 \\*
BindCraft & 11.00 & 9.00 & 7.86 & 6.00 & 0.5993 \\*
PXDesign & 9.17 & 7.00 & 6.68 & 6.20 & 0.5561 \\[4pt]
\multicolumn{6}{@{}l}{\textbf{H3}}\\*
\method & 1.83 & 1.00 & 0.81 & 1.02 & 0.1378 \\*
Claude Science & 0.00 & 2.00 & 1.63 & 1.43 & 0.1282 \\*
Biomni & 0.00 & 0.00 & 0.81 & 1.02 & 0.0907 \\*
STELLA & 0.00 & 1.00 & 0.81 & 0.41 & 0.0830 \\*
Codex & 0.00 & 1.00 & 0.81 & 0.41 & 0.0805 \\*
BoltzGen & 0.00 & 1.00 & 0.81 & 0.41 & 0.0824 \\*
BindCraft & 0.00 & 2.00 & 2.04 & 1.63 & 0.1267 \\*
PXDesign & 1.83 & 2.00 & 2.44 & 1.43 & 0.1118 \\[4pt]
\multicolumn{6}{@{}l}{\textbf{ALK}}\\*
\method & 11.00 & 11.00 & 9.62 & 8.16 & 0.8138 \\*
Claude Science & 11.00 & 9.43 & 8.25 & 6.39 & 0.6961 \\*
Biomni & 11.00 & 11.00 & 8.25 & 6.39 & 0.6847 \\*
STELLA & 8.25 & 7.86 & 6.88 & 5.68 & 0.5902 \\*
Codex & 8.25 & 9.43 & 6.88 & 5.68 & 0.5895 \\*
BoltzGen & 8.25 & 9.43 & 7.56 & 6.03 & 0.6290 \\*
BindCraft & 11.00 & 9.43 & 8.94 & 7.45 & 0.7032 \\*
PXDesign & 5.50 & 6.29 & 4.12 & 3.55 & 0.3184 \\[4pt]
\multicolumn{6}{@{}l}{\textbf{IL-7R\(\alpha\)}}\\*
\method & 11.00 & 11.00 & 8.46 & 6.60 & 0.6778 \\*
Claude Science & 7.33 & 8.80 & 9.31 & 6.16 & 0.6721 \\*
Biomni & 11.00 & 11.00 & 9.31 & 7.04 & 0.7550 \\*
STELLA & 7.33 & 8.80 & 9.31 & 7.04 & 0.6762 \\*
Codex & 11.00 & 11.00 & 9.31 & 7.92 & 0.7711 \\*
BoltzGen & 11.00 & 8.80 & 10.15 & 6.60 & 0.7188 \\*
BindCraft & 7.33 & 4.40 & 5.92 & 5.72 & 0.4634 \\*
PXDesign & 3.67 & 2.20 & 5.92 & 5.72 & 0.4362 \\[4pt]
\multicolumn{6}{@{}l}{\textbf{SARS-CoV-2 RBD}}\\*
\method & 11.00 & 11.00 & 8.80 & 4.95 & 0.5757 \\*
Claude Science & 11.00 & 8.25 & 5.50 & 3.30 & 0.3523 \\*
Biomni & 11.00 & 8.25 & 5.50 & 3.85 & 0.3978 \\*
STELLA & 11.00 & 8.25 & 6.60 & 4.40 & 0.4419 \\*
Codex & 11.00 & 8.25 & 5.50 & 3.30 & 0.3461 \\*
BoltzGen & 11.00 & 8.25 & 5.50 & 3.30 & 0.3596 \\*
BindCraft & 5.50 & 5.50 & 3.30 & 2.75 & 0.2438 \\*
PXDesign & 11.00 & 11.00 & 6.60 & 3.85 & 0.4472 \\[4pt]
\multicolumn{6}{@{}l}{\textbf{EGFR}}\\*
\method & 11.00 & 8.25 & 3.67 & 1.94 & 0.3256 \\*
Claude Science & 11.00 & 8.25 & 4.89 & 2.59 & 0.3288 \\*
Biomni & 11.00 & 8.25 & 4.89 & 2.59 & 0.3069 \\*
STELLA & 5.50 & 5.50 & 3.67 & 1.94 & 0.2199 \\*
Codex & 5.50 & 5.50 & 3.67 & 1.94 & 0.2253 \\*
BoltzGen & 5.50 & 5.50 & 2.44 & 1.94 & 0.2178 \\*
BindCraft & 5.50 & 2.75 & 1.22 & 2.59 & 0.2002 \\*
PXDesign & 5.50 & 5.50 & 2.44 & 1.29 & 0.1643 \\[4pt]
\multicolumn{6}{@{}l}{\textbf{TrkA}}\\*
\method & 11.00 & 11.00 & 9.17 & 6.00 & 0.7168 \\*
Claude Science & 11.00 & 7.33 & 9.17 & 7.00 & 0.7128 \\*
Biomni & 11.00 & 11.00 & 11.00 & 8.00 & 0.8554 \\*
STELLA & 11.00 & 11.00 & 7.33 & 7.00 & 0.7224 \\*
Codex & 11.00 & 7.33 & 7.33 & 8.00 & 0.6912 \\*
BoltzGen & 11.00 & 11.00 & 9.17 & 7.00 & 0.7724 \\*
BindCraft & 5.50 & 7.33 & 7.33 & 6.00 & 0.5599 \\*
PXDesign & 11.00 & 7.33 & 7.33 & 6.00 & 0.5997 \\[4pt]
\multicolumn{6}{@{}l}{\textbf{TIE2}}\\*
\method & 0.00 & 0.00 & 0.00 & 0.00 & 0.0886 \\*
Claude Science & 0.00 & 0.00 & 0.00 & 0.00 & 0.0865 \\*
Biomni & 0.00 & 0.00 & 0.00 & 0.00 & 0.0856 \\*
STELLA & 0.00 & 0.00 & 0.00 & 0.00 & 0.1093 \\*
Codex & 0.00 & 0.00 & 0.00 & 0.00 & 0.0872 \\*
BoltzGen & 0.00 & 0.00 & 0.00 & 0.00 & 0.0850 \\*
BindCraft & 0.00 & 0.00 & 0.00 & 0.00 & 0.0839 \\*
PXDesign & 0.00 & 0.00 & 0.00 & 0.00 & 0.1102 \\[4pt]
\end{longtable}
\endgroup

\textbf{pMHC test results.} \Cref{fig:supp-protein-pmhc} compares test AUPRC for the five agent workflows and the three fixed expert-defined objectives on MART-1. Agent workflows were selected using the gp100 search data and frozen before test evaluation. 
\method achieved the highest AUPRC on MART-1 (0.0172), compared with positive-fraction references of 0.0070 and the second highest AUPRC values of 0.0122 for BindCraft.

\Cref{tab:protein-ef-pmhc} reports all four EF cutoffs and AUPRC. \method achieved the highest EF across all four cutoffs among agent baselines on MART-1. Against the three fixed expert-defined objectives, \method achieved higher AUPRC and the highest EF in three of the four EF cutoffs. These results show broadly improved ranking performance across both precision--recall and enrichment metrics.

\begin{table}[htbp]
\centering\small
\caption{Held-out pMHC test performance. Agent workflows were selected using gp100 data; fixed expert-defined objectives used the same test candidates. EF values are shown at four cutoffs, alongside AUPRC. Values retain the precision of the source export.}
\label{tab:protein-ef-pmhc}
\setlength{\tabcolsep}{5pt}
\begin{tabular}{@{}lrrrrr@{}}
\toprule
\textbf{Method} & \textbf{EF@1}\% & \textbf{EF@2}\% & \textbf{EF@5}\% & \textbf{EF@10}\% & \textbf{AUPRC} \\
\midrule
\method & 3.65 & 5.55 & 2.21 & 2.22 & 0.0172 \\
Claude Science & 0.00 & 1.85 & 0.74 & 1.48 & 0.0089 \\
Biomni & 0.00 & 1.85 & 0.74 & 1.11 & 0.0087 \\
STELLA & 0.00 & 1.85 & 1.48 & 1.48 & 0.0095 \\
Codex & 0.00 & 0.00 & 1.48 & 0.74 & 0.0088 \\
BoltzGen & 0.00 & 1.85 & 0.74 & 0.74 & 0.0067 \\
BindCraft & 7.30 & 3.70 & 1.48 & 1.48 & 0.0122 \\
PXDesign & 0.00 & 0.00 & 0.74 & 0.37 & 0.0065 \\
\bottomrule
\end{tabular}
\end{table}

\textbf{Effect of source task memory.} The pMHC searches with and without miniprotein-binder memory used the same resource library, candidate data, search controller, development objective and evaluation procedure, with a 300 A100 GPU-hour budget and allowance of 20 attempted workflow evaluations. Both conditions used the complete gp100 collections for search and retained memory updates during the current search. The selected workflows were frozen before test evaluation. 

Source-task memory increased MART-1 AUPRC from 0.0125 to 0.0172, a relative improvement of approximately 37\% (\Cref{fig:supp-protein-pmhc}). It increased EF from 0.00 to 3.65 at the 1\% cutoff and from 0.00 to 5.55 at the 2\% cutoff. At the 5\% and 10\% cutoffs, EF increased from 1.48 to 2.21 and from 1.48 to 2.22, respectively (\Cref{tab:protein-memory-results}).

\begin{table}[htbp]
\centering\small
\caption{Effect of source task memory on MART-1 test performance. With and without RSI denote searches with and without prior miniprotein-binder experience.}
\label{tab:protein-memory-results}
\setlength{\tabcolsep}{5pt}
\begin{tabular}{@{}lrrrrr@{}}
\toprule
\textbf{Condition} & \textbf{EF@1\%} & \textbf{EF@2\%} & \textbf{EF@5\%} & \textbf{EF@10\%} & \textbf{AUPRC} \\
\midrule
With RSI & 3.65 & 5.55 & 2.21 & 2.22 & 0.0172 \\
Without RSI & 0.00 & 0.00 & 1.48 & 1.48 & 0.0125 \\
\bottomrule
\end{tabular}
\end{table}

\textbf{Selected pMHC ranking workflow.} \method selected a workflow using ESMFold2-Fast~\citep{candido2026language} and an equal-weight sum of eight terms: \texttt{binder\_target\_contact}, \texttt{helix\_score}, \texttt{binder\_ptm}, \texttt{binder\_target\_iptm}, \texttt{ptm\_energy}, \texttt{ipsae\_min}, \texttt{monomer\_plddt} and \texttt{monomer\_helix}. Each term was oriented so that a larger value represented a more favorable predicted property before aggregation.

\subsection{Binder generation and candidate selection}
\label{app:protein-generation}

\textbf{Design targets.}
\label{app:pmhc-design}
\label{app:pmhc-targets} 
Each prospective design task was specified by an HLA allele, an on-target and off-target peptide pair, source structures for sequence and hotspot provenance, chain assignments, a binder-length range, and exposed peptide hotspots (\Cref{tab:pmhc-design-inputs}). Within each target pair, the HLA heavy-chain and \(\beta_2\)-microglobulin sequences were held fixed and only the peptide sequence was changed. The source PDB coordinates were not supplied as design inputs. They were used only to establish the target-chain sequences and select peptide hotspots. Following previous work~\citep{liu2025design}, hotspots were selected from solvent-exposed, upward-facing peptide residues with relative solvent-accessible surface area (relSASA) of at least 20\% in the binder-free pMHC complex. Individual binder lengths were sampled within the target-specific allowed range. 

\begin{table*}[t]
\centering
\small
\caption{Inputs and target-specific constraints for the two prospective pMHC binder-design tasks. Source chains give the HLA heavy-chain, \(\beta_2\)-microglobulin, and peptide chain identifiers in the corresponding PDB entry. Peptide positions are one-based. Numbers in parentheses are zero-based indices in the concatenated target sequence.}
\label{tab:pmhc-design-inputs}
\begin{tabularx}{\textwidth}{@{}>{\raggedright\arraybackslash}p{4.6cm}>{\raggedright\arraybackslash}X>{\raggedright\arraybackslash}X@{}}
\toprule
\textbf{Input} & \textbf{MART-1} & \textbf{SARS-CoV-1} \\
\midrule
HLA allele &
HLA-A*02:01 & HLA-A*02:01 \\
\addlinespace
On-target peptide &
ELAGIGILTV & GLMWLSYFV \\
\addlinespace
Off-target peptide &
GLMGAGIAQV & GLIFSSYFV \\
\addlinespace
Structure provenance &
PDB 5NHT & PDB 3I6G \\
\addlinespace
Source PDB chains
(HLA heavy chain / \(\beta_2\)m / peptide) &
H/L/P & A/B/C \\
\addlinespace
Binder-length range &
80--120 residues & 80--120 residues \\
\addlinespace
Peptide hotspots &
P4 Gly, P5 Ile, P8 Leu, P9 Thr
(378, 379, 382, 383) &
P4 Trp, P5 Leu, P6 Ser, P8 Phe
(378, 379, 380, 382) \\
\bottomrule
\end{tabularx}
\end{table*}

\textbf{Generation workflow search configuration.} \label{app:pmhc-generation} The available generation methods were RFDiffusion3~\citep{butcher2025novo}, Proteina-Complexa~\citep{didi2026scaling}, PXDesign~\citep{team2025pxdesign}, BoltzGen~\citep{stark2025boltzgen}, and the MOSAIC~\citep{boyd2025mosaic} simplex accelerated proximal gradient method (APGM). The methods differed at the generation stage and entered the same inverse-folding and ranking procedure.

The target-specific searches ran in parallel without exchanging evaluations or search records. Each search was allocated 30 attempted rounds and 1,000 A100 GPU-hours. At each node, \method evaluated a generator configuration by generating 20 backbones and applying SolubleMPNN~\citep{goverde2024computational} to generate eight sequences for each retained backbone, followed by the remaining scoring steps. Cysteine was excluded from the inverse-folding sequence alphabet in all workflows. The ranking score among the top-ranked 5\% of designs defined the configuration's development reward. 

After the 30-round search for each target, the three highest-scoring workflows were each used to generate 8,000 additional designs. The final candidate pool comprised all designs from the search and expansion stages and was then subjected to deduplication, filtering, and ranking. We denote the exported fields \texttt{on\_target\_ranking\_score} and \texttt{off\_target\_ranking\_score} by $s_{\mathrm{on}}$ and $s_{\mathrm{off}}$, respectively. For final \method selection, designs were retained when $\texttt{on\_ipsae\_min}\geq0.50$, $\texttt{monomer\_plddt}\geq0.80$, at least 15 HLA contacts ($\texttt{hla\_res}\geq15$), and $\texttt{s8\_on}-\texttt{s8\_off}>0$. Passing designs were ranked by $\texttt{s8\_on}$.

\subsection{Comparison with reference design workflows}
\label{app:protein-workflow-comparison}

\textbf{Prospective agent comparison.} We selected Claude Science as the prospective baseline because it achieved the highest performance on the miniprotein-binder benchmark and had previously demonstrated autonomous design of experimentally validated protein binders~\citep{claudescience2026binder}. The completed comparison comprised independent target-specific campaigns for MART-1 and SARS-CoV-1. The target and peptide-swapped off-target specifications, binder-length range, design objectives and total computational budget were held fixed, while each system retained its native generation, refinement and ranking procedure. Both systems received 1,000 A100 GPU-hours per target. The 30-round allowance applied to \method's workflow search; Claude Science instead controlled job scheduling within the shared GPU-hour budget.

\textbf{Claude Science run configuration for prospective binder design.}
\label{app:claude-prospective}
We initialized one independent Claude Science session per target, using Claude Opus~5 and an adaptation of Anthropic's multi-target binder-design protocol~\citep{claudescience2026binder}. The prompt source is provided in \Cref{app:protein-baselines}; this section summarizes the study-specific adaptations and run configuration.

We adapted the instructions to peptide-specific pMHC design and the available computing platform. Inputs comprised HLA heavy-chain and \(\beta_2\)-microglobulin sequences and MSAs, on- and off-target peptide sequences, an 80--120-residue binder range, epitope hints and novelty references. Every scoring prediction was required to include the HLA heavy chain, \(\beta_2\)-microglobulin, peptide and binder. Epitope positions were hints, not a mandatory hotspot-contact filter. %

The campaign was given a computational budget of 1,000 A100 GPU-hours per target, with job state recorded in files on a shared volume. SolubleMPNN~\citep{goverde2024computational} and SolubleCaliby~\citep{shuai2025caliby} were named for designs selected for the final computational output, with their base variants permitted during backbone search. The specified pipeline placed checks for sequence and structural novelty, liabilities, monomer foldability, structural plausibility and redundancy before any complex prediction.

The instructions required retained candidates to be scored against the on- and off-target complexes using ESMFold2-Fast, ESMFold2-Full~\citep{candido2026language} and Protenix v2~\citep{zhang2026protenix}. For each scoring model, the fixed criterion combined minimum ipSAE~\citep{dunbrack2025res}, design-to-prediction scDockQ~\citep{basu2016dockq} and the on-target-minus-off-target minimum-ipSAE difference. Scores were standardized within the candidate cohort and combined with relative weights of \(4{:}1{:}4\). The minimum scDockQ across the scoring models used also entered the ranking key at a prespecified pose threshold of 0.23, ordering designs that met it above those that did not. At least two rounds of partial diffusion, prediction followed by redesign, or sequence resampling were required.
The requested output was a ranked set of 500 designs per target, excluding exact and near duplicates and limiting representation from individual design lineages and structural clusters. At least three structure generators had to be represented, with no more than 50\% of designs from one generator or two-thirds from one sequence-design method.

The authors specified the target inputs, scoring criterion, output format, diversity limits and computational budget. Claude Science chose the generation strategy and parameters, sampling allocation, use of epitope hints, monomer and structural-plausibility thresholds, parent selection and refinement operations. The monomer and structural-plausibility thresholds were fixed after selection. The comparison therefore used its native binder-design protocol with pMHC- and platform-specific adaptations.

\stopcontents[supplementary]

\clearpage
\putbib
\end{bibunit}


\begin{thebibliography}{56}
\providecommand{\natexlab}[1]{#1}
\providecommand{\url}[1]{\texttt{#1}}
\expandafter\ifx\csname urlstyle\endcsname\relax
  \providecommand{\doi}[1]{doi: #1}\else
  \providecommand{\doi}{doi: \begingroup \urlstyle{rm}\Url}\fi

\bibitem[Wong et~al.(2024)Wong, Zheng, Valeri, Donghia, Anahtar, Omori, Li, Cubillos-Ruiz, Krishnan, Jin, Manson, Friedrichs, Helbig, Hajian, Fiejtek, Wagner, Soutter, Earl, Stokes, Renner, and Collins]{wong2024discovery}
Felix Wong, Erica~J. Zheng, Jacqueline~A. Valeri, Nina~M. Donghia, Melis~N. Anahtar, Satotaka Omori, Alicia Li, Andres Cubillos-Ruiz, Aarti Krishnan, Wengong Jin, Abigail~L. Manson, Jens Friedrichs, Ralf Helbig, Behnoush Hajian, Dawid~K. Fiejtek, Florence~F. Wagner, Holly~H. Soutter, Ashlee~M. Earl, Jonathan~M. Stokes, Lars~D. Renner, and James~J. Collins.
\newblock Discovery of a structural class of antibiotics with explainable deep learning.
\newblock \emph{Nature}, 626\penalty0 (7997):\penalty0 177--185, 2024.
\newblock \doi{10.1038/s41586-023-06887-8}.
\newblock URL \url{https://doi.org/10.1038/s41586-023-06887-8}.

\bibitem[Liu et~al.(2025)Liu, Greenwood, Bonzanini, Motmaen, Meyerberg, Dao, Xiang, Ault, Sharp, Wang, Visani, Vafeados, Roullier, Nourmohammad, Scheinberg, Garcia, and Baker]{liu2025design}
Bingxu Liu, Nathan~F. Greenwood, Julia~E. Bonzanini, Amir Motmaen, Jeremy Meyerberg, Tao Dao, Xinyu Xiang, Russell Ault, Jazmin Sharp, Chunyu Wang, Gian~Marco Visani, Dionne~K. Vafeados, Nicole Roullier, Armita Nourmohammad, David~A. Scheinberg, K.~Christopher Garcia, and David Baker.
\newblock Design of high-specificity binders for peptide–mhc-i complexes.
\newblock \emph{Science}, 389\penalty0 (6758):\penalty0 386--391, 2025.
\newblock \doi{10.1126/science.adv0185}.
\newblock URL \url{https://doi.org/10.1126/science.adv0185}.

\bibitem[Bennett et~al.(2023)Bennett, Coventry, Goreshnik, Huang, Allen, Vafeados, Peng, Dauparas, Baek, Stewart, DiMaio, De~Munck, Savvides, and Baker]{bennett2023improving}
Nathaniel~R. Bennett, Brian Coventry, Inna Goreshnik, Buwei Huang, Aza Allen, Dionne Vafeados, Ying~Po Peng, Justas Dauparas, Minkyung Baek, Lance Stewart, Frank DiMaio, Steven De~Munck, Savvas~N. Savvides, and David Baker.
\newblock Improving de novo protein binder design with deep learning.
\newblock \emph{Nature Communications}, 14\penalty0 (1):\penalty0 2625, 2023.
\newblock \doi{10.1038/s41467-023-38328-5}.
\newblock URL \url{https://doi.org/10.1038/s41467-023-38328-5}.

\bibitem[Dauparas et~al.(2022)Dauparas, Anishchenko, Bennett, Bai, Ragotte, Milles, Wicky, Courbet, de~Haas, Bethel, Leung, Huddy, Pellock, Tischer, Chan, Koepnick, Nguyen, Kang, Sankaran, Bera, King, and Baker]{dauparas2022robust}
J.~Dauparas, I.~Anishchenko, N.~Bennett, H.~Bai, R.~J. Ragotte, L.~F. Milles, B.~I.~M. Wicky, A.~Courbet, R.~J. de~Haas, N.~Bethel, P.~J.~Y. Leung, T.~F. Huddy, S.~Pellock, D.~Tischer, F.~Chan, B.~Koepnick, H.~Nguyen, A.~Kang, B.~Sankaran, A.~K. Bera, N.~P. King, and D.~Baker.
\newblock Robust deep learning–based protein sequence design using proteinmpnn.
\newblock \emph{Science}, 378\penalty0 (6615):\penalty0 49--56, 2022.
\newblock \doi{10.1126/science.add2187}.
\newblock URL \url{https://doi.org/10.1126/science.add2187}.

\bibitem[Watson et~al.(2023)Watson, Juergens, Bennett, Trippe, Yim, Eisenach, Ahern, Borst, Ragotte, Milles, Wicky, Hanikel, Pellock, Courbet, Sheffler, Wang, Venkatesh, Sappington, Torres, Lauko, De~Bortoli, Mathieu, Ovchinnikov, Barzilay, Jaakkola, DiMaio, Baek, and Baker]{watson2023novo}
Joseph~L. Watson, David Juergens, Nathaniel~R. Bennett, Brian~L. Trippe, Jason Yim, Helen~E. Eisenach, Woody Ahern, Andrew~J. Borst, Robert~J. Ragotte, Lukas~F. Milles, Basile I.~M. Wicky, Nikita Hanikel, Samuel~J. Pellock, Alexis Courbet, William Sheffler, Jue Wang, Preetham Venkatesh, Isaac Sappington, Susana~Vázquez Torres, Anna Lauko, Valentin De~Bortoli, Emile Mathieu, Sergey Ovchinnikov, Regina Barzilay, Tommi~S. Jaakkola, Frank DiMaio, Minkyung Baek, and David Baker.
\newblock De novo design of protein structure and function with rfdiffusion.
\newblock \emph{Nature}, 620\penalty0 (7976):\penalty0 1089--1100, 2023.
\newblock \doi{10.1038/s41586-023-06415-8}.
\newblock URL \url{https://doi.org/10.1038/s41586-023-06415-8}.

\bibitem[Yang et~al.(2019)Yang, Swanson, Jin, Coley, Eiden, Gao, Guzman-Perez, Hopper, Kelley, Mathea, Palmer, Settels, Jaakkola, Jensen, and Barzilay]{yang2019analyzing}
Kevin Yang, Kyle Swanson, Wengong Jin, Connor Coley, Philipp Eiden, Hua Gao, Angel Guzman-Perez, Timothy Hopper, Brian Kelley, Miriam Mathea, Andrew Palmer, Volker Settels, Tommi Jaakkola, Klavs Jensen, and Regina Barzilay.
\newblock Analyzing learned molecular representations for property prediction.
\newblock \emph{Journal of Chemical Information and Modeling}, 59\penalty0 (8):\penalty0 3370--3388, 2019.
\newblock \doi{10.1021/acs.jcim.9b00237}.
\newblock URL \url{https://pubs.acs.org/doi/10.1021/acs.jcim.9b00237}.

\bibitem[Wu et~al.(2018)Wu, Ramsundar, Feinberg, Gomes, Geniesse, Pappu, Leswing, and Pande]{wu2018moleculenet}
Zhenqin Wu, Bharath Ramsundar, Evan~N. Feinberg, Joseph Gomes, Caleb Geniesse, Aneesh~S. Pappu, Karl Leswing, and Vijay Pande.
\newblock {MoleculeNet}: a benchmark for molecular machine learning.
\newblock \emph{Chemical Science}, 9\penalty0 (2):\penalty0 513--530, 2018.
\newblock \doi{10.1039/C7SC02664A}.
\newblock URL \url{https://doi.org/10.1039/C7SC02664A}.

\bibitem[Liu et~al.(2023)Liu, Catacutan, Rathod, Swanson, Jin, Mohammed, Chiappino-Pepe, Syed, Fragis, Rachwalski, Magolan, Surette, Coombes, Jaakkola, Barzilay, Collins, and Stokes]{Liu2023Abaucin}
Gary Liu, Denise~B. Catacutan, Khushi Rathod, Kyle Swanson, Wengong Jin, Jody~C. Mohammed, Anush Chiappino-Pepe, Saad~A. Syed, Meghan Fragis, Kenneth Rachwalski, Jakob Magolan, Michael~G. Surette, Brian~K. Coombes, Tommi Jaakkola, Regina Barzilay, James~J. Collins, and Jonathan~M. Stokes.
\newblock {Deep learning-guided discovery of an antibiotic targeting Acinetobacter baumannii}.
\newblock \emph{Nature Chemical Biology}, 19\penalty0 (11):\penalty0 1342--1350, 2023.
\newblock \doi{10.1038/s41589-023-01349-8}.
\newblock URL \url{https://www.nature.com/articles/s41589-023-01349-8}.

\bibitem[Stokes et~al.(2020)Stokes, Yang, Swanson, Jin, Cubillos-Ruiz, Donghia, MacNair, French, Carfrae, Bloom-Ackermann, Tran, Chiappino-Pepe, Badran, Andrews, Chory, Church, Brown, Jaakkola, Barzilay, and Collins]{Stokes2020AntibioticDiscovery}
Jonathan~M. Stokes, Kevin Yang, Kyle Swanson, Wengong Jin, Andres Cubillos-Ruiz, Nina~M. Donghia, Craig~R. MacNair, Shawn French, Lindsey~A. Carfrae, Zohar Bloom-Ackermann, Victoria~M. Tran, Anush Chiappino-Pepe, Ahmed~H. Badran, Ian~W. Andrews, Emma~J. Chory, George~M. Church, Eric~D. Brown, Tommi~S. Jaakkola, Regina Barzilay, and James~J. Collins.
\newblock {A Deep Learning Approach to Antibiotic Discovery}.
\newblock \emph{Cell}, 180\penalty0 (4):\penalty0 688--702.e13, 2020.
\newblock \doi{10.1016/j.cell.2020.01.021}.
\newblock URL \url{https://doi.org/10.1016/j.cell.2020.01.021}.

\bibitem[Feurer et~al.(2015)Feurer, Klein, Eggensperger, Springenberg, Blum, and Hutter]{feurer2015autosklearn}
Matthias Feurer, Aaron Klein, Katharina Eggensperger, Jost~Tobias Springenberg, Manuel Blum, and Frank Hutter.
\newblock Efficient and robust automated machine learning.
\newblock In \emph{Advances in Neural Information Processing Systems}, volume~28, 2015.
\newblock URL \url{https://papers.nips.cc/paper_files/paper/2015/hash/11d0e6287202fced83f79975ec59a3a6-Abstract.html}.

\bibitem[Zhang et~al.(2025)Zhang, Xiang, Yu, Teng, Chen, Chen, Zhuge, Cheng, Hong, Wang, Zheng, Liu, Luo, and Wu]{zhang2025aflow}
Jiayi Zhang, Jinyu Xiang, Zhaoyang Yu, Fengwei Teng, Xiong-Hui Chen, Jiaqi Chen, Mingchen Zhuge, Xin Cheng, Sirui Hong, Jinlin Wang, Bingnan Zheng, Bang Liu, Yuyu Luo, and Chenglin Wu.
\newblock {AFlow}: Automating agentic workflow generation.
\newblock In \emph{The Thirteenth International Conference on Learning Representations}, 2025.
\newblock URL \url{https://openreview.net/forum?id=z5uVAKwmjf}.

\bibitem[Feurer et~al.(2022)Feurer, Eggensperger, Falkner, Lindauer, and Hutter]{feurer2022autosklearn2}
Matthias Feurer, Katharina Eggensperger, Stefan Falkner, Marius Lindauer, and Frank Hutter.
\newblock {Auto-Sklearn 2.0}: Hands-free {AutoML} via meta-learning.
\newblock \emph{Journal of Machine Learning Research}, 23\penalty0 (261):\penalty0 1--61, 2022.
\newblock URL \url{https://jmlr.org/papers/v23/21-0992.html}.

\bibitem[Toledo et~al.(2025)Toledo, Hambardzumyan, Josifoski, Hazra, Baldwin, Audran-Reiss, Kuchnik, Magka, Jiang, Lupidi, Lupu, Raileanu, Shavrina, Niu, Gagnon-Audet, Shvartsman, Sodhani, Miller, Charnalia, Dunfield, Wu, Stenetorp, Cancedda, Foerster, and Bachrach]{Toledo2025AIRA}
Edan Toledo, Karen Hambardzumyan, Martin Josifoski, Rishi Hazra, Nicolas Baldwin, Alexis Audran-Reiss, Michael Kuchnik, Despoina Magka, Minqi Jiang, Alisia~Maria Lupidi, Andrei Lupu, Roberta Raileanu, Tatiana Shavrina, Kelvin Niu, Jean-Christophe Gagnon-Audet, Michael Shvartsman, Shagun Sodhani, Alexander~H. Miller, Abhishek Charnalia, Derek Dunfield, Carole-Jean Wu, Pontus Stenetorp, Nicola Cancedda, Jakob~Nicolaus Foerster, and Yoram Bachrach.
\newblock {AI Research Agents for Machine Learning: Search, Exploration, and Generalization in MLE-bench}.
\newblock In \emph{Advances in Neural Information Processing Systems}, volume~38, pages 39717--39756, 2025.
\newblock \doi{10.52202/085713-1186}.
\newblock URL \url{https://proceedings.neurips.cc/paper_files/paper/2025/hash/328b81881da145412f2bc56c998dfb6a-Abstract-Conference.html}.

\bibitem[M.~Bran et~al.(2024)M.~Bran, Cox, Schilter, Baldassari, White, and Schwaller]{m2024augmenting}
Andres M.~Bran, Sam Cox, Oliver Schilter, Carlo Baldassari, Andrew~D. White, and Philippe Schwaller.
\newblock Augmenting large language models with chemistry tools.
\newblock \emph{Nature Machine Intelligence}, 6\penalty0 (5):\penalty0 525--535, 2024.
\newblock \doi{10.1038/s42256-024-00832-8}.
\newblock URL \url{https://doi.org/10.1038/s42256-024-00832-8}.

\bibitem[Swanson et~al.(2025)Swanson, Wu, Bulaong, Pak, and Zou]{swanson2025virtual}
Kyle Swanson, Wesley Wu, Nash~L. Bulaong, John~E. Pak, and James Zou.
\newblock The {Virtual Lab} of {AI} agents designs new {SARS-CoV-2} nanobodies.
\newblock \emph{Nature}, 646\penalty0 (8085):\penalty0 716--723, 2025.
\newblock \doi{10.1038/s41586-025-09442-9}.
\newblock URL \url{https://www.nature.com/articles/s41586-025-09442-9}.

\bibitem[Miao et~al.(2026)Miao, Davis, Zhang, Pritchard, and Zou]{miao2026paper2agent}
Jiacheng Miao, Joe~R. Davis, Yaohui Zhang, Jonathan~K. Pritchard, and James Zou.
\newblock Reimagining research papers as interactive and reliable {AI} agents.
\newblock \emph{Nature}, 658\penalty0 (8138):\penalty0 1272--1280, 2026.
\newblock \doi{10.1038/s41586-026-11044-y}.
\newblock URL \url{https://www.nature.com/articles/s41586-026-11044-y}.

\bibitem[Ayg{\"u}n et~al.(2026)Ayg{\"u}n, Belyaeva, Comanici, Coram, Cui, Garrison, Johnston, Kast, McLean, Norgaard, Shamsi, Smalling, Thompson, Venugopalan, Williams, He, Martinson, Plomecka, Wei, Zhou, Zhu, Abraham, Brand, Bulanova, Cardille, Co, Ellsworth, Joseph, Kane, Krueger, Kartiwa, Liebling, Lueckmann, Raccuglia, Wang, Chou, Manyika, Matias, Platt, Dorfman, Mourad, and Brenner]{aygun2026empirical}
Eser Ayg{\"u}n, Anastasiya Belyaeva, Gheorghe Comanici, Marc Coram, Hao Cui, Jake Garrison, Renee Johnston, Anton Kast, Cory~Y. McLean, Peter Norgaard, Zahra Shamsi, David Smalling, James Thompson, Subhashini Venugopalan, Brian~P. Williams, Chujun He, Sarah Martinson, Martyna Plomecka, Lai Wei, Yuchen Zhou, Qian-Ze Zhu, Matthew Abraham, Erica Brand, Anna Bulanova, Jeffrey~A. Cardille, Chris Co, Scott Ellsworth, Grace Joseph, Malcolm Kane, Ryan Krueger, Johan Kartiwa, Dan Liebling, Jan-Matthis Lueckmann, Paul Raccuglia, Xuefei~Julie Wang, Katherine Chou, James Manyika, Yossi Matias, John~C. Platt, Lizzie Dorfman, Shibl Mourad, and Michael~P. Brenner.
\newblock An {AI} system to help scientists write expert-level empirical software.
\newblock \emph{Nature}, 654\penalty0 (8120):\penalty0 909--916, 2026.
\newblock \doi{10.1038/s41586-026-10658-6}.

\bibitem[Lu et~al.(2026)Lu, Lu, Lange, Yamada, Hu, Foerster, Ha, and Clune]{Lu2026AIScientist}
Chris Lu, Cong Lu, Robert~Tjarko Lange, Yutaro Yamada, Shengran Hu, Jakob Foerster, David Ha, and Jeff Clune.
\newblock {Towards end-to-end automation of AI research}.
\newblock \emph{Nature}, 651\penalty0 (8107):\penalty0 914--919, 2026.
\newblock \doi{10.1038/s41586-026-10265-5}.
\newblock URL \url{https://www.nature.com/articles/s41586-026-10265-5}.

\bibitem[Schmidgall et~al.(2025)Schmidgall, Su, Wang, Sun, Wu, Yu, Liu, Moor, Liu, and Barsoum]{Schmidgall2025AgentLaboratory}
Samuel Schmidgall, Yusheng Su, Ze~Wang, Ximeng Sun, Jialian Wu, Xiaodong Yu, Jiang Liu, Michael Moor, Zicheng Liu, and Emad Barsoum.
\newblock {Agent Laboratory: Using LLM Agents as Research Assistants}.
\newblock In \emph{Findings of the Association for Computational Linguistics: EMNLP 2025}, pages 5977--6043, Suzhou, China, November 2025. Association for Computational Linguistics.
\newblock \doi{10.18653/v1/2025.findings-emnlp.320}.
\newblock URL \url{https://aclanthology.org/2025.findings-emnlp.320/}.

\bibitem[Martinek et~al.(2026)Martinek, Gariboldi, Tzimotoudis, Galea, Zacharopoulou, Alberdi~Escudero, Blake, Čechák, Cassar, Balestrucci, and Alexiou]{Martinek2026Agentomics}
Vlastimil Martinek, Andrea Gariboldi, Dimosthenis Tzimotoudis, Mark Galea, Elissavet Zacharopoulou, Aitor Alberdi~Escudero, Edward Blake, David Čechák, Luke Cassar, Alessandro Balestrucci, and Panagiotis Alexiou.
\newblock {Agentomics: an agentic system that autonomously develops novel state-of-the-art solutions for biomedical machine learning tasks}.
\newblock \emph{Bioinformatics}, 42\penalty0 (Supplement 1):\penalty0 btag250, 2026.
\newblock \doi{10.1093/bioinformatics/btag250}.
\newblock URL \url{https://academic.oup.com/bioinformatics/article/42/Supplement_1/btag250/8726289}.

\bibitem[Huang et~al.(2026)Huang, Zhang, Wang, Qu, Lu, Li, Roohani, Qiu, Cao, Li, Zhang, Yin, Wierenga, Kavi, Liu, She, Marwaha, Carter, Zhou, Wheeler, Bernstein, Wang, He, Zhou, Snyder, Cong, Regev, and Leskovec]{Huang2026BiomniScience}
Kexin Huang, Serena Zhang, Hanchen Wang, Yuanhao Qu, Yingzhou Lu, Ryan Li, Yusuf Roohani, Lin Qiu, Shiyi Cao, Gavin Li, Junze Zhang, Di~Yin, Rick Wierenga, Deniz Kavi, Sherry Liu, Tianwei She, Shruti Marwaha, Jennefer~N. Carter, Xin Zhou, Matthew~T. Wheeler, Jonathan~A. Bernstein, Mengdi Wang, Peng He, Jingtian Zhou, Michael~P. Snyder, Le~Cong, Aviv Regev, and Jure Leskovec.
\newblock {Autonomous biomedical research with an artificial intelligence agent}.
\newblock \emph{Science}, 393\penalty0 (6813):\penalty0 eadz4351, 2026.
\newblock \doi{10.1126/science.adz4351}.
\newblock URL \url{https://www.science.org/doi/10.1126/science.adz4351}.

\bibitem[{Claude Science} and Shanehsazzadeh(2026)]{claudescience2026binder}
{Claude Science} and Amir Shanehsazzadeh.
\newblock Autonomous de novo protein binder design with claude.
\newblock Technical report, Anthropic, aug 2026.
\newblock URL \url{https://www-cdn.anthropic.com/30bf50e22a01388bb29bf077ee3f244531594b7a.pdf}.

\bibitem[{Claude Science} et~al.(2026){Claude Science}, Horwitz, van Geelen, Kajderowicz, and Shanehsazzadeh]{claudescience2026scorpion}
{Claude Science}, Nathaniel Horwitz, James van Geelen, Kat Kajderowicz, and Amir Shanehsazzadeh.
\newblock De novo design of protein binders against scorpion toxins with {Claude}.
\newblock Technical report, Anthropic, oct 2026.
\newblock URL \url{https://www-cdn.anthropic.com/31ac578806cffd8b3d92d8affe6edc4c04d502e1.pdf}.

\bibitem[Fang et~al.(2026)Fang, Liang, Wang, Wu, Qiao, Xie, Huang, Chen, and Zhang]{fang2025memp}
Runnan Fang, Yuan Liang, Xiaobin Wang, Jialong Wu, Shuofei Qiao, Pengjun Xie, Fei Huang, Huajun Chen, and Ningyu Zhang.
\newblock Memp: Exploring agent procedural memory.
\newblock In \emph{Findings of the Association for Computational Linguistics: ACL 2026}, pages 17490--17502, San Diego, California, United States, jul 2026. Association for Computational Linguistics.
\newblock \doi{10.18653/v1/2026.findings-acl.866}.
\newblock URL \url{https://aclanthology.org/2026.findings-acl.866/}.

\bibitem[Kim et~al.(2026)Kim, Kang, Kim, Yang, Ren, and Hwang]{kim2026memory}
Kangsan Kim, Minki Kang, Taeil Kim, Yanlai Yang, Mengye Ren, and Sung~Ju Hwang.
\newblock Memory transfer learning: How memories are transferred across domains in coding agents.
\newblock In \emph{Proceedings of the 2026 Conference on Empirical Methods in Natural Language Processing}, pages 60677--60697, Budapest, Hungary, oct 2026. Association for Computational Linguistics.
\newblock URL \url{https://aclanthology.org/2026.emnlp-main.2666/}.

\bibitem[Kocsis and Szepesv{\'a}ri(2006)]{kocsis2006bandit}
Levente Kocsis and Csaba Szepesv{\'a}ri.
\newblock Bandit based {Monte-Carlo} planning.
\newblock In \emph{Machine Learning: {ECML} 2006}, volume 4212 of \emph{Lecture Notes in Computer Science}, pages 282--293. Springer, 2006.
\newblock \doi{10.1007/11871842_29}.
\newblock URL \url{https://link.springer.com/chapter/10.1007/11871842_29}.

\bibitem[{GBD 2021 Antimicrobial Resistance Collaborators}(2024)]{gbd2024amr}
{GBD 2021 Antimicrobial Resistance Collaborators}.
\newblock Global burden of bacterial antimicrobial resistance 1990--2021: a systematic analysis with forecasts to 2050.
\newblock \emph{The Lancet}, 404\penalty0 (10459):\penalty0 1199--1226, 2024.
\newblock \doi{10.1016/S0140-6736(24)01867-1}.
\newblock URL \url{https://doi.org/10.1016/S0140-6736(24)01867-1}.

\bibitem[{Anthropic}(2025)]{anthropic_claude_code}
{Anthropic}.
\newblock Claude code documentation.
\newblock \url{https://code.claude.com/docs/en/overview}, 2025.
\newblock URL \url{https://code.claude.com/docs/en/overview}.
\newblock Accessed: 2026-10-28.

\bibitem[Huang et~al.(2021)Huang, Fu, Gao, Zhao, Roohani, Leskovec, Coley, Xiao, Sun, and Zitnik]{huang2021therapeutics}
Kexin Huang, Tianfan Fu, Wenhao Gao, Yue Zhao, Yusuf Roohani, Jure Leskovec, Connor~W Coley, Cao Xiao, Jimeng Sun, and Marinka Zitnik.
\newblock Therapeutics data commons: Machine learning datasets and tasks for drug discovery and development.
\newblock In \emph{Proceedings of the Neural Information Processing Systems Track on Datasets and Benchmarks}, volume~1. Neural information processing systems foundation, 2021.
\newblock URL \url{https://datasets-benchmarks-proceedings.neurips.cc/paper/2021/hash/4c56ff4ce4aaf9573aa5dff913df997a-Abstract-round1.html}.

\bibitem[van~der Maaten and Hinton(2008)]{vandermaaten2008visualizing}
Laurens van~der Maaten and Geoffrey Hinton.
\newblock Visualizing data using {t-SNE}.
\newblock \emph{Journal of Machine Learning Research}, 9\penalty0 (86):\penalty0 2579--2605, 2008.
\newblock URL \url{https://jmlr.org/papers/v9/vandermaaten08a.html}.

\bibitem[Nathan et~al.(2021)Nathan, Hassel, Rutkowski, Baurain, Butler, Schlaak, Sullivan, Ochsenreither, Dummer, Kirkwood, Joshua, Sacco, Shoushtari, Orloff, Piulats, Milhem, Salama, Curti, Demidov, Gastaud, Mauch, Yushak, Carvajal, Hamid, Abdullah, Holland, Goodall, Piperno-Neumann, and {IMCgp100-202 Investigators}]{Nathan2021Tebentafusp}
Paul Nathan, Jessica~C. Hassel, Piotr Rutkowski, Jean-Fran{\c{c}}ois Baurain, Marcus~O. Butler, Max Schlaak, Ryan~J. Sullivan, Sebastian Ochsenreither, Reinhard Dummer, John~M. Kirkwood, Anthony~M. Joshua, Joseph~J. Sacco, Alexander~N. Shoushtari, Marlana Orloff, Josep~M. Piulats, Mohammed Milhem, April K.~S. Salama, Brendan Curti, Lev Demidov, Lauris Gastaud, Cornelia Mauch, Melinda Yushak, Richard~D. Carvajal, Omid Hamid, Shaad~E. Abdullah, Chris Holland, Howard Goodall, Sophie Piperno-Neumann, and {IMCgp100-202 Investigators}.
\newblock Overall survival benefit with tebentafusp in metastatic uveal melanoma.
\newblock \emph{New England Journal of Medicine}, 385\penalty0 (13):\penalty0 1196--1206, 2021.
\newblock \doi{10.1056/NEJMoa2103485}.
\newblock URL \url{https://doi.org/10.1056/NEJMoa2103485}.

\bibitem[Jumper et~al.(2021)Jumper, Evans, Pritzel, Green, Figurnov, Ronneberger, Tunyasuvunakool, Bates, Žídek, Potapenko, Bridgland, Meyer, Kohl, Ballard, Cowie, Romera-Paredes, Nikolov, Jain, Adler, Back, Petersen, Reiman, Clancy, Zielinski, Steinegger, Pacholska, Berghammer, Bodenstein, Silver, Vinyals, Senior, Kavukcuoglu, Kohli, and Hassabis]{jumper2021highly}
John Jumper, Richard Evans, Alexander Pritzel, Tim Green, Michael Figurnov, Olaf Ronneberger, Kathryn Tunyasuvunakool, Russ Bates, Augustin Žídek, Anna Potapenko, Alex Bridgland, Clemens Meyer, Simon A.~A. Kohl, Andrew~J. Ballard, Andrew Cowie, Bernardino Romera-Paredes, Stanislav Nikolov, Rishub Jain, Jonas Adler, Trevor Back, Stig Petersen, David Reiman, Ellen Clancy, Michal Zielinski, Martin Steinegger, Michalina Pacholska, Tamas Berghammer, Sebastian Bodenstein, David Silver, Oriol Vinyals, Andrew~W. Senior, Koray Kavukcuoglu, Pushmeet Kohli, and Demis Hassabis.
\newblock Highly accurate protein structure prediction with alphafold.
\newblock \emph{Nature}, 596\penalty0 (7873):\penalty0 583--589, 2021.
\newblock \doi{10.1038/s41586-021-03819-2}.
\newblock URL \url{https://doi.org/10.1038/s41586-021-03819-2}.

\bibitem[Motmaen et~al.(2023)Motmaen, Dauparas, Baek, Abedi, Baker, and Bradley]{motmaen2023peptide}
Amir Motmaen, Justas Dauparas, Minkyung Baek, Mohamad~H. Abedi, David Baker, and Philip Bradley.
\newblock Peptide-binding specificity prediction using fine-tuned protein structure prediction networks.
\newblock \emph{Proceedings of the National Academy of Sciences}, 120\penalty0 (9):\penalty0 e2216697120, 2023.
\newblock \doi{10.1073/pnas.2216697120}.
\newblock URL \url{https://doi.org/10.1073/pnas.2216697120}.

\bibitem[{Anthropic}(2026)]{anthropic_claude_science}
{Anthropic}.
\newblock Claude science.
\newblock \url{https://claude.com/product/claude-science}, 2026.
\newblock URL \url{https://claude.com/product/claude-science}.
\newblock Accessed: 2026-10-28.

\bibitem[Jin et~al.(2025)Jin, Zhang, Wang, and Cong]{jin2025stella}
Ruofan Jin, Zaixi Zhang, Mengdi Wang, and Le~Cong.
\newblock Stella: Self-evolving llm agent for biomedical research.
\newblock \emph{arXiv preprint arXiv:2507.02004}, 2025.
\newblock \doi{10.48550/arXiv.2507.02004}.
\newblock URL \url{https://arxiv.org/abs/2507.02004}.

\bibitem[{OpenAI}(2025)]{openai_codex}
{OpenAI}.
\newblock Codex is now generally available.
\newblock \url{https://openai.com/index/codex-now-generally-available/}, 2025.
\newblock Published October 6, 2025.

\bibitem[Cao et~al.(2020)Cao, Goreshnik, Coventry, Case, Miller, Kozodoy, Chen, Carter, Walls, Park, Strauch, Stewart, Diamond, Veesler, and Baker]{cao2020novo}
Longxing Cao, Inna Goreshnik, Brian Coventry, James~Brett Case, Lauren Miller, Lisa Kozodoy, Rita~E. Chen, Lauren Carter, Alexandra~C. Walls, Young-Jun Park, Eva-Maria Strauch, Lance Stewart, Michael~S. Diamond, David Veesler, and David Baker.
\newblock De novo design of picomolar sars-cov-2 miniprotein inhibitors.
\newblock \emph{Science}, 370\penalty0 (6515):\penalty0 426--431, 2020.
\newblock \doi{10.1126/science.abd9909}.
\newblock URL \url{https://doi.org/10.1126/science.abd9909}.

\bibitem[Cao et~al.(2022)Cao, Coventry, Goreshnik, Huang, Sheffler, Park, Jude, Marković, Kadam, Verschueren, Verstraete, Walsh, Bennett, Phal, Yang, Kozodoy, DeWitt, Picton, Miller, Strauch, DeBouver, Pires, Bera, Halabiya, Hammerson, Yang, Bernard, Stewart, Wilson, Ruohola-Baker, Schlessinger, Lee, Savvides, Garcia, and Baker]{cao2022design}
Longxing Cao, Brian Coventry, Inna Goreshnik, Buwei Huang, William Sheffler, Joon~Sung Park, Kevin~M. Jude, Iva Marković, Rameshwar~U. Kadam, Koen H.~G. Verschueren, Kenneth Verstraete, Scott Thomas~Russell Walsh, Nathaniel Bennett, Ashish Phal, Aerin Yang, Lisa Kozodoy, Michelle DeWitt, Lora Picton, Lauren Miller, Eva-Maria Strauch, Nicholas~D. DeBouver, Allison Pires, Asim~K. Bera, Samer Halabiya, Bradley Hammerson, Wei Yang, Steffen Bernard, Lance Stewart, Ian~A. Wilson, Hannele Ruohola-Baker, Joseph Schlessinger, Sangwon Lee, Savvas~N. Savvides, K.~Christopher Garcia, and David Baker.
\newblock Design of protein-binding proteins from the target structure alone.
\newblock \emph{Nature}, 605\penalty0 (7910):\penalty0 551--560, 2022.
\newblock \doi{10.1038/s41586-022-04654-9}.
\newblock URL \url{https://doi.org/10.1038/s41586-022-04654-9}.

\bibitem[Dunbrack(2025)]{dunbrack2025res}
Roland~L. Dunbrack, Jr.
\newblock Rēs ipsae loquuntur: What's wrong with alphafold's iptm score and how to fix it.
\newblock \emph{bioRxiv}, 2025.
\newblock \doi{10.1101/2025.02.10.637595}.
\newblock URL \url{https://www.biorxiv.org/content/10.1101/2025.02.10.637595v2}.
\newblock Preprint; version 2 posted 2025-12-15.

\bibitem[McInnes et~al.(2018)McInnes, Healy, Saul, and Gro{\ss}berger]{mcinnes2018umap}
Leland McInnes, John Healy, Nathaniel Saul, and Lukas Gro{\ss}berger.
\newblock {UMAP}: Uniform manifold approximation and projection.
\newblock \emph{Journal of Open Source Software}, 3\penalty0 (29):\penalty0 861, 2018.
\newblock \doi{10.21105/joss.00861}.
\newblock URL \url{https://joss.theoj.org/papers/10.21105/joss.00861}.

\bibitem[Zhang and Skolnick(2005)]{zhang2005tmalign}
Yang Zhang and Jeffrey Skolnick.
\newblock {TM-align}: a protein structure alignment algorithm based on the {TM-score}.
\newblock \emph{Nucleic Acids Research}, 33\penalty0 (7):\penalty0 2302--2309, 2005.
\newblock \doi{10.1093/nar/gki524}.
\newblock URL \url{https://doi.org/10.1093/nar/gki524}.

\bibitem[Baell and Holloway(2010)]{baell2010pains}
Jonathan~B. Baell and Georgina~A. Holloway.
\newblock New substructure filters for removal of pan assay interference compounds ({PAINS}) from screening libraries and for their exclusion in bioassays.
\newblock \emph{Journal of Medicinal Chemistry}, 53\penalty0 (7):\penalty0 2719--2740, 2010.
\newblock \doi{10.1021/jm901137j}.
\newblock URL \url{https://pubs.acs.org/doi/10.1021/jm901137j}.

\bibitem[Brenk et~al.(2008)Brenk, Schipani, James, Krasowski, Gilbert, Frearson, and Wyatt]{brenk2008lessons}
Ruth Brenk, Alessandro Schipani, Daniel James, Agata Krasowski, Ian~Hugh Gilbert, Julie Frearson, and Paul~Graham Wyatt.
\newblock Lessons learnt from assembling screening libraries for drug discovery for neglected diseases.
\newblock \emph{ChemMedChem}, 3\penalty0 (3):\penalty0 435--444, 2008.
\newblock \doi{10.1002/cmdc.200700139}.
\newblock URL \url{https://chemistry-europe.onlinelibrary.wiley.com/doi/10.1002/cmdc.200700139}.

\bibitem[Kl{\"a}ser et~al.(2024)Kl{\"a}ser, Banaszewski, Maddrell-Mander, McLean, M{\"u}ller, Parviz, Huang, and Fitzgibbon]{klaser2024minimol}
Kerstin Kl{\"a}ser, B{\l}a{\.z}ej Banaszewski, Samuel Maddrell-Mander, Callum McLean, Luis M{\"u}ller, Ali Parviz, Shenyang Huang, and Andrew Fitzgibbon.
\newblock {MiniMol}: A parameter-efficient foundation model for molecular learning.
\newblock In \emph{ICML Workshop on Accessible and Efficient Foundation Models for Biological Discovery}, 2024.
\newblock \doi{10.48550/arXiv.2404.14986}.
\newblock URL \url{https://openreview.net/forum?id=uA8zvXmiot}.
\newblock Workshop paper; arXiv version: 2404.14986.

\bibitem[Rogers and Hahn(2010)]{rogers2010extended}
David Rogers and Mathew Hahn.
\newblock Extended-connectivity fingerprints.
\newblock \emph{Journal of Chemical Information and Modeling}, 50\penalty0 (5):\penalty0 742--754, 2010.
\newblock \doi{10.1021/ci100050t}.
\newblock URL \url{https://pubs.acs.org/doi/10.1021/ci100050t}.

\bibitem[{RDKit contributors}()]{rdkit}
{RDKit contributors}.
\newblock {RDKit}: Open-source cheminformatics.
\newblock \url{https://www.rdkit.org}.
\newblock URL \url{https://www.rdkit.org}.
\newblock Accessed 28 October 2026.

\bibitem[Butina(1999)]{butina1999unsupervised}
Darko Butina.
\newblock Unsupervised data base clustering based on daylight's fingerprint and tanimoto similarity: A fast and automated way to cluster small and large data sets.
\newblock \emph{Journal of Chemical Information and Computer Sciences}, 39\penalty0 (4):\penalty0 747--750, 1999.
\newblock \doi{10.1021/ci9803381}.
\newblock URL \url{https://pubs.acs.org/doi/10.1021/ci9803381}.

\bibitem[Butcher et~al.(2025)Butcher, Krishna, Mitra, Brent, Li, Corley, Kim, Funk, Mathis, Salike, Muraishi, Eisenach, Thompson, Chen, Politanska, Sehgal, Coventry, Zhang, Qiang, Didi, Kazman, DiMaio, and Baker]{butcher2025novo}
Jasper Butcher, Rohith Krishna, Raktim Mitra, Rafael~I. Brent, Yanjing Li, Nathaniel Corley, Paul~T. Kim, Jonathan Funk, Simon Mathis, Saman Salike, Aiko Muraishi, Helen Eisenach, Tuscan~Rock Thompson, Jie Chen, Yuliya Politanska, Enisha Sehgal, Brian Coventry, Odin Zhang, Bo~Qiang, Kieran Didi, Max Kazman, Frank DiMaio, and David Baker.
\newblock De novo design of all-atom biomolecular interactions with rfdiffusion3.
\newblock \emph{bioRxiv}, 2025.
\newblock \doi{10.1101/2025.09.18.676967}.
\newblock URL \url{https://www.biorxiv.org/content/10.1101/2025.09.18.676967v2}.
\newblock Preprint; version 2 posted 2025-11-19.

\bibitem[Didi et~al.(2026)Didi, Zhang, Zhou, Reidenbach, Cao, Cha, Geffner, Dallago, Tang, Bronstein, Steinegger, Kucukbenli, Vahdat, and Kreis]{didi2026scaling}
Kieran Didi, Zuobai Zhang, Guoqing Zhou, Danny Reidenbach, Zhonglin Cao, Sooyoung Cha, Tomas Geffner, Christian Dallago, Jian Tang, Michael Bronstein, Martin Steinegger, Emine Kucukbenli, Arash Vahdat, and Karsten Kreis.
\newblock Scaling atomistic protein binder design with generative pretraining and test-time compute.
\newblock In C.~Vondrick, B.~Hariharan, C.~Raffel, L.~Pinto, D.~Yang, and A.~Faust, editors, \emph{International Conference on Learning Representations}, volume 2026, pages 61056--61109, 2026.
\newblock URL \url{https://proceedings.iclr.cc/paper_files/paper/2026/file/639d992f819c2b40387d4d5170b8ffd7-Paper-Conference.pdf}.

\bibitem[{Protenix Team} et~al.(2025){Protenix Team}, Ren, Sun, Guan, Liu, Gong, Wang, Wang, Cai, Ma, Zhang, Liu, Zhang, Chen, and Xiao]{team2025pxdesign}
{Protenix Team}, Milong Ren, Jinyuan Sun, Jiaqi Guan, Cong Liu, Chengyue Gong, Yuzhe Wang, Lan Wang, Qixu Cai, Wenzhi Ma, Yuxuan Zhang, Zhenyu Liu, Hanyu Zhang, Xinshi Chen, and Wenzhi Xiao.
\newblock Pxdesign: Fast, modular, and accurate de novo design of protein binders.
\newblock \emph{bioRxiv}, 2025.
\newblock \doi{10.1101/2025.08.15.670450}.
\newblock URL \url{https://www.biorxiv.org/content/10.1101/2025.08.15.670450v3}.
\newblock Preprint; version 3 posted 2025-12-17.

\bibitem[Stark et~al.(2025)Stark, Faltings, Choi, Xie, Hur, O’Donnell, Bushuiev, Uçar, Passaro, Mao, Reveiz, Bushuiev, Portnoi, Pluskal, Sivic, Kreis, Vahdat, Ray, Goldstein, Savinov, Hambalek, Gupta, Taquiri-Diaz, Zhang, Snyder, Hatstat, Arada, Kim, Fan, Tackie-Yarboi, Boselli, Schnaider, Liu, Li, Hnisz, Sabatini, DeGrado, Wohlwend, Corso, Barzilay, and Jaakkola]{stark2025boltzgen}
Hannes Stark, Felix Faltings, MinGyu Choi, Yuxin Xie, Eunsu Hur, Timothy O’Donnell, Anton Bushuiev, Talip Uçar, Saro Passaro, Weian Mao, Mateo Reveiz, Roman Bushuiev, Tally Portnoi, Tomáš Pluskal, Josef Sivic, Karsten Kreis, Arash Vahdat, Shamayeeta Ray, Jonathan~T. Goldstein, Andrew Savinov, Jacob~A. Hambalek, Anshika Gupta, Diego~A. Taquiri-Diaz, Yaotian Zhang, Samuel~J. Snyder, A.~Katherine Hatstat, Angelika Arada, Nam~Hyeong Kim, Haoyu Fan, Ethel Tackie-Yarboi, Dylan Boselli, Lee Schnaider, Chang~C. Liu, Gene-Wei Li, Denes Hnisz, David~M. Sabatini, William~F. DeGrado, Jeremy Wohlwend, Gabriele Corso, Regina Barzilay, and Tommi Jaakkola.
\newblock Boltzgen: Toward universal binder design.
\newblock \emph{bioRxiv}, 2025.
\newblock \doi{10.1101/2025.11.20.689494}.
\newblock URL \url{https://www.biorxiv.org/content/10.1101/2025.11.20.689494v2}.
\newblock Preprint; version 2 posted 2026-06-16.

\bibitem[Boyd et~al.(2025)Boyd, Guns, and {Escalante Bio}]{boyd2025mosaic}
N.~Boyd, S.~Guns, and {Escalante Bio}.
\newblock {MOSAIC}.
\newblock \url{https://github.com/escalante-bio/mosaic}, 2025.
\newblock URL \url{https://github.com/escalante-bio/mosaic}.

\bibitem[Goverde et~al.(2024)Goverde, Pacesa, Goldbach, Dornfeld, Balbi, Georgeon, Rosset, Kapoor, Choudhury, Dauparas, Schellhaas, Kozlov, Baker, Ovchinnikov, Vecchio, and Correia]{goverde2024computational}
Casper~A. Goverde, Martin Pacesa, Nicolas Goldbach, Lars~J. Dornfeld, Petra E.~M. Balbi, Sandrine Georgeon, Stéphane Rosset, Srajan Kapoor, Jagrity Choudhury, Justas Dauparas, Christian Schellhaas, Simon Kozlov, David Baker, Sergey Ovchinnikov, Alex~J. Vecchio, and Bruno~E. Correia.
\newblock Computational design of soluble and functional membrane protein analogues.
\newblock \emph{Nature}, 631\penalty0 (8020):\penalty0 449--458, 2024.
\newblock \doi{10.1038/s41586-024-07601-y}.
\newblock URL \url{https://doi.org/10.1038/s41586-024-07601-y}.

\bibitem[Esvelt et~al.(2011)Esvelt, Carlson, and Liu]{esvelt2011system}
Kevin~M Esvelt, Jacob~C Carlson, and David~R Liu.
\newblock A system for the continuous directed evolution of biomolecules.
\newblock \emph{Nature}, 472\penalty0 (7344):\penalty0 499--503, 2011.
\newblock \doi{10.1038/nature09929}.
\newblock URL \url{https://www.nature.com/articles/nature09929}.

\bibitem[Badran et~al.(2016)Badran, Guzov, Huai, Kemp, Vishwanath, Kain, Nance, Evdokimov, Moshiri, Turner, Wang, Malvar, and Liu]{badran2016continuous}
Ahmed~H. Badran, Victor~M. Guzov, Qing Huai, Melissa~M. Kemp, Prashanth Vishwanath, Wendy Kain, Autumn~M. Nance, Artem Evdokimov, Farhad Moshiri, Keith~H. Turner, Ping Wang, Thomas Malvar, and David~R. Liu.
\newblock Continuous evolution of bacillus thuringiensis toxins overcomes insect resistance.
\newblock \emph{Nature}, 533\penalty0 (7601):\penalty0 58--63, 2016.
\newblock \doi{10.1038/nature17938}.
\newblock URL \url{https://www.nature.com/articles/nature17938}.

\bibitem[Styles et~al.(2025)Styles, Pixley, Wei, Basile, Lu, and Dickinson]{styles2025pancs}
Matthew~J Styles, Joshua~A Pixley, Tongyao Wei, Christopher Basile, Shannon~S Lu, and Bryan~C Dickinson.
\newblock Pancs-binders: a rapid, high-throughput binder discovery platform.
\newblock \emph{Nature Methods}, 22\penalty0 (8):\penalty0 1720--1730, 2025.
\newblock \doi{10.1038/s41592-025-02740-0}.
\newblock URL \url{https://www.nature.com/articles/s41592-025-02740-0}.

\end{thebibliography}


\begin{thebibliography}{43}
\providecommand{\natexlab}[1]{#1}
\providecommand{\url}[1]{\texttt{#1}}
\expandafter\ifx\csname urlstyle\endcsname\relax
  \providecommand{\doi}[1]{doi: #1}\else
  \providecommand{\doi}{doi: \begingroup \urlstyle{rm}\Url}\fi

\bibitem[Liu et~al.(2022)Liu, Tam, Muqeeth, Mohta, Huang, Bansal, and Raffel]{liu2022fewshot}
Haokun Liu, Derek Tam, Mohammed Muqeeth, Jay Mohta, Tenghao Huang, Mohit Bansal, and Colin~A. Raffel.
\newblock Few-shot parameter-efficient fine-tuning is better and cheaper than in-context learning.
\newblock In \emph{Advances in Neural Information Processing Systems}, volume~35, pages 1950--1965, 2022.
\newblock \doi{10.52202/068431-0142}.
\newblock URL \url{https://proceedings.neurips.cc/paper_files/paper/2022/hash/0cde695b83bd186c1fd456302888454c-Abstract-Conference.html}.

\bibitem[Auer et~al.(2002)Auer, Cesa-Bianchi, and Fischer]{auer2002finite}
Peter Auer, Nicolò Cesa-Bianchi, and Paul Fischer.
\newblock Finite-time analysis of the multiarmed bandit problem.
\newblock \emph{Machine Learning}, 47\penalty0 (2-3):\penalty0 235--256, 2002.
\newblock \doi{10.1023/a:1013689704352}.
\newblock URL \url{https://doi.org/10.1023/a:1013689704352}.

\bibitem[Huang et~al.(2026)Huang, Zhang, Wang, Qu, Lu, Li, Roohani, Qiu, Cao, Li, Zhang, Yin, Wierenga, Kavi, Liu, She, Marwaha, Carter, Zhou, Wheeler, Bernstein, Wang, He, Zhou, Snyder, Cong, Regev, and Leskovec]{Huang2026BiomniScience}
Kexin Huang, Serena Zhang, Hanchen Wang, Yuanhao Qu, Yingzhou Lu, Ryan Li, Yusuf Roohani, Lin Qiu, Shiyi Cao, Gavin Li, Junze Zhang, Di~Yin, Rick Wierenga, Deniz Kavi, Sherry Liu, Tianwei She, Shruti Marwaha, Jennefer~N. Carter, Xin Zhou, Matthew~T. Wheeler, Jonathan~A. Bernstein, Mengdi Wang, Peng He, Jingtian Zhou, Michael~P. Snyder, Le~Cong, Aviv Regev, and Jure Leskovec.
\newblock {Autonomous biomedical research with an artificial intelligence agent}.
\newblock \emph{Science}, 393\penalty0 (6813):\penalty0 eadz4351, 2026.
\newblock \doi{10.1126/science.adz4351}.
\newblock URL \url{https://www.science.org/doi/10.1126/science.adz4351}.

\bibitem[{Anthropic}(2026)]{anthropic_claude_science}
{Anthropic}.
\newblock Claude science.
\newblock \url{https://claude.com/product/claude-science}, 2026.
\newblock URL \url{https://claude.com/product/claude-science}.
\newblock Accessed: 2026-10-28.

\bibitem[Jin et~al.(2025)Jin, Zhang, Wang, and Cong]{jin2025stella}
Ruofan Jin, Zaixi Zhang, Mengdi Wang, and Le~Cong.
\newblock Stella: Self-evolving llm agent for biomedical research.
\newblock \emph{arXiv preprint arXiv:2507.02004}, 2025.
\newblock \doi{10.48550/arXiv.2507.02004}.
\newblock URL \url{https://arxiv.org/abs/2507.02004}.

\bibitem[{OpenAI}(2025)]{openai_codex}
{OpenAI}.
\newblock Codex is now generally available.
\newblock \url{https://openai.com/index/codex-now-generally-available/}, 2025.
\newblock Published October 6, 2025.

\bibitem[{Anthropic}(2025)]{anthropic_claude_code}
{Anthropic}.
\newblock Claude code documentation.
\newblock \url{https://code.claude.com/docs/en/overview}, 2025.
\newblock URL \url{https://code.claude.com/docs/en/overview}.
\newblock Accessed: 2026-10-28.

\bibitem[{Claude Science} and Shanehsazzadeh(2026)]{claudescience2026binder}
{Claude Science} and Amir Shanehsazzadeh.
\newblock Autonomous de novo protein binder design with claude.
\newblock Technical report, Anthropic, aug 2026.
\newblock URL \url{https://www-cdn.anthropic.com/30bf50e22a01388bb29bf077ee3f244531594b7a.pdf}.

\bibitem[Huang et~al.(2021)Huang, Fu, Gao, Zhao, Roohani, Leskovec, Coley, Xiao, Sun, and Zitnik]{huang2021therapeutics}
Kexin Huang, Tianfan Fu, Wenhao Gao, Yue Zhao, Yusuf Roohani, Jure Leskovec, Connor~W Coley, Cao Xiao, Jimeng Sun, and Marinka Zitnik.
\newblock Therapeutics data commons: Machine learning datasets and tasks for drug discovery and development.
\newblock In \emph{Proceedings of the Neural Information Processing Systems Track on Datasets and Benchmarks}, volume~1. Neural information processing systems foundation, 2021.
\newblock URL \url{https://datasets-benchmarks-proceedings.neurips.cc/paper/2021/hash/4c56ff4ce4aaf9573aa5dff913df997a-Abstract-round1.html}.

\bibitem[Wong et~al.(2024)Wong, Zheng, Valeri, Donghia, Anahtar, Omori, Li, Cubillos-Ruiz, Krishnan, Jin, Manson, Friedrichs, Helbig, Hajian, Fiejtek, Wagner, Soutter, Earl, Stokes, Renner, and Collins]{wong2024discovery}
Felix Wong, Erica~J. Zheng, Jacqueline~A. Valeri, Nina~M. Donghia, Melis~N. Anahtar, Satotaka Omori, Alicia Li, Andres Cubillos-Ruiz, Aarti Krishnan, Wengong Jin, Abigail~L. Manson, Jens Friedrichs, Ralf Helbig, Behnoush Hajian, Dawid~K. Fiejtek, Florence~F. Wagner, Holly~H. Soutter, Ashlee~M. Earl, Jonathan~M. Stokes, Lars~D. Renner, and James~J. Collins.
\newblock Discovery of a structural class of antibiotics with explainable deep learning.
\newblock \emph{Nature}, 626\penalty0 (7997):\penalty0 177--185, 2024.
\newblock \doi{10.1038/s41586-023-06887-8}.
\newblock URL \url{https://doi.org/10.1038/s41586-023-06887-8}.

\bibitem[Rogers and Hahn(2010)]{rogers2010extended}
David Rogers and Mathew Hahn.
\newblock Extended-connectivity fingerprints.
\newblock \emph{Journal of Chemical Information and Modeling}, 50\penalty0 (5):\penalty0 742--754, 2010.
\newblock \doi{10.1021/ci100050t}.
\newblock URL \url{https://pubs.acs.org/doi/10.1021/ci100050t}.

\bibitem[{RDKit contributors}()]{rdkit}
{RDKit contributors}.
\newblock {RDKit}: Open-source cheminformatics.
\newblock \url{https://www.rdkit.org}.
\newblock URL \url{https://www.rdkit.org}.
\newblock Accessed 28 October 2026.

\bibitem[Kl{\"a}ser et~al.(2024)Kl{\"a}ser, Banaszewski, Maddrell-Mander, McLean, M{\"u}ller, Parviz, Huang, and Fitzgibbon]{klaser2024minimol}
Kerstin Kl{\"a}ser, B{\l}a{\.z}ej Banaszewski, Samuel Maddrell-Mander, Callum McLean, Luis M{\"u}ller, Ali Parviz, Shenyang Huang, and Andrew Fitzgibbon.
\newblock {MiniMol}: A parameter-efficient foundation model for molecular learning.
\newblock In \emph{ICML Workshop on Accessible and Efficient Foundation Models for Biological Discovery}, 2024.
\newblock \doi{10.48550/arXiv.2404.14986}.
\newblock URL \url{https://openreview.net/forum?id=uA8zvXmiot}.
\newblock Workshop paper; arXiv version: 2404.14986.

\bibitem[Vinyals et~al.(2016)Vinyals, Bengio, and Kudlur]{vinyals2016order}
Oriol Vinyals, Samy Bengio, and Manjunath Kudlur.
\newblock Order matters: Sequence to sequence for sets.
\newblock In \emph{International Conference on Learning Representations}, 2016.
\newblock URL \url{https://research.google/pubs/order-matters-sequence-to-sequence-for-sets/}.

\bibitem[Chen and Guestrin(2016)]{chen2016xgboost}
Tianqi Chen and Carlos Guestrin.
\newblock {XGBoost}: A scalable tree boosting system.
\newblock In \emph{Proceedings of the 22nd ACM SIGKDD International Conference on Knowledge Discovery and Data Mining}, pages 785--794, 2016.
\newblock \doi{10.1145/2939672.2939785}.
\newblock URL \url{https://doi.org/10.1145/2939672.2939785}.

\bibitem[Lin et~al.(2017)Lin, Goyal, Girshick, He, and Doll{\'a}r]{lin2017focal}
Tsung-Yi Lin, Priya Goyal, Ross Girshick, Kaiming He, and Piotr Doll{\'a}r.
\newblock Focal loss for dense object detection.
\newblock In \emph{Proceedings of the IEEE International Conference on Computer Vision}, pages 2999--3007, 2017.
\newblock \doi{10.1109/ICCV.2017.324}.
\newblock URL \url{https://openaccess.thecvf.com/content_iccv_2017/html/Lin_Focal_Loss_for_ICCV_2017_paper.html}.

\bibitem[Loshchilov and Hutter(2017)]{loshchilov2017sgdr}
Ilya Loshchilov and Frank Hutter.
\newblock {SGDR}: Stochastic gradient descent with warm restarts.
\newblock In \emph{International Conference on Learning Representations}, 2017.
\newblock URL \url{https://arxiv.org/abs/1608.03983}.

\bibitem[Izmailov et~al.(2018)Izmailov, Podoprikhin, Garipov, Vetrov, and Wilson]{izmailov2018averaging}
Pavel Izmailov, Dmitrii Podoprikhin, Timur Garipov, Dmitry Vetrov, and Andrew~Gordon Wilson.
\newblock Averaging weights leads to wider optima and better generalization.
\newblock In \emph{Conference on Uncertainty in Artificial Intelligence}, 2018.
\newblock URL \url{https://arxiv.org/abs/1803.05407}.

\bibitem[Baell and Holloway(2010)]{baell2010pains}
Jonathan~B. Baell and Georgina~A. Holloway.
\newblock New substructure filters for removal of pan assay interference compounds ({PAINS}) from screening libraries and for their exclusion in bioassays.
\newblock \emph{Journal of Medicinal Chemistry}, 53\penalty0 (7):\penalty0 2719--2740, 2010.
\newblock \doi{10.1021/jm901137j}.
\newblock URL \url{https://pubs.acs.org/doi/10.1021/jm901137j}.

\bibitem[Brenk et~al.(2008)Brenk, Schipani, James, Krasowski, Gilbert, Frearson, and Wyatt]{brenk2008lessons}
Ruth Brenk, Alessandro Schipani, Daniel James, Agata Krasowski, Ian~Hugh Gilbert, Julie Frearson, and Paul~Graham Wyatt.
\newblock Lessons learnt from assembling screening libraries for drug discovery for neglected diseases.
\newblock \emph{ChemMedChem}, 3\penalty0 (3):\penalty0 435--444, 2008.
\newblock \doi{10.1002/cmdc.200700139}.
\newblock URL \url{https://chemistry-europe.onlinelibrary.wiley.com/doi/10.1002/cmdc.200700139}.

\bibitem[Butina(1999)]{butina1999unsupervised}
Darko Butina.
\newblock Unsupervised data base clustering based on daylight's fingerprint and tanimoto similarity: A fast and automated way to cluster small and large data sets.
\newblock \emph{Journal of Chemical Information and Computer Sciences}, 39\penalty0 (4):\penalty0 747--750, 1999.
\newblock \doi{10.1021/ci9803381}.
\newblock URL \url{https://pubs.acs.org/doi/10.1021/ci9803381}.

\bibitem[Yang et~al.(2019)Yang, Swanson, Jin, Coley, Eiden, Gao, Guzman-Perez, Hopper, Kelley, Mathea, Palmer, Settels, Jaakkola, Jensen, and Barzilay]{yang2019analyzing}
Kevin Yang, Kyle Swanson, Wengong Jin, Connor Coley, Philipp Eiden, Hua Gao, Angel Guzman-Perez, Timothy Hopper, Brian Kelley, Miriam Mathea, Andrew Palmer, Volker Settels, Tommi Jaakkola, Klavs Jensen, and Regina Barzilay.
\newblock Analyzing learned molecular representations for property prediction.
\newblock \emph{Journal of Chemical Information and Modeling}, 59\penalty0 (8):\penalty0 3370--3388, 2019.
\newblock \doi{10.1021/acs.jcim.9b00237}.
\newblock URL \url{https://pubs.acs.org/doi/10.1021/acs.jcim.9b00237}.

\bibitem[Wilson(1927)]{wilson1927probable}
Edwin~B. Wilson.
\newblock Probable inference, the law of succession, and statistical inference.
\newblock \emph{Journal of the American Statistical Association}, 22\penalty0 (158):\penalty0 209--212, 1927.
\newblock \doi{10.1080/01621459.1927.10502953}.
\newblock URL \url{https://www.tandfonline.com/doi/abs/10.1080/01621459.1927.10502953}.

\bibitem[Cao et~al.(2022)Cao, Coventry, Goreshnik, Huang, Sheffler, Park, Jude, Marković, Kadam, Verschueren, Verstraete, Walsh, Bennett, Phal, Yang, Kozodoy, DeWitt, Picton, Miller, Strauch, DeBouver, Pires, Bera, Halabiya, Hammerson, Yang, Bernard, Stewart, Wilson, Ruohola-Baker, Schlessinger, Lee, Savvides, Garcia, and Baker]{cao2022design}
Longxing Cao, Brian Coventry, Inna Goreshnik, Buwei Huang, William Sheffler, Joon~Sung Park, Kevin~M. Jude, Iva Marković, Rameshwar~U. Kadam, Koen H.~G. Verschueren, Kenneth Verstraete, Scott Thomas~Russell Walsh, Nathaniel Bennett, Ashish Phal, Aerin Yang, Lisa Kozodoy, Michelle DeWitt, Lora Picton, Lauren Miller, Eva-Maria Strauch, Nicholas~D. DeBouver, Allison Pires, Asim~K. Bera, Samer Halabiya, Bradley Hammerson, Wei Yang, Steffen Bernard, Lance Stewart, Ian~A. Wilson, Hannele Ruohola-Baker, Joseph Schlessinger, Sangwon Lee, Savvas~N. Savvides, K.~Christopher Garcia, and David Baker.
\newblock Design of protein-binding proteins from the target structure alone.
\newblock \emph{Nature}, 605\penalty0 (7910):\penalty0 551--560, 2022.
\newblock \doi{10.1038/s41586-022-04654-9}.
\newblock URL \url{https://doi.org/10.1038/s41586-022-04654-9}.

\bibitem[Cao et~al.(2020)Cao, Goreshnik, Coventry, Case, Miller, Kozodoy, Chen, Carter, Walls, Park, Strauch, Stewart, Diamond, Veesler, and Baker]{cao2020novo}
Longxing Cao, Inna Goreshnik, Brian Coventry, James~Brett Case, Lauren Miller, Lisa Kozodoy, Rita~E. Chen, Lauren Carter, Alexandra~C. Walls, Young-Jun Park, Eva-Maria Strauch, Lance Stewart, Michael~S. Diamond, David Veesler, and David Baker.
\newblock De novo design of picomolar sars-cov-2 miniprotein inhibitors.
\newblock \emph{Science}, 370\penalty0 (6515):\penalty0 426--431, 2020.
\newblock \doi{10.1126/science.abd9909}.
\newblock URL \url{https://doi.org/10.1126/science.abd9909}.

\bibitem[Bennett et~al.(2023)Bennett, Coventry, Goreshnik, Huang, Allen, Vafeados, Peng, Dauparas, Baek, Stewart, DiMaio, De~Munck, Savvides, and Baker]{bennett2023improving}
Nathaniel~R. Bennett, Brian Coventry, Inna Goreshnik, Buwei Huang, Aza Allen, Dionne Vafeados, Ying~Po Peng, Justas Dauparas, Minkyung Baek, Lance Stewart, Frank DiMaio, Steven De~Munck, Savvas~N. Savvides, and David Baker.
\newblock Improving de novo protein binder design with deep learning.
\newblock \emph{Nature Communications}, 14\penalty0 (1):\penalty0 2625, 2023.
\newblock \doi{10.1038/s41467-023-38328-5}.
\newblock URL \url{https://doi.org/10.1038/s41467-023-38328-5}.

\bibitem[Liu et~al.(2025)Liu, Greenwood, Bonzanini, Motmaen, Meyerberg, Dao, Xiang, Ault, Sharp, Wang, Visani, Vafeados, Roullier, Nourmohammad, Scheinberg, Garcia, and Baker]{liu2025design}
Bingxu Liu, Nathan~F. Greenwood, Julia~E. Bonzanini, Amir Motmaen, Jeremy Meyerberg, Tao Dao, Xinyu Xiang, Russell Ault, Jazmin Sharp, Chunyu Wang, Gian~Marco Visani, Dionne~K. Vafeados, Nicole Roullier, Armita Nourmohammad, David~A. Scheinberg, K.~Christopher Garcia, and David Baker.
\newblock Design of high-specificity binders for peptide–mhc-i complexes.
\newblock \emph{Science}, 389\penalty0 (6758):\penalty0 386--391, 2025.
\newblock \doi{10.1126/science.adv0185}.
\newblock URL \url{https://doi.org/10.1126/science.adv0185}.

\bibitem[Stark et~al.(2025)Stark, Faltings, Choi, Xie, Hur, O’Donnell, Bushuiev, Uçar, Passaro, Mao, Reveiz, Bushuiev, Portnoi, Pluskal, Sivic, Kreis, Vahdat, Ray, Goldstein, Savinov, Hambalek, Gupta, Taquiri-Diaz, Zhang, Snyder, Hatstat, Arada, Kim, Fan, Tackie-Yarboi, Boselli, Schnaider, Liu, Li, Hnisz, Sabatini, DeGrado, Wohlwend, Corso, Barzilay, and Jaakkola]{stark2025boltzgen}
Hannes Stark, Felix Faltings, MinGyu Choi, Yuxin Xie, Eunsu Hur, Timothy O’Donnell, Anton Bushuiev, Talip Uçar, Saro Passaro, Weian Mao, Mateo Reveiz, Roman Bushuiev, Tally Portnoi, Tomáš Pluskal, Josef Sivic, Karsten Kreis, Arash Vahdat, Shamayeeta Ray, Jonathan~T. Goldstein, Andrew Savinov, Jacob~A. Hambalek, Anshika Gupta, Diego~A. Taquiri-Diaz, Yaotian Zhang, Samuel~J. Snyder, A.~Katherine Hatstat, Angelika Arada, Nam~Hyeong Kim, Haoyu Fan, Ethel Tackie-Yarboi, Dylan Boselli, Lee Schnaider, Chang~C. Liu, Gene-Wei Li, Denes Hnisz, David~M. Sabatini, William~F. DeGrado, Jeremy Wohlwend, Gabriele Corso, Regina Barzilay, and Tommi Jaakkola.
\newblock Boltzgen: Toward universal binder design.
\newblock \emph{bioRxiv}, 2025.
\newblock \doi{10.1101/2025.11.20.689494}.
\newblock URL \url{https://www.biorxiv.org/content/10.1101/2025.11.20.689494v2}.
\newblock Preprint; version 2 posted 2026-06-16.

\bibitem[Pacesa et~al.(2025)Pacesa, Nickel, Schellhaas, Schmidt, Pyatova, Kissling, Barendse, Choudhury, Kapoor, Alcaraz-Serna, Cho, Ghamary, Vinué, Yachnin, Wollacott, Buckley, Westphal, Lindhoud, Georgeon, Goverde, Hatzopoulos, Gönczy, Muller, Schwank, Swarts, Vecchio, Schneider, Ovchinnikov, and Correia]{pacesa2025one}
Martin Pacesa, Lennart Nickel, Christian Schellhaas, Joseph Schmidt, Ekaterina Pyatova, Lucas Kissling, Patrick Barendse, Jagrity Choudhury, Srajan Kapoor, Ana Alcaraz-Serna, Yehlin Cho, Kourosh~H. Ghamary, Laura Vinué, Brahm~J. Yachnin, Andrew~M. Wollacott, Stephen Buckley, Adrie~H. Westphal, Simon Lindhoud, Sandrine Georgeon, Casper~A. Goverde, Georgios~N. Hatzopoulos, Pierre Gönczy, Yannick~D. Muller, Gerald Schwank, Daan~C. Swarts, Alex~J. Vecchio, Bernard~L. Schneider, Sergey Ovchinnikov, and Bruno~E. Correia.
\newblock One-shot design of functional protein binders with bindcraft.
\newblock \emph{Nature}, 646\penalty0 (8084):\penalty0 483--492, 2025.
\newblock \doi{10.1038/s41586-025-09429-6}.
\newblock URL \url{https://doi.org/10.1038/s41586-025-09429-6}.

\bibitem[{Protenix Team} et~al.(2025){Protenix Team}, Ren, Sun, Guan, Liu, Gong, Wang, Wang, Cai, Ma, Zhang, Liu, Zhang, Chen, and Xiao]{team2025pxdesign}
{Protenix Team}, Milong Ren, Jinyuan Sun, Jiaqi Guan, Cong Liu, Chengyue Gong, Yuzhe Wang, Lan Wang, Qixu Cai, Wenzhi Ma, Yuxuan Zhang, Zhenyu Liu, Hanyu Zhang, Xinshi Chen, and Wenzhi Xiao.
\newblock Pxdesign: Fast, modular, and accurate de novo design of protein binders.
\newblock \emph{bioRxiv}, 2025.
\newblock \doi{10.1101/2025.08.15.670450}.
\newblock URL \url{https://www.biorxiv.org/content/10.1101/2025.08.15.670450v3}.
\newblock Preprint; version 3 posted 2025-12-17.

\bibitem[Passaro et~al.(2025)Passaro, Corso, Wohlwend, Reveiz, Thaler, Somnath, Getz, Portnoi, Roy, Stark, Kwabi-Addo, Beaini, Jaakkola, and Barzilay]{passaro2025boltz}
Saro Passaro, Gabriele Corso, Jeremy Wohlwend, Mateo Reveiz, Stephan Thaler, Vignesh~Ram Somnath, Noah Getz, Tally Portnoi, Julien Roy, Hannes Stark, David Kwabi-Addo, Dominique Beaini, Tommi Jaakkola, and Regina Barzilay.
\newblock Boltz-2: Towards accurate and efficient binding affinity prediction.
\newblock \emph{bioRxiv}, 2025.
\newblock \doi{10.1101/2025.06.14.659707}.
\newblock URL \url{https://www.biorxiv.org/content/10.1101/2025.06.14.659707v1}.
\newblock Preprint; version 1 posted 2025-06-18.

\bibitem[Salentin et~al.(2015)Salentin, Schreiber, Haupt, Adasme, and Schroeder]{salentin2015plip}
Sebastian Salentin, Sven Schreiber, V.~Joachim Haupt, Melissa~F. Adasme, and Michael Schroeder.
\newblock {PLIP}: fully automated protein--ligand interaction profiler.
\newblock \emph{Nucleic Acids Research}, 43\penalty0 (W1):\penalty0 W443--W447, 2015.
\newblock \doi{10.1093/nar/gkv315}.
\newblock URL \url{https://academic.oup.com/nar/article/43/W1/W443/2467865}.

\bibitem[Evans et~al.(2021)Evans, O’Neill, Pritzel, Antropova, Senior, Green, Žídek, Bates, Blackwell, Yim, Ronneberger, Bodenstein, Zielinski, Bridgland, Potapenko, Cowie, Tunyasuvunakool, Jain, Clancy, Kohli, Jumper, and Hassabis]{evans2021protein}
Richard Evans, Michael O’Neill, Alexander Pritzel, Natasha Antropova, Andrew Senior, Tim Green, Augustin Žídek, Russ Bates, Sam Blackwell, Jason Yim, Olaf Ronneberger, Sebastian Bodenstein, Michal Zielinski, Alex Bridgland, Anna Potapenko, Andrew Cowie, Kathryn Tunyasuvunakool, Rishub Jain, Ellen Clancy, Pushmeet Kohli, John Jumper, and Demis Hassabis.
\newblock Protein complex prediction with alphafold-multimer.
\newblock \emph{bioRxiv}, 2021.
\newblock \doi{10.1101/2021.10.04.463034}.
\newblock URL \url{https://www.biorxiv.org/content/10.1101/2021.10.04.463034v2}.
\newblock Preprint; version 2 posted 2022-03-10.

\bibitem[Gong et~al.(2025)Gong, Chen, Zhang, Song, Zhou, and Xiao]{gong2025protenixmini}
Chengyue Gong, Xinshi Chen, Yuxuan Zhang, Yuxuan Song, Hao Zhou, and Wenzhi Xiao.
\newblock {Protenix-Mini}: Efficient structure predictor via compact architecture, few-step diffusion and switchable {pLM}, 2025.
\newblock URL \url{https://arxiv.org/abs/2507.11839}.

\bibitem[Candido et~al.(2026)Candido, Hayes, Derry, Rao, Lin, Verkuil, Wu, Lee, Bruguera, Keval, Kopylov, Pak, Wu, Thomas, Mataraso, Hsu, Trotman-Grant, Fatras, dos Santos~Costa, Badkundri, Akın, Oktay, Deaton, Montabana, Sitwala, Yu, Wiggert, Carlin, Goering, Blazejewski, Sandora, Hla, Jia, Kloker, Sofroniew, Uehara, Pannu, Bachas, Liu, Sercu, and Rives]{candido2026language}
Salvatore Candido, Thomas Hayes, Alexander Derry, Roshan Rao, Zeming Lin, Robert Verkuil, Bryan Wu, Jin~Sub Lee, Elise Bruguera, Jehan Keval, Mykhailo Kopylov, John Pak, Wesley Wu, Neil Thomas, Samson Mataraso, Alvin Hsu, Ashton Trotman-Grant, Kilian Fatras, Allan dos Santos~Costa, Rohil Badkundri, Halil Akın, Deniz Oktay, Jonathan Deaton, Elizabeth Montabana, Hrishita Sitwala, Yue Yu, Marius Wiggert, Dylan~Alexander Carlin, Anthony Goering, Tomasz Blazejewski, McCullen Sandora, Michael Hla, Tina Jia, Leon Kloker, Nicholas Sofroniew, Masatoshi Uehara, Jassi Pannu, Sharrol Bachas, Daniel Liu, Tom Sercu, and Alexander Rives.
\newblock Language modeling materializes a world model of protein biology.
\newblock \emph{bioRxiv}, 2026.
\newblock \doi{10.64898/2026.06.03.729735}.
\newblock URL \url{https://www.biorxiv.org/content/10.64898/2026.06.03.729735v1}.
\newblock Preprint; version 1 posted 2026-06-04.

\bibitem[Butcher et~al.(2025)Butcher, Krishna, Mitra, Brent, Li, Corley, Kim, Funk, Mathis, Salike, Muraishi, Eisenach, Thompson, Chen, Politanska, Sehgal, Coventry, Zhang, Qiang, Didi, Kazman, DiMaio, and Baker]{butcher2025novo}
Jasper Butcher, Rohith Krishna, Raktim Mitra, Rafael~I. Brent, Yanjing Li, Nathaniel Corley, Paul~T. Kim, Jonathan Funk, Simon Mathis, Saman Salike, Aiko Muraishi, Helen Eisenach, Tuscan~Rock Thompson, Jie Chen, Yuliya Politanska, Enisha Sehgal, Brian Coventry, Odin Zhang, Bo~Qiang, Kieran Didi, Max Kazman, Frank DiMaio, and David Baker.
\newblock De novo design of all-atom biomolecular interactions with rfdiffusion3.
\newblock \emph{bioRxiv}, 2025.
\newblock \doi{10.1101/2025.09.18.676967}.
\newblock URL \url{https://www.biorxiv.org/content/10.1101/2025.09.18.676967v2}.
\newblock Preprint; version 2 posted 2025-11-19.

\bibitem[Didi et~al.(2026)Didi, Zhang, Zhou, Reidenbach, Cao, Cha, Geffner, Dallago, Tang, Bronstein, Steinegger, Kucukbenli, Vahdat, and Kreis]{didi2026scaling}
Kieran Didi, Zuobai Zhang, Guoqing Zhou, Danny Reidenbach, Zhonglin Cao, Sooyoung Cha, Tomas Geffner, Christian Dallago, Jian Tang, Michael Bronstein, Martin Steinegger, Emine Kucukbenli, Arash Vahdat, and Karsten Kreis.
\newblock Scaling atomistic protein binder design with generative pretraining and test-time compute.
\newblock In C.~Vondrick, B.~Hariharan, C.~Raffel, L.~Pinto, D.~Yang, and A.~Faust, editors, \emph{International Conference on Learning Representations}, volume 2026, pages 61056--61109, 2026.
\newblock URL \url{https://proceedings.iclr.cc/paper_files/paper/2026/file/639d992f819c2b40387d4d5170b8ffd7-Paper-Conference.pdf}.

\bibitem[Boyd et~al.(2025)Boyd, Guns, and {Escalante Bio}]{boyd2025mosaic}
N.~Boyd, S.~Guns, and {Escalante Bio}.
\newblock {MOSAIC}.
\newblock \url{https://github.com/escalante-bio/mosaic}, 2025.
\newblock URL \url{https://github.com/escalante-bio/mosaic}.

\bibitem[Goverde et~al.(2024)Goverde, Pacesa, Goldbach, Dornfeld, Balbi, Georgeon, Rosset, Kapoor, Choudhury, Dauparas, Schellhaas, Kozlov, Baker, Ovchinnikov, Vecchio, and Correia]{goverde2024computational}
Casper~A. Goverde, Martin Pacesa, Nicolas Goldbach, Lars~J. Dornfeld, Petra E.~M. Balbi, Sandrine Georgeon, Stéphane Rosset, Srajan Kapoor, Jagrity Choudhury, Justas Dauparas, Christian Schellhaas, Simon Kozlov, David Baker, Sergey Ovchinnikov, Alex~J. Vecchio, and Bruno~E. Correia.
\newblock Computational design of soluble and functional membrane protein analogues.
\newblock \emph{Nature}, 631\penalty0 (8020):\penalty0 449--458, 2024.
\newblock \doi{10.1038/s41586-024-07601-y}.
\newblock URL \url{https://doi.org/10.1038/s41586-024-07601-y}.

\bibitem[Shuai et~al.(2025)Shuai, Lu, Bhatti, Kouba, and Huang]{shuai2025caliby}
Richard~W. Shuai, Tianyu Lu, Subhang Bhatti, Petr Kouba, and Po-Ssu Huang.
\newblock Ensemble-conditioned protein sequence design with {Caliby}.
\newblock \emph{bioRxiv}, 2025.
\newblock \doi{10.1101/2025.09.30.679633}.
\newblock URL \url{https://www.biorxiv.org/content/10.1101/2025.09.30.679633}.
\newblock Preprint; also presented at the Machine Learning in Structural Biology workshop (MLSB 2025).

\bibitem[Zhang et~al.(2026)Zhang, Gong, Sun, Guan, Ren, Xue, Zhang, Ma, Liu, Chen, and Xiao]{zhang2026protenix}
Yuxuan Zhang, Chengyue Gong, Jinyuan Sun, Jiaqi Guan, Milong Ren, Song Xue, Hanyu Zhang, Wenzhi Ma, Zhenyu Liu, Xinshi Chen, and Wenzhi Xiao.
\newblock Protenix-v2: Broadening the reach of structure prediction and biomolecular design.
\newblock \emph{bioRxiv}, 2026.
\newblock \doi{10.64898/2026.04.10.717613}.
\newblock URL \url{https://www.biorxiv.org/content/10.64898/2026.04.10.717613v1}.
\newblock Preprint; version 1 posted 2026-04-11.

\bibitem[Dunbrack(2025)]{dunbrack2025res}
Roland~L. Dunbrack, Jr.
\newblock Rēs ipsae loquuntur: What's wrong with alphafold's iptm score and how to fix it.
\newblock \emph{bioRxiv}, 2025.
\newblock \doi{10.1101/2025.02.10.637595}.
\newblock URL \url{https://www.biorxiv.org/content/10.1101/2025.02.10.637595v2}.
\newblock Preprint; version 2 posted 2025-12-15.

\bibitem[Basu and Wallner(2016)]{basu2016dockq}
Sankar Basu and Bj{\"o}rn Wallner.
\newblock {DockQ}: A quality measure for protein--protein docking models.
\newblock \emph{PLOS ONE}, 11\penalty0 (8):\penalty0 e0161879, 2016.
\newblock \doi{10.1371/journal.pone.0161879}.
\newblock URL \url{https://journals.plos.org/plosone/article?id=10.1371/journal.pone.0161879}.

\end{thebibliography}
\end{document}